\documentclass[final,5p,times,twocolumn,authoryear]{elsarticle}

\usepackage{lineno}
\usepackage[hidelinks]{hyperref}
\usepackage{url}
\usepackage{siunitx}
\usepackage{enumitem}
\usepackage{booktabs}
\usepackage{graphicx}
\usepackage{lscape}
\usepackage{multirow}
\usepackage{longtable,tabu}
\usepackage{perpage}
\usepackage{makecell}
\MakePerPage{footnote}

\begin{document}

\begin{frontmatter}



\title{A review of machine learning in processing remote sensing data for mineral exploration}


\author[fn1]{Hojat Shirmard}
\fntext[fn1]{School of Mining Engineering, College of Engineering, University of Tehran, Tehran, Iran}

\author[fn2]{Ehsan Farahbakhsh\corref{cor1}}
\fntext[fn2]{Department of Mining Engineering, Amirkabir University of Technology (Tehran Polytechnic), Tehran, Iran}
\ead{e.farahbakhsh@aut.ac.ir}
\cortext[cor1]{Corresponding author}

\author[fn3]{R. Dietmar M\"uller}
\fntext[fn3]{EarthByte Group, School of Geosciences, University of Sydney, NSW 2006, Australia}

\author[fn4,fn5]{Rohitash Chandra}
\fntext[fn4]{School of Mathematics and Statistics, University of New South Wales, Sydney, NSW 2052, Australia}
\fntext[fn5]{UNSW Data Science Hub, University of New South Wales, Sydney, NSW 2052, Australia}


\begin{abstract}
The decline of the number of newly discovered mineral deposits and increase in demand for different minerals in recent years has led exploration geologists to look for more efficient and innovative methods for processing different data types at each stage of mineral exploration. As a primary step, various features, such as lithological units, alteration types, structures, and indicator minerals, are mapped to aid decision-making in targeting ore deposits. Different types of remote sensing datasets, such as satellite and airborne data, make it possible to overcome common problems associated with mapping geological features. The rapid increase in the volume of remote sensing data obtained from different platforms has encouraged scientists to develop advanced, innovative, and robust data processing methodologies. Machine learning methods can help process a wide range of remote sensing datasets and determine the relationship between components such as the reflectance continuum and features of interest. These methods are robust in processing spectral and ground truth measurements against noise and uncertainties. In recent years, many studies have been carried out by supplementing geological surveys with remote sensing datasets, which is now prominent in geoscience research. This paper provides a comprehensive review of the implementation and adaptation of some popular and recently established machine learning methods for processing different types of remote sensing data and investigates their applications for detecting various ore deposit types. We demonstrate the high capability of combining remote sensing data and machine learning methods for mapping different geological features that are critical for providing potential maps. Moreover, we find there is scope for advanced methods such as deep learning to process the new generation of remote sensing data that provide high spatial and spectral resolution for creating improved mineral prospectivity maps.
\end{abstract}


\end{frontmatter}


\section{Introduction}
\label{sec1}
One of the fundamental steps in mineral exploration is localizing the geological features related to target mineralization by providing and investigating geological maps. These maps involve different features such as lithological units, alteration types, structures, and indicator minerals \citep{Brimhall2005,Ninomiya2005,Rowan2006,Gad2007,Pour2019}. Over time, geological mapping methods have evolved; and nowadays, the combination of remote sensing data and advanced data analytics such as machine learning is gaining much attention \citep{Cracknell2014,Harvey2016,Bachri2019,Chakouri2020}. This combination helps geologists overcome common challenges of traditional methods such as subjective judgment that can provide reliable maps and avoid wasting money on prospecting for barren regions.\\
There are numerous difficulties associated with mapping geological features, particularly in areas that are hard to access. Traditionally, this task has been carried out by expert knowledge of survey lines, navigation systems, and data collection in the field. Moreover, fieldwork is affected by climate conditions, topography, field experts, and operating approaches \citep{Latifovic2018,Sang2020}. Remote sensing data collected in different spatial, spectral, and temporal resolutions enable geologists to provide a solution for most of the challenges and shortcomings involved with geological field mapping \citep{Harris2011,Pour2016a,Dai2017}. Based on the data type and exploration stage, geological maps can be created on small or large scales \citep{Usui2010,Bartalev2014}. Remote sensing technology can play an important role in geological survey, mapping, analysis, and interpretation. It offers a unique ability to investigate geological features of remote regions on the Earth's surface \citep{Al-Nahmi2017}.\\
Remotely sensed multispectral imaging (MSI) has historically been used for the visual analysis of geological formations and lithological units \citep{Goetz1981}. Early studies with prototypes of airborne imaging spectrometers (AIS) demonstrated their ability for detecting different features such as indicator minerals \citep{Asadzadeh2016}. The use of satellite imagery for mapping geological features has been extended with the latest advances in multispectral and hyperspectral remote sensing instruments such as enhanced thematic mapper plus (ETM+), operational land imager (OLI), advanced space-borne thermal emission and reflection radiometer (ASTER), and Hyperion \citep{Rezaei2020}. A new insight into geological mapping improvement has been offered by remote sensing data and innovations in digital image processing methods \citep{Bachri2019}. Lithological discrimination \citep{Leverington2012,Black2016,Testa2018,Metelka2018,Ninomiya2019}, alteration \citep{Rowan2006,Kratt2010} and structural mapping \citep{Raharimahefa2009}, and delineating rocks and minerals \citep{Mahanta2018} have been made possible by processing spectral images obtained by airborne to space-borne instruments.\\
The unprecedented rise in the diversity of remote sensing data obtained from different platforms, as well as ground measurements, has enabled scientists to provide innovative and effective methodologies for data processing \citep{Ali2015}. A large number of image processing methods have been developed over the last decades to identify, discriminate, and enhance features such as lithological units, alteration zones, and structures to aid in discovering mineral deposits using remote sensing data \citep{Shirmard2020}. Image processing methods are mainly used for enhancement, feature extraction and detection, segmentation or classification, fusion, change detection, and compression of satellite images \citep{Asokan2020}. These tasks are conducted by applying various mathematical algorithms to extract useful information \citep{Babbar2019}.\\
The abundance of computational power and the advent of big data and machine learning help geologists overcome the issues they need to deal with during different stages of mineral exploration. Mathematical geologists have adopted advanced computer and software tools for tasks such as data interpolation (e.g., using geostatistical methods) \citep{VanderMeer1994}, mapping singularities \citep{Cheng1999}, separating anomalous zones \citep{Cheng2007}, and integrating data layers \citep{Farahbakhsh2020}. The new knowledge obtained through digital analysis and novel methods of data mining is greatly benefitting human decision-making. Machine learning as a subdomain of artificial intelligence has been considered reliable since it can accurately and efficiently classify remotely sensed imagery \citep{Maxwell2018}. The combination of machine learning with other methods like geostatistics can also help in analyzing remote sensing data \citep{Varouchakis2021}. However, this combination has less been considered for mineral exploration. Machine learning makes it possible to manage high dimensional data and map features with complicated characteristics \citep{Maxwell2018}. The combined use of remote sensing data and machine learning algorithms have proven to facilitate and improve mineral exploration. Machine learning methods draw a growing interest in the area of remote sensing data analysis as a solution to the problems of geological or mineral exploration \citep{Bachri2019}. It is important to provide a roadmap of work in this area of interest, given the rapid development of machine learning and deep learning methods.\\
In this paper, we provide a comprehensive review of the application and adoption of machine learning methods in remote sensing data processing for modeling geological patterns and exploring ore deposits. Firstly, we discuss the characteristics of remote sensing data that are popular in the community of exploration geologists and obtained by satellites, airplanes, drones, and ground-based instruments. Secondly, we review the application of remote sensing data in mapping lithological units, alteration types, structures, and indicator minerals known as key features in discovering ore deposits. Thirdly, we look into the progress of the most popular and recently used machine learning methods for processing remote sensing data focusing on mineral exploration. We classify the machine learning methods in our study into five groups that include dimensionality reduction, classification, clustering, regression, and deep learning methods. Finally, we discuss challenges and highlight potential future works given interdisciplinary focus of our review. This paper provides a roadmap of the development of the field with some of the most recent deep learning approaches that have not been used for remote sensing beforehand, such as graph deep learning methods, Bayesian deep learning, variational autoencoders, and other newcomers such as transformer recurrent neural networks.

\section{Methodology}
\label{sec2}
We begin by providing a categorization scheme that groups different machine methods and then present a literature review of machine learning methods relevant to the mineral exploration industry. We review those publications demonstrating the application of such methods in remote sensing data processing via detection of geological target features. The main keywords used for searching documents on Google Scholar and Scopus include "machine learning", "remote sensing", and "mineral exploration", determined by consulting with experts and according to previous studies. Moreover, we use appropriate keywords to search for the documents focused on applying machine learning methods for delineating geological target features, including lithological units, alteration types, structures, and indicator minerals known as key elements in mineral exploration that can be mapped using remote sensing data.\\
We plot the frequency of publications on this topic using different keywords. Figure 1a shows the number of publications in recent years using the keywords of "machine learning" and "remote sensing" based on the results obtained by searching documents in the Scopus research publication database. Figure 1b shows the number of publications using the keywords that include  "machine learning", "remote sensing", and "mineral". Figure 1c shows the number of publications using the keywords that include "machine learning", "remote sensing", and "mineral exploration", which are in the scope of this review paper. As shown in these plots, the number of publications that focus on the applications of machine learning methods in processing remote sensing data has continuously increased in the last decade.\\
Remote sensing is a broad field with a wide range of applications, and only a small portion of these studies are related to mineral exploration. Although the number of publications shown in Figure 1c is much lower than Figure 1a, both show an increasing trend that indicates the combination of remote sensing data and machine learning methods is a hot topic and is gradually getting attention from industry experts and researchers. The low number of publications in Figure 1c also reveals the gap for developing the application of remote sensing and machine learning methods in detecting mineral deposits with potential benefits to the community of exploration geologists.

\begin{figure}
  \centering
  \frame{\includegraphics[width=0.6\linewidth]{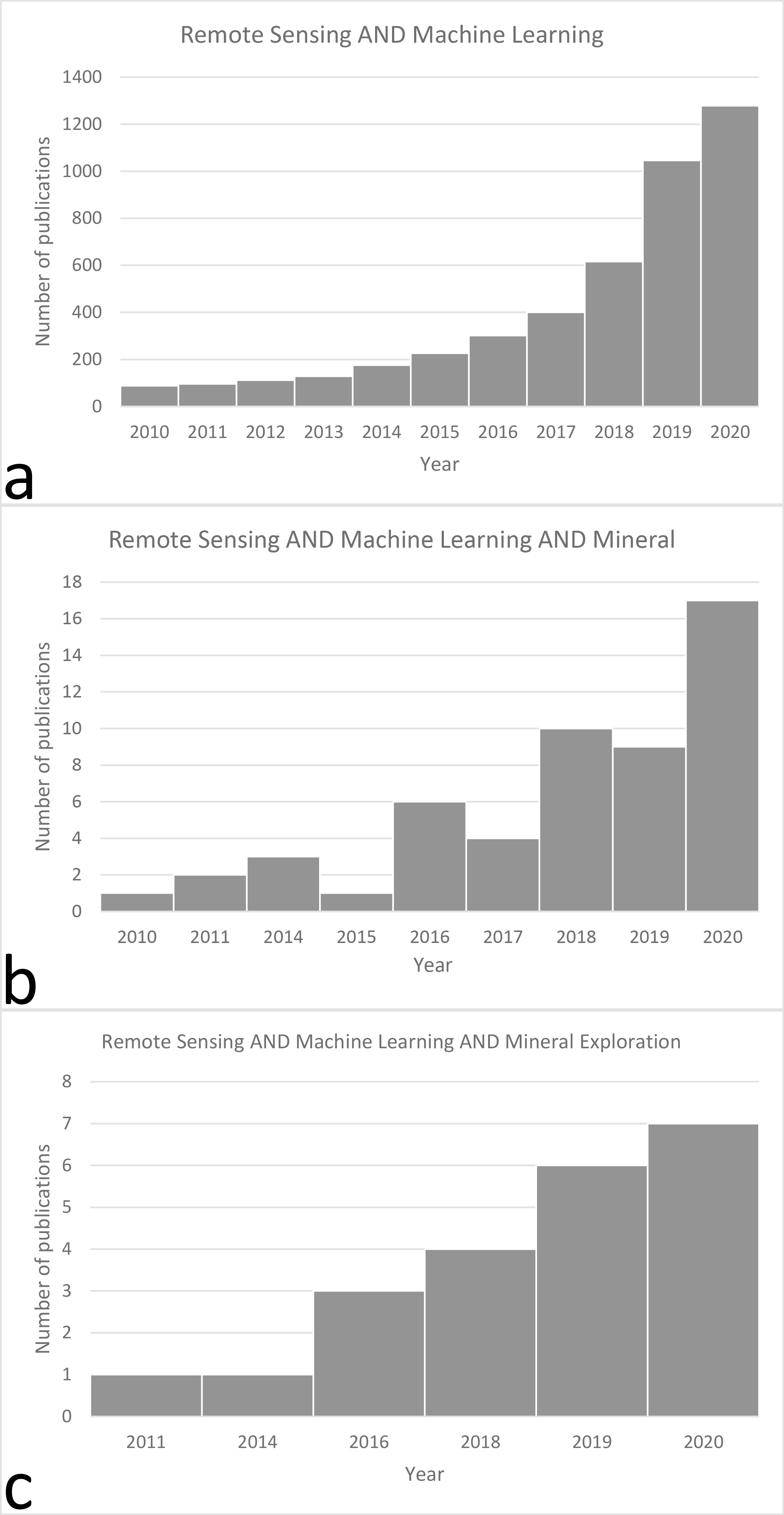}}
  \caption{Number of publications in the last decade based on the Scopus research publication database and obtained by searching keywords: a) “machine learning” and “remote sensing”, b) “machine learning”, “remote sensing”, and ‘mineral’, and c) “machine learning”, “remote sensing”, and “mineral exploration”.}
  \label{fig1}
\end{figure}

\section{Remote sensing data}
\label{sec3}
Our focus is to review and categorize remote sensing data that are usually used for mapping geological features, particularly related to mineralization based on previous studies. The data acquiring platforms include satellites, airborne, and ground-based instruments \citep{Prost2014,Toth2016}. We summarize the characteristics of different popular remote sensing data in mineral exploration based on their platforms in Table 1.

\begin{table*}[htbp!]
\centering
\caption{Characteristics of popular remote sensing data obtained by different platforms that are used for mapping geological features.}
\label{table1_1}
\resizebox{\textwidth}{!}{%
\begin{tabular}{|l|l|l|l|l|l|l|}
\hline
Satellite/Sensor & Subsystem & Band Number/Name & Spectral Range ($\mu$) & Ground Resolution (m) & Swath Width (km) & Year of Launch \\ \hline
\multirow{11}{*}{Landsat 5} & \multirow{4}{*}{MSS} & Band 4 Green & 0.50--0.60 & \multirow{4}{*}{57$\times$79} & \multirow{11}{*}{185} & \multirow{11}{*}{1984} \\ \cline{3-4}
 &  & Band 5 Red & 0.60--0.70 &  &  &  \\ \cline{3-4}
 &  & Band 6 NIR 1 & 0.70--0.80 &  &  &  \\ \cline{3-4}
 &  & Band 7 NIR 2 & 0.80--1.10 &  &  &  \\ \cline{2-5}
 & \multirow{8}{*}{TM} & Band 1 Blue & 0.45--0.52 & \multirow{5}{*}{30} &  &  \\ \cline{3-4}
 &  & Band 2 Red & 0.52--0.60 &  &  &  \\ \cline{3-4}
 &  & Band 3 Green & 0.63--0.69 &  &  &  \\ \cline{3-4}
 &  & Band 4 NIR & 0.76--0.90 &  &  &  \\ \cline{3-4}
 &  & Band 5 SWIR 1 & 1.55--1.75 &  &  &  \\ \cline{3-5}
 &  & Band 6 Thermal & 10.40--12.50 & 120 &  &  \\ \cline{3-5}
 &  & Band 7 SWIR 2 & 2.08--2.35 & 30 &  &  \\ \hline
\multirow{8}{*}{Landsat 7} & \multirow{8}{*}{ETM+} & Band 1 Blue & 0.45--0.52 & \multirow{5}{*}{30} & \multirow{8}{*}{185} & \multirow{8}{*}{1999} \\ \cline{3-4}
 &  & Band 2 Red & 0.52--0.60 &  &  &  \\ \cline{3-4}
 &  & Band 3 Green & 0.63--0.69 &  &  &  \\ \cline{3-4}
 &  & Band 4 NIR & 0.77--0.90 &  &  &  \\ \cline{3-4}
 &  & Band 5 SWIR 1 & 1.55--1.75 &  &  &  \\ \cline{3-5}
 &  & Band 6 Thermal & 10.40--12.50 & 60 &  &  \\ \cline{3-5}
 &  & Band 7 SWIR 2 & 2.08--2.35 & 30 &  &  \\ \cline{3-5}
 &  & Band 8 Panchromatic & 0.52--0.90 & 15 &  &  \\ \hline
\multirow{11}{*}{Landsat 8} & \multirow{9}{*}{OLI} & Band 1 Coastal Aerosol & 0.43--0.45 & \multirow{7}{*}{30} & \multirow{11}{*}{185} & \multirow{11}{*}{2013} \\ \cline{3-4}
 &  & Band 2 Blue & 0.45--0.51 &  &  &  \\ \cline{3-4}
 &  & Band 3 Green & 0.53--0.59 &  &  &  \\ \cline{3-4}
 &  & Band 4 Red & 0.64--0.67 &  &  &  \\ \cline{3-4}
 &  & Band 5 NIR & 0.85--0.88 &  &  &  \\ \cline{3-4}
 &  & Band 6 SWIR 1 & 1.57--1.65 &  &  &  \\ \cline{3-4}
 &  & Band 7 SWIR 2 & 2.11--2.29 &  &  &  \\ \cline{3-5}
 &  & Band 8 Panchromatic & 0.50--0.68 & 15 &  &  \\ \cline{3-5}
 &  & Band 9 Cirrus & 1.36--1.38 & 30 &  &  \\ \cline{2-5}
 & \multirow{2}{*}{TIRS} & Band 10 TIRS 1 & 10.60--11.19 & \multirow{2}{*}{100} &  &  \\ \cline{3-4}
 &  & Band 11 TIRS 2 & 11.50--12.51 &  &  &  \\ \hline
\multirow{14}{*}{ASTER} & \multirow{3}{*}{VNIR} & Band 1 & 0.520--0.600 & \multirow{3}{*}{15} & \multirow{14}{*}{60} & \multirow{14}{*}{1999} \\ \cline{3-4}
 &  & Band 2 & 0.630--0.690 &  &  &  \\ \cline{3-4}
 &  & Band 3 & 0.780--0.860 &  &  &  \\ \cline{2-5}
 & \multirow{6}{*}{SWIR} & Band 4 & 1.600--1.700 & \multirow{6}{*}{30} &  &  \\ \cline{3-4}
 &  & Band 5 & 2.145--2.185 &  &  &  \\ \cline{3-4}
 &  & Band 6 & 2.185--2.225 &  &  &  \\ \cline{3-4}
 &  & Band 7 & 2.235--2.285 &  &  &  \\ \cline{3-4}
 &  & Band 8 & 2.295--2.365 &  &  &  \\ \cline{3-4}
 &  & Band 9 & 2.360--2.430 &  &  &  \\ \cline{2-5}
 & \multirow{5}{*}{TIR} & Band 10 & 8.125--8.475 & \multirow{5}{*}{90} &  &  \\ \cline{3-4}
 &  & Band 11 & 8.475--8.825 &  &  &  \\ \cline{3-4}
 &  & Band 12 & 8.925--9.275 &  &  &  \\ \cline{3-4}
 &  & Band 13 & 10.250--10.950 &  &  &  \\ \cline{3-4}
 &  & Band 14 & 10.950--11.650 &  &  &  \\ \hline
\end{tabular}%
}
\end{table*}

\begin{table*}[htbp!]
\centering
\label{table1_2}
{\footnotesize Continued from the previous page.}
\resizebox{\textwidth}{!}{%
\begin{tabular}{|l|l|l|l|l|l|l|}
\hline
Satellite/Sensor & Subsystem & Band Number/Name & Spectral Range ($\mu$) & Ground Resolution (m) & Swath Width (km) & Year of Launch \\ \hline
\multicolumn{2}{|l|}{WorldView-1} & Panchromatic & 0.45--0.90 & 0.5 & 17.6 & 2007 \\ \hline
\multicolumn{2}{|l|}{\multirow{9}{*}{WorldView-2}} & Panchromatic & 0.450--0.800 & 0.46 & \multirow{9}{*}{16.4} & \multirow{9}{*}{2009} \\ \cline{3-5}
\multicolumn{2}{|l|}{} & Bnad 1 Coastal Blue & 0.400--0.450 & \multirow{8}{*}{1.84} &  &  \\ \cline{3-4}
\multicolumn{2}{|l|}{} & Bnad 2 Blue & 0.450--0.510 &  &  &  \\ \cline{3-4}
\multicolumn{2}{|l|}{} & Bnad 3 Green & 0.510--0.580 &  &  &  \\ \cline{3-4}
\multicolumn{2}{|l|}{} & Band 4 Yellow & 0.585--0.625 &  &  &  \\ \cline{3-4}
\multicolumn{2}{|l|}{} & Bnad 5 Red & 0.630--0.690 &  &  &  \\ \cline{3-4}
\multicolumn{2}{|l|}{} & Band 6 Red Edge & 0.705--0.745 &  &  &  \\ \cline{3-4}
\multicolumn{2}{|l|}{} & Bnad 7 NIR 1 & 0.770--0.895 &  &  &  \\ \cline{3-4}
\multicolumn{2}{|l|}{} & Band 8 NIR 2 & 0.860--1.040 &  &  &  \\ \hline
\multicolumn{2}{|l|}{\multirow{27}{*}{WorldView-3}} & Panchromatic & 0.450--0.800 & 0.31 & \multirow{27}{*}{13.1} & \multirow{27}{*}{2014} \\ \cline{2-5}
 & \multirow{8}{*}{VNIR} & Coastal Blue & 0.400--0.450 & \multirow{8}{*}{1.24} &  &  \\ \cline{3-4}
 &  & Blue & 0.450--0.510 &  &  &  \\ \cline{3-4}
 &  & Green & 0.510--0.580 &  &  &  \\ \cline{3-4}
 &  & Yellow & 0.585--0.625 &  &  &  \\ \cline{3-4}
 &  & Red & 0.630--0.690 &  &  &  \\ \cline{3-4}
 &  & Red Edge & 0.705--0.745 &  &  &  \\ \cline{3-4}
 &  & NIR 1 & 0.770--0.895 &  &  &  \\ \cline{3-4}
 &  & NIR 2 & 0.860--1.040 &  &  &  \\ \cline{2-5}
 & \multirow{8}{*}{SWIR} & Band 1 & 1.195--1.225 & \multirow{8}{*}{3.7} &  &  \\ \cline{3-4}
 &  & Band 2 & 1.550--1.590 &  &  &  \\ \cline{3-4}
 &  & Band 3 & 1.640--1.680 &  &  &  \\ \cline{3-4}
 &  & Band 4 & 1.710--1.750 &  &  &  \\ \cline{3-4}
 &  & Band 5 & 2.145--2.185 &  &  &  \\ \cline{3-4}
 &  & Band 6 & 2.185--2.225 &  &  &  \\ \cline{3-4}
 &  & Band 7 & 2.235--2.285 &  &  &  \\ \cline{3-4}
 &  & Band 8 & 2.295--2.365 &  &  &  \\ \cline{2-5}
 & \multirow{11}{*}{CAVIS} & Desert Clouds & 0.405--0.420 & \multirow{11}{*}{30} &  &  \\ \cline{3-4}
 &  & Aerosols 1 & 0.459--0.509 &  &  &  \\ \cline{3-4}
 &  & Green & 0.525--0.585 &  &  &  \\ \cline{3-4}
 &  & Aerosols 2 & 0.620--0.670 &  &  &  \\ \cline{3-4}
 &  & Water 1 & 0.845--0.885 &  &  &  \\ \cline{3-4}
 &  & Water 2 & 0.897--0.927 &  &  &  \\ \cline{3-4}
 &  & Water 3 & 0.930--0.965 &  &  &  \\ \cline{3-4}
 &  & NDVI-SWIR & 1.220--1.252 &  &  &  \\ \cline{3-4}
 &  & Cirrus & 1.350--1.410 &  &  &  \\ \cline{3-4}
 &  & Snow & 1.620--1.680 &  &  &  \\ \cline{3-4}
 &  & Aerosol 3 & 2.105--2.245 &  &  &  \\ \hline
\end{tabular}%
}
\end{table*}

\begin{table*}[htbp!]
\centering
{\footnotesize Continued from the previous page.}
\label{table1_3}
\resizebox{\textwidth}{!}{%
\begin{tabular}{|l|l|l|l|l|l|l|}
\hline
Satellite/Sensor & Subsystem & Band Number/Name & \makecell{Spectral Range ($\mu$m)/\\Frequency} & Ground Resolution (m) & Swath Width (km) & \makecell{Year of Launch/\\First Flight or Operation} \\ \hline
\multicolumn{2}{|l|}{\multirow{12}{*}{Sentinel-2}} & Band 1 Coastal Aerosol & 0.433--0.453 & 60 & \multirow{12}{*}{290} & \multirow{12}{*}{2015} \\ \cline{3-5}
\multicolumn{2}{|l|}{} & Band 2 Blue & 0.458--0.523 & \multirow{3}{*}{10} &  &  \\ \cline{3-4}
\multicolumn{2}{|l|}{} & Band 3 Green & 0.543--0.578 &  &  &  \\ \cline{3-4}
\multicolumn{2}{|l|}{} & Band 4 Red & 0.650--0.680 &  &  &  \\ \cline{3-5}
\multicolumn{2}{|l|}{} & Band 5 Red Edge 1 & 0.698--0.713 & \multirow{3}{*}{20} &  &  \\ \cline{3-4}
\multicolumn{2}{|l|}{} & Band 6 Red Edge 2 & 0.733--0.748 &  &  &  \\ \cline{3-4}
\multicolumn{2}{|l|}{} & Band 7 Red Edge 3 & 0.773--0.793 &  &  &  \\ \cline{3-5}
\multicolumn{2}{|l|}{} & Band 8 NIR & 0.785--0.900 & 10 &  &  \\ \cline{3-5}
\multicolumn{2}{|l|}{} & Band 8A Narrow NIR & 0.855--0.875 & 20 &  &  \\ \cline{3-5}
\multicolumn{2}{|l|}{} & Band 9 Water Vapor & 0.935--0.955 & \multirow{2}{*}{60} &  &  \\ \cline{3-4}
\multicolumn{2}{|l|}{} & Band 10 SWIR/Cirrus & 1.360--1.390 &  &  &  \\ \cline{3-5}
\multicolumn{2}{|l|}{} & Band 11 SWIR 1 & 1.565--1.655 & \multirow{2}{*}{20} &  &  \\ \cline{3-4}
\multicolumn{2}{|l|}{} & Band 12 SWIR 2 & 2.100--2.280 &  &  &  \\ \hline
\multicolumn{2}{|l|}{\multirow{5}{*}{RapidEye}} & Blue & 0.440--0.510 & \multirow{5}{*}{5} & \multirow{5}{*}{77} & \multirow{5}{*}{2008} \\ \cline{3-4}
\multicolumn{2}{|l|}{} & Green & 0.520--0.590 &  &  &  \\ \cline{3-4}
\multicolumn{2}{|l|}{} & Red & 0.630--0.685 &  &  &  \\ \cline{3-4}
\multicolumn{2}{|l|}{} & Red Edge & 0.690--0.730 &  &  &  \\ \cline{3-4}
\multicolumn{2}{|l|}{} & NIR & 0.760--0.850 &  &  &  \\ \hline
\multicolumn{2}{|l|}{\multirow{5}{*}{SPOT 5}} & Panchromatic & 0.48--0.71 & 5 & \multirow{5}{*}{60} & \multirow{5}{*}{2002} \\ \cline{3-5}
\multicolumn{2}{|l|}{} & Band 1 Green & 0.50--0.59 & \multirow{3}{*}{10} &  &  \\ \cline{3-4}
\multicolumn{2}{|l|}{} & Band 2 Red & 0.61--0.68 &  &  &  \\ \cline{3-4}
\multicolumn{2}{|l|}{} & Band 3 NIR & 0.78--0.89 &  &  &  \\ \cline{3-5}
\multicolumn{2}{|l|}{} & Band 4 SWIR & 1.58--1.75 & 20 &  &  \\ \hline
\multicolumn{2}{|l|}{\multirow{5}{*}{SPOT 7}} & Panchromatic & 0.45--0.75 & 1.5 & \multirow{5}{*}{60} & \multirow{5}{*}{2014} \\ \cline{3-5}
\multicolumn{2}{|l|}{} & Band 1 Blue & 0.45--0.52 & \multirow{4}{*}{6} &  &  \\ \cline{3-4}
\multicolumn{2}{|l|}{} & Band 2 Green & 0.53--0.60 &  &  &  \\ \cline{3-4}
\multicolumn{2}{|l|}{} & Band 3 Red & 0.62--0.69 &  &  &  \\ \cline{3-4}
\multicolumn{2}{|l|}{} & Band 4 NIR & 0.76--0.89 &  &  &  \\ \hline
\multicolumn{2}{|l|}{Hyperion} & 242 spectral bands & 0.357--2.576 & 30 & 7.7$\times$42 & 2000 \\ \hline
\multicolumn{2}{|l|}{HyMap} & 126 spectral bands & 0.45–2.50 & $\sim$5 & - & - \\ \hline
\multicolumn{2}{|l|}{AVIRIS} & 224 spectral bands & 0.36--2.50 & $\sim$20 & $\sim$10.5 & 1987 \\ \hline
\multicolumn{2}{|l|}{AVIRIS-NG} & 425 spectral bands & 0.38--2.51 & $\sim$5 & 4--6 & 2012 \\ \hline
\multicolumn{2}{|l|}{Geoscan Airborne Multispectral Scanner (AMSS)} & 46 spectral bands & 0.49--12 & $\sim$10 & - & - \\ \hline
\multicolumn{2}{|l|}{UHD 285 Hedgehog Camera} & 285 spectral bands & 0.45--0.95 & - & - & 2013 \\ \hline
\multicolumn{2}{|l|}{NEO HySpex VNIR-1600} & 160 spectral bands & 0.41--0.99 & - & - & - \\ \hline
RADARSAT-1 & SAR & C-band & 5.3 GHz & 10--100 & $\sim$100 & 1995 \\ \hline
RADARSAT-2 & SAR & C-band & 5.405 GHz & 3--100 & $\sim$50 & 2007 \\ \hline
\multirow{4}{*}{Sentinel-1} & Stripmap & C-band & \multirow{4}{*}{5.405 GHz} & 5 & 80 & \multirow{4}{*}{2014} \\ \cline{2-3} \cline{5-6}
 & Interferometric wide swath & C-band &  & 5$\times$20 & 250 &  \\ \cline{2-3} \cline{5-6}
 & Extra wide swath & C-band &  & 20$\times$40 & 400 &  \\ \cline{2-3} \cline{5-6}
 & Wave & C-band &  & 5 & 20 &  \\ \hline
\multicolumn{2}{|l|}{ALOS PALSAR} & L-band & 1.3 GHz & 10--100 & 30--350 & 2006 \\ \hline
\end{tabular}%
}
\end{table*}

\subsection{Satellite data}
\label{sec3-1}
In general, satellite datasets are acquired by passive and active remote sensing systems differentiated by the source of energy used to collect data. Passive systems rely on ambient energy from an external source, mostly the sun on Earth, while active systems create their own energy. Optical and radar (radio detection and ranging) data, that are  categorized as passive and active remote sensing data, respectively, constitute the most important data types used for geological mapping. Optical sensors measure the intensity of the electromagnetic spectrum over a small range of wavelengths, and the resulting electronic signal is referred to as a channel when processed \citep{Lee2020}. On the other hand, a radar remote sensing system works in the microwave portion of the electromagnetic spectrum, defined as wavelengths between 1 millimeter (mm) and 1 meter (m) \citep{Zhou2011}. Next, we review both data types in mapping important features for detecting potential mineralization zones.

\subsubsection{Optical data}
\label{sec3-1-1}
Landsat satellites are the most famous satellites that provide optical data and have been widely used in geological mapping. They have continuously monitored Earth's surface to satisfy various information and data requirements for more than four decades \citep{Wulder2008}. Among these satellites, Landsat 5, Landsat 7, and Landsat 8 have been more of interest to exploration geologists for mapping different geological features in recent years. Landsat 5 was launched in 1984, and in addition to the multispectral scanner (MSS), it carried the thematic mapper (TM) sensor. The TM sensor gathered data in seven bands in the visible, shortwave infrared, and thermal regions. The TM sensor's spatial resolution is 120 m for the thermal band and 30 m for the other bands \citep{Banskota2014}.\\
Landsat 7 was launched on 15 April 1999 and carries the ETM+ sensor. It acquires data in eight spectral bands with different spatial resolutions, including visible and near-infrared (VNIR) bands 1--4 and shortwave infrared (SWIR) bands 5 and 7 with a spatial resolution of 30 m and panchromatic band 8 with a spatial resolution of 15 m. The thermal infrared band 6 provides a spatial resolution of 60 m \citep{RajanGirija2019}. Geologists benefit from using shortwave infrared bands because they are sensitive to soil and rock content changes, making it possible to differentiate some basic rock types. Landsat 8 was launched on 11 February 2013 and carries two sensors, including OLI and thermal infrared sensor (TIRS). It offers images in 11 spectral bands with the same resolution as that of ETM+ in the VNIR and SWIR bands 1--7 and the panchromatic band 8. The spatial resolution of band nine that is used for cirrus cloud detection is 30 m. The last two thermal bands of 10 and 11 have a 100 m resolution. The spectral ranges of OLI bands are designed to prevent the atmospheric absorption properties within the ETM+ bands \citep{Zhang2016}. In the band 4 (0.780--0.900 $\mu$m) of the ETM+ sensor, the water vapor absorption characteristics exist at 0.825 $\mu$m and they are eliminated in OLI by adding band 4 (0.630--0.680 $\mu$m) and band 5 (0.850--0.880 $\mu$m) \citep{Zhang2016}.\\
ASTER is part of the Earth observation system (EOS) Terra platform and tracks solar radiation in 14 spectral bands. ASTER measures reflected radiation in three bands ranging from 0.52 to 0.86 $\mu$m (VNIR) and six bands ranging from 1.6 to 2.43 $\mu$m (SWIR) with a resolution of 15 and 30 m, respectively. ASTER has five bands in the range of thermal infrared (TIR) from 8.125 to 11.65 $\mu$m wavelengths. Each scene in ASTER covers an area of 60$\times$60 km \citep{Rowan2003}. The European space agency has been developing a new family of missions called Sentinels specifically for the operational needs of the Copernicus program. Each Sentinel mission is based on a constellation of two satellites to fulfill revisit and coverage requirements, providing robust datasets for Copernicus services. These missions carry a range of technologies, such as radar and multi-spectral imaging instruments for land, ocean, and atmospheric monitoring. Sentinel-2 is a polar-orbiting, multispectral high-resolution imaging mission for land monitoring. Sentinel-2A was launched on 23 June 2015, and Sentinel-2B followed on 7 March 2017. They provide images that can be used for detecting mineral occurrences on the Earth's surface and consist of 13 VNIR and SWIR spectral bands with a spatial resolution of 10 m for four bands, 20 m for six bands, and 60 m for three bands \citep{Drusch2012}.\\
High-resolution images have to be usually purchased from commercial vendors such as DigitalGlobe (Maxar Technologies) \footnote{https://www.maxar.com}, Planet Labs or Planet \footnote{https://www.planet.com}, and Spot Image (Airbus Defence and Space) \footnote{https://www.intelligence-airbusds.com}. DigitalGlobe is an American commercial vendor of space imagery and geospatial content and operator of civilian remote sensing spacecrafts, such as IKONOS, QuickBird, GeoEye-1, and WorldView satellite system that are a network of commercial and orbital platforms designed by Ball Aerospace and Technologies \citep{Good2018}. In 2007, WorldView-1 (WV-1) was launched with a 50-centimeter (cm) spatial resolution panchromatic (PAN) imaging system. The primary objective of the single-band PAN system with no multispectral bands on board was to quickly collect high spatial resolution imagery, best suited for generating detailed data from the digital elevation model (DEM). The next major breakthrough was the launch of WorldView-2 (WV-2) in 2009, which provided high-resolution PAN data at a pixel size of 46 cm plus VNIR bands at a spatial resolution of 1.85 m. The first instrument for collecting eight high-resolution multispectral bands ranging from 0.4 to 1.04 $\mu$m wavelengths was WV-2 \citep{Kruse2013}. The only 16-band commercial high-resolution Earth-imaging satellite currently in space is WorldView-3 (WV-3), launched in August 2014. WV-3 has the enhanced capability of eight SWIR (1.2--2.33 $\mu$m) bands with a spatial resolution of 3.7 m, in addition to eight VNIR (0.42--1.04 $\mu$m) bands at a spatial resolution of 1.2 m \citep{Kruse2015}.\\
The Planet is an American private Earth-imaging company with three distinct constellations of satellites, including Doves, SkySats, and RapidEye. Among them, RapidEye data, in combination with other satellite data types, have been used for mapping geological features, such as pegmatite deposits \citep{Peng2013}. Spot Image is headquartered in France and is mostly known as the commercial operator for SPOT (Satellite Pour l'Observation de la Terre) Earth observation satellites. This company also distributes multi-resolution data from other optical and radar satellites, such as very high-resolution Pleiades satellites. SPOT is a commercial high-resolution optical Earth-imaging satellite system operating from space. It has been designed to improve the knowledge and management of the Earth by exploring the Earth's resources, detecting and forecasting phenomena involving climatology and oceanography, and monitoring human activities and natural phenomena. The SPOT system includes a series of satellites (SPOT 1--7) and ground control resources for satellite control and programming, image production, and distribution. Among different SPOT satellites, the combination of SPOT 5 and other multispectral data has been widely used for mapping geological features \citep{Harbi2014,Ahmadirouhani2018,Bishta2018,Bishta2021}.\\
Google Earth made it possible to view, map, and navigate any remote locations on the Earth's surface free of cost \citep{Bailey2012,Fisher2012} and is considered as one of the most efficient tools for geoscience. Google Earth comprises a wide range of true-color visible spectrum satellite imagery obtained from different Landsat imagery and high-resolution data available from commercial vendors \citep{tewksbury2012,Fisher2012}. It is of enormous use to identify main outcrop positions in pre-field planning and reconnaissance surveys and connect remote areas with outcrops confirmed by field data. In addition, it is a beneficial instrument in remote and war-torn or politically troubled areas.\\
The invention of the Hyperion sensor marked the beginning of hyperspectral remote sensing. Hyperion is the first space-borne hyperspectral sensor, which is capable of providing data in the spectrum of VNIR and SWIR, launched in November 2000 as part of NASA's EO-1 Millennium Mission \citep{Pearlman2003}. It comprises the 0.36--2.58 $\mu$m spectrum of VNIR and SWIR regions with 242 spectral bands at roughly 10 nanometers (nm) spectral resolution and 30 m spatial resolution \citep{Pearlman2003}.

\subsubsection{Radar data}
\label{sec3-1-2}
The Canadian government approved an Earth observation program named RADARSAT (1980). Since the launch of RADARSAT-1 (1995) and the release of RADARSAT-2 (2007), Canada has been supplying C-band synthetic aperture radar (SAR) data without interruption. With the latest planning of the next generation mission, the RADARSAT constellation, there is also a solid commitment to ensuring data reliability in the future. This perennial data supply allows users to incorporate this important pool of knowledge into their operating applications at the national and international levels \citep{Iris2019}. Sentinel-1 is a polar-orbiting, all-weather, day-and-night radar imaging mission for land and ocean services. Sentinel-1A was launched on 3 April 2014 and Sentinel-1B on 25 April 2016. Sentinel-1 C-band SAR sensors have dual-polarization (co-polarized VV or HH, and cross-polarized VH or HV), wide-swath interferometric mode, and spatial resolution of 5$\times$20 m \citep{Zoheir2019}. SAR data from microwaves is an excellent component of professional data used to map geological structures. The phased array type L-band synthetic aperture radar (PALSAR) sensor is a fully polarized (HH, HV, VH, and VV) L-band SAR sensor and multi-observation modes (fine, polarimetric, and ScanSar) with a spatial resolution of 10, 30, and 100 m, respectively \citep{Ma2017}.

\subsection{Airborne data}
\label{sec3-2}
Nowadays, airborne data are collected using both airplanes and drones, and specific sensors have been developed for them. Multispectral and hyperspectral airborne data have been available from the Geoscan AMSS MKI and MKII sensors since 1997 \citep{Agar1994}. The design of advanced visible infrared imaging spectrometer-new generation (AVIRIS-NG) was developed based on AVIRIS with some modifications and improvements. AVIRIS measures solar irradiance with a whisk-broom scanning mechanism with 224 bands of 10 nm spectral resolution across-track components. The AVIRIS-NG data has a high signal-to-noise ratio and is free of smile/spikes and keystone errors. AVIRIS-NG is a hyperspectral imaging spectrometer with a spatial resolution of 8.1 m with 427 contiguously spaced 5 nm bands from 0.38 to 2.51 $\mu$m \citep{Hamlin2011}. In October 1999, the hyperspectral mapper (HyMap) was developed and operated in Australia. This sensor covers the wavelength of 0.45 to 2.48 $\mu$m and consists of 126 spectral bands, and provides a spatial resolution of 2 to 10 m. With the exception of absorption levels near 1.4 and 1.9 $\mu$m, due to atmospheric water vapor, HyMap can achieve a continuous continuum \citep{Ishidoshiro2016}. The airborne sensors such as Geoscan AMSS MKI, GER DAIS 63, AVIRIS, De Beers AMS, TEEMS, HyMap, CASI, SFSI, and SpecTir have been reviewed with details of spectral and spatial characteristics \citep{Agar2007}.\\
Unmanned aerial systems (UAS) are known under several different names and acronyms such as unmanned aerial vehicle (UAV), aerial robot, or simply drone, with the most common words being UAV and drone \citep{Colomina2014}. Drones have been progressively used in mineral exploration. Rugged terrain or outcrops difficult to reach on foot or by cars may be quickly surveyed with UAVs from a safe distance with limited human resources on-site, ensuring protection, speed, and quality. The importance of using UAV-based hyperspectral images for geological exploration mapping has been shown in limited studies (e.g., \cite{Booysen2020}). Hyperspectral sensors have recently begun to be installed on UAVs by the Helmholtz Institute Freiberg for Resource Technologies \citep{Jakob2017,Jackisch2020}. It is possible to obtain higher spatial resolutions (from millimeters to tens of centimeters) and a number of scanning viewpoints are possible with hyperspectral sensors. \cite{Colomina2014} described the fundamental characteristics of some common and representative sensors, including visible-band, near-infrared (NIR), thermal, multispectral and hyperspectral cameras, laser scanners, and synthetic aperture radars for UAVs. \cite{Heincke2019} developed multi-sensor drones for geological mapping and mineral exploration with a new integrated positioning system that is not based on global positioning system (GPS). This allows precise positioning across areas with low GPS reception, such as mine tunnels and narrow valleys.

\subsection{Ground-based data}
\label{sec3-3}
Ground-based hyperspectral sensing, especially in the VNIR and SWIR portions of the electromagnetic spectrum, has grown more popular for geological purposes. Remote measurements with ever greater spectral and spatial resolutions have been achieved, although proximal sensing with high spectral resolution (10 nm) for quick characterization of rocks, minerals, and soil is widespread \citep{Salazar2020}. Hyperspectral images captured by specialised cameras can be used for mapping alteration minerals. In the Gongchangling iron deposit, a typical profile of a high-grade iron ore body has been examined and analyzed \citep{Song2020}. Polarized microscopy has been used to examine the materials, and changes in zonation were determined based on hydrothermal mineral assemblages and paragenesis. Furthermore, the Norsk Elektro Optikk (NEO) HySpex imagining system was used to produce hyperspectral pictures of wall rocks from each alteration zone. Their findings show that spectral characteristics reveal obvious regular variations; for example, as the high-grade iron ore body moves from proximal to distal, the wavelengths of chlorite and garnet account for the majority of the hydrothermal alteration minerals grow longer and the absorption depths shrink.\\
A hyperspectral scanner can be mounted on a platform, for example, about 4 m above ground \citep{Krupnik2016}. For instance, HySpex can be used for scanning field sceneries, which may be readily put on a tripod and rotation stage. A two-dimensional charge-coupled device (CCD) sensor array is used in this camera; the first dimension is utilized for spectrum separation, while the second is employed for imaging in a single spatial direction. The sensor's movement covers the second spatial dimension. The camera captures 1600 pixels on each track, resulting in a less than a millimeter spatial resolution. The rotating stage's speed can be adjusted such that the single lines captured by the camera create a picture with nearly square pixels. One hundred sixty bands can be recorded in the spectrum range of 410 to 990 nm with a spectral sampling distance of 3.7 nm by binning the 320 sensor pixels in the spectral direction. The data can be captured with a radiometric resolution of 12 bits \citep{Buddenbaum2012}.\\
The hyperspectral snapshot camera approach enables quick picture data capture in a portable manner. The UHD 285 hyperspectral snapshot camera is a non-scanning hyperspectral camera primarily intended for real-time data capture. The full-frame photos are captured with a dynamic picture resolution of 14 bits on a silicon CCD chip with a sensor resolution of 970$\times$970 pixels in the 450 to 950 nm region. The integration time of capturing one hyperspectral data cube in a regular solar light environment is 1 millisecond (ms). The camera can acquire over 15 spectral data cubes per second, making hyperspectral video recording possible \citep{Jung2015}.

\section{Target features}
\label{sec4}
In one of the first studies on the application of image processing techniques for mineral exploration, \cite{Sabins1999} proposed two key approaches for targeting mineral deposits. The first is lithological and structural mapping, and the second is mapping hydrothermal alteration zones. Later, three methods were suggested by \cite{Rajesh2004}: (i) lithological; (ii) structural; and (iii) alteration mapping. Identifying mineralization zones as the fourth approach is also important for vectoring different deposits \citep{VanderMeer2012}. Since the 1970s, image processing algorithms have been implemented for detecting different types of mineral deposits, particularly porphyry copper and gold deposits. Schematic cross-sections showing the association of alteration and mineralization zones in a porphyry copper system is illustrated in Figure 2. Moreover, they have been used for exploring iron ore, volcanogenic massive sulfide (VMS), skarn-hosted, chromite, rare earth elements (REE), brine and evaporite, porphyry molybdenum, zinc, and lead, diamond, and bauxite deposits \citep{Cardoso-Fernandes2020}. Next, we discuss the most important target features in mineral exploration.

\begin{figure}
  \centering
  \frame{\includegraphics[width=\linewidth]{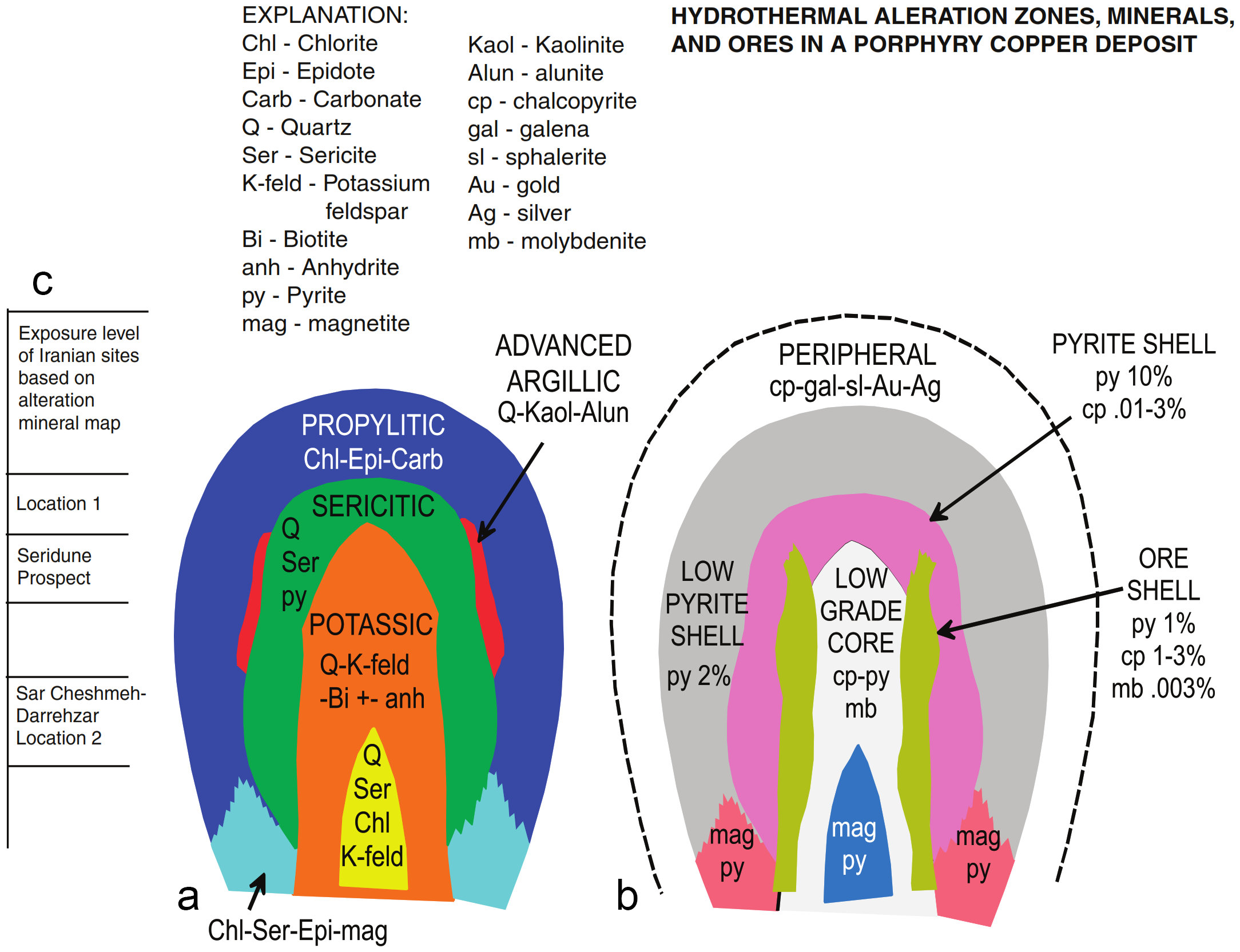}}
  \caption{An illustrated deposit model of a porphyry copper deposit (modified from \cite{Lowell1970}). Panel (a) presents a schematic cross-section of hydrothermal alteration minerals and types, which includes propylitic, sericitic, advanced argillic, and potassic alteration. Panel (b) presents a schematic cross-section of ores associated with each alteration type. Panel (c) presents a scale showing the level of interpreted exposure for Iranian alteration sites based on ASTER mapped alteration units.}
  \label{fig2}
\end{figure}

\subsection{Lithology}
\label{sec4-1}
Ore deposits are volumes of rock containing desired elements in appropriate amounts and quantities that can be economically exploited. Any ore-forming hydrothermal transport system includes massive amounts of rock (or magma) through which ore-forming elements are released. For example, VMS deposits are high-grade metal accumulations associated with submarine magmatic rocks \citep{Heinrich2014}. Mineral deposits in sedimentary rocks are the major resource of lead and zinc that occur as either Mississippi valley type or stratiform clastic sediment-hosted deposits. These deposits consist of a variety of minerals hosted by a wide range of carbonate and siliciclastic rocks \citep{Leach2005}. The major resources of Cu, Mo, Sn, W, In, and Re and a significant source of Au, Ag, Pb, Zn, and other minor and rare metals are associated with intrusive rocks magmatic-hydrothermal ore systems. Porphyry and epithermal Cu, Mo, and Au deposits are some examples of these deposits \citep{Richards2011}. Therefore, discriminating and identifying different rock units are inseparable and fundamental parts of the mineral exploration process, and remote sensing data can provide extraction of critical geological features.\\
Several image processing algorithms have been examined in poorly mapped or unmapped areas to represent and optimize the distinction of various spectral lithological divisions in order to differentiate lithological units utilizing various types of satellite data \citep{Pour2019}. Dimensionality reduction techniques such as principal component analysis (PCA), independent component analysis (ICA), and minimum noise fraction (MNF), have been applied on OLI and ASTER spectral bands to map lithological units \citep{Pour2019,Ali2014}. Classification methods such as spectral angle mapper (SAM) and maximum likelihood estimation (MLE) have also been applied on Hyperion, ALI, and ASTER data for mapping different lithological units \citep{Pour2014}. \cite{Amer2010} used the images obtained by PCA and a newly evolved ASTER band ratio to produce lithological maps in the Central Eastern Desert of Egypt. Within this study, different ophiolitic rocks, including serpentinite, metagabbro, metabasalt, and granitic rocks such as grey and pink granites, were discriminated against for vectoring chromite deposits. \cite{Yu2012} used ASTER imagery along with ASTER-derived DEM and aeromagnetic data to apply the support vector machine (SVM) method for automated lithological classification of a study area in northwestern India. In this study, the SVM was compared with the maximum likelihood classifier (MLC). Results revealed that the SVM had better precision in the classification of independent validation samples and similarities to the available lithological bedrock database. In recent years, modern sensors such as WV-3 that provide high-resolution images have been evaluated for lithological mapping, and results have been compared to ASTER and OLI data to satisfy the criteria for large-scale geological mapping \cite{Ye2017}.

\subsection{Alteration types}
\label{sec4-2}
Geochemical properties of wall rocks in an ore deposition region cause alteration and eventually forming a diversity of mineral deposit types and metals such as porphyry copper and epithermal gold systems \citep{Sillitoe2010}. Porphyry deposits are characterized as large volumes (10--$>$100 $km^3$) of hydrothermally altered rocks clustered on porphyry stocks that may also include precious metallic mineralization of skarn, carbonate replacement, sediment-hosted, and epithermal types \citep{Richards2003}. Argillic, phyllic, propylitic, and silicic alteration patterns are some of the important indicators for vectoring porphyry and epithermal deposits \citep{Testa2018}. In the case of sediment-hosted deposits, dissolution and hydrothermal brecciation of carbonate host rocks that occur from acid-producing reactions typically associated with fluid mixing is the most frequent alteration observed in MVT deposits. The alteration and halo production style associated with SEDEX deposits relies on the host sedimentary structure and permeability plus porosity characteristics. Since the alteration amplitude is usually much lower in SEDEX systems than in VMS systems, the scale of halos across favorable formations can be much greater. Iron-manganese carbonate and silicate alteration are two major alteration types that can be mapped by the detailed analysis of target deposits \cite{Leach2005}. Therefore, distinguishing altered rocks and identifying alteration patterns are so critical for delineating the favorable area of mineralization; hence, remote sensing data analysis is an effective tool.\\
The ability to discriminate between altered and unaltered zones is essential in geological mapping for different purposes, such as mineral exploration. Due to various mineral assemblages, each alteration type indicates a particular spectral pattern. Geologists use these spectral characteristics as diagnostic features to identify and distinguish between various alteration types using remote sensing data. Multispectral and hyperspectral remote sensing instruments provide detailed spectral data on the geochemistry of rocks comprising the Earth's crust. This technology has been used for decades to map weathering characteristics in different regions, e.g., \cite{VanderMeer2012,Pour2019,Pour2019a,Bolouki2020}. PCA \citep{Ghulam2010}, SAM \citep{Ferrier2002}, spectral information divergence (SID) \citep{Sheikhrahimi2019}, and the integration of selective dimensionality reduction techniques \citep{Shirmard2020} are some of the methods that have been applied on remote sensing data to map hydrothermal alteration zones. Different alteration types such as propylitic, phyllic, argillic, and advanced argillic have been discriminated by applying various methods such as selective PCA on ASTER data \citep{Noori2019}. Moreover, AVIRIS-NG hyperspectral data with 5 nm spectral resolution allowed the identification of various altered, and weathered clay groups in target areas \citep{Tripathi2020}.

\subsection{Structures}
\label{sec4-3}
Several types of mineral deposits such as epithermal, mesothermal, carlin-type gold, and other hydrothermal deposits are usually associated with fault, vein, and shear zone systems. Such structural features can be useful to future exploration efforts in regions associated with structurally controlled minerals \citep{Grebby2012}. The structural analysis applied to mineral exploration aims to identify how deformation has impacted the permeability in rocks, either spatially or overtime \citep{Micklethwaite2010}. The extraction of tectonic lineaments using satellite data is a basic application of remote sensing data analysis. For a number of applications, mapping tectonic lineaments such as faults and dykes are of high importance, mainly due to their relationship with hydrothermal mineralization \citep{Farahbakhsh2020a}.\\
Recent advances in remote sensing technology have improved the application of optical and radar remote sensing data for investigating geological structures such as tectonic lineaments that involve rectilinear and curvilinear structures in covered and uncovered areas. Structural lineament interpretation helps to understand the tectonic and geodynamic processes of an area \citep{Chinkaka2019}. A variety of algorithms and remote sensing data formats have been used to map geological structures. For instance, the spatial convolution filtering technique has been widely used for processing PALSAR data \citep{Pour2016}. Lineaments have been automatically extracted by applying edge detection methods on shuttle radar topography mission (SRTM), OLI, and ASTER data \citep{Hamimi2020}. Various methods such as spectral band ratio indices, supervised classification techniques namely SAM, SID, directional filtering technique, and manually extracting tectonic lineaments have been applied on ASTER, SAR/RADARSAT-1, ASAR/ENVISAT, SPOT 5, and SPOT 7 data for mapping geological structures \citep{Pour2018,Sheikhrahimi2019,Tagnon2020,Ibrahim2017}. Image processing methods can be applied on radar remote sensing data for evaluating the regional structural control of scattered gold anomalies \citep{Zoheir2019}. SAR data with the help of remote sensing data processing, enhanced a vivid litho-tectonic insight for the South Eastern Desert in Egypt \citep{Zoheir2019}.

\subsection{Minerals}
\label{sec4-4}
It is crucial to differentiate unique minerals as an indicator of high economic potential zones. Identifying mineralization zones using remote sensing data has been widely done for locating porphyry copper, epithermal gold, and VMS deposits around the world \citep{Bolouki2020}. Quantitative and validated (subpixel) surface mineralogic mapping was initiated with the advent of hyperspectral remote sensing data \citep{RajanGirija2019}. This led to a variety of techniques for matching image pixel spectra to library and field spectra and unraveling mixed pixel spectra to pure end-member spectra for extracting compositional details from a subpixel surface \citep{VanderMeer2012}. For example, the footprint of minerals such as clay minerals (e.g., kaolinite and illite), sulfate minerals (e.g., alunite), carbonate minerals (e.g., calcite and dolomite), iron oxides (e.g., hematite and goethite), and silica (quartz) enabled to map alteration facies (propylitic, argillic, etc.), which are the key indicators for targeting epithermal and porphyry-related deposits \citep{Testa2018,RajanGirija2019,VanderMeer2012}.\\
Ultramafic igneous rocks can be classified into those where oxide minerals host the metals of interest and those where the metals of interest are maintained as sulfides or are closely correlated with sulfides. Stratiform, podiform, and breccia-related chromite, magnetite-rich layers (often Ti and V-bearing), and ilmenite-rich layers or discordant bodies are found in oxide ore deposits. Massive, net-textured, and disseminated Ni-Cu-PGE (platinum group elements) occurrences, and PGE-rich reefs containing disseminated sulfides have ore minerals \citep{Ripley2018}. Therefore, remote sensing data can be used extensively and effectively for mapping these minerals \citep{Awad2018}. There are many methods used for extracting indicator minerals from remote sensing data. PCA and ICA are two of the commonly-used image processing algorithms in the mineral mapping \citep{Cardoso-Fernandes2020,Farahbakhsh2016}. Several spectral analyses have been implemented on short-wave infrared bands of ASTER for detecting spectral features of interest attributed to alteration mineral assemblages at different scales \citep{Pour2019}. ASTER, ALI, and Hyperion are some of the most common remote sensing data used for mapping minerals \citep{Pour2014}. \cite{Kruse2013} extracted minerals as regions of interest from airborne visible/infrared imaging spectrometer (AVIRIS), ASTER, and World View-3 data by applying minimum distance classification. In this study, calcite, buddingtonite, alunite, kaolinite, muscovite, and silica can be named as some of the minerals extracted and mapped by mentioned remote sensing data. Moreover, image processing methods have been applied to lightweight drone-based hyperspectral data for direct mapping of REEs in Marinkas Quellen, Namibia, and Siilinj\"{a}rvi, Finland \citep{Booysen2020}.

\section{Machine learning}
\label{sec5}
Mapping geological features is a fundamental step in mineral exploration. The combined use of machine learning methods and remote sensing data can be considered an easy and inexpensive approach for mapping lithological units, alteration zones, structures, and indicator minerals associated with mineral deposits. In several fields, rapid advancements in acquiring high-resolution remote sensing data have led to the explosion of big data that offers a new opportunity for data-driven discovery \citep{Sun2019}. Machine learning methods are effective for remote sensing data analysis since they can automatically learn the relationship between input features such as reflectance continuum with desired outputs for prediction or classification. Moreover, they are robust in spectral and ground truth measurements against noise and uncertainties \citep{Gewali2018}.\\
Broadly speaking, machine learning methods are of two major types, which include supervised and unsupervised learning. Supervised machine learning methods require labeled data that are used for regression and classification problems to model the relationship between input features and outcomes \citep{Kotsiantis2007}. Machine learning-based classification methods for different datasets (problems) offer different outcomes, i.e., various classified maps. A machine learning method that provides the best accuracy for solving a problem may not work for another problem or dataset. Hence, prior to problem-solving, various methods for a given dataset must be examined \citep{Cgsar2019}.\\
Unsupervised learning methods have the ability to recognize patterns in data without the need for target labels. Examples of unsupervised learning include clustering and data reduction strategies such as PCA. Clustering methods discover structures in data using a given measure of similarity amongst data instances in order to develop clusters (groups). Clustering methods are commonly used in machine learning and image processing \citep{Xie2020}. Reducing the feature set dimensionality is important in machine learning in order to decrease the complexity of a problem, remove outliers and noise, and finally shorten the model training time. In smaller datasets, simplified models are often more robust and are less influenced by variations due to noise or outliers \citep{Caggiano2018}.\\
Deep learning is a branch of machine learning that seeks to model high-level abstractions in data by applying multiple processing layers with complex structures. Deep learning methods such as recurrent neural networks (RNNs), convolutional neural networks (CNNs), autoencoders, deep belief networks, and restricted Boltzmann machines have been successfully applied to transform different fields \citep{Benuwa2016,Schmidhuber2015,Velliangiri2019}. Ensemble methods integrate multiple machine learning models for classification or regression problems that typically outperform standalone methods \citep{Sagi2018}. Random forest is an example of an ensemble method based on the bagging paradigm and decision trees, which can be used for classification or regression problems \citep{Rokach2010}. The combination of selected machine learning methods into ensemble learning paradigms such as boosting, stacking, and bagging has been prominent \citep{Dietterich2002,Guan2014,Yang2010}.\\
Machine learning methods are essentially data-driven approaches that can be used in several ways, such as processing high-dimensional data into lower dimensions, predicting certain trends in the data, and identifying certain characteristics or components in the data. Hence, there is a great potential and opportunity for applying machine learning methods for addressing the increasingly growing size and complexity of remote sensing data \citep{Cracknell2014}. The synergy of machine learning and remote sensing data, including satellite, airborne, and ground-based data, could be helpful in mineral exploration. Next, we review the most prominent and recently used machine learning algorithms in processing remote sensing data for mineral exploration. Machine learning methods are categorized into dimensionality reduction, classification, clustering, regression, and deep learning methods. Table 2 provides a list of sample studies focused on using machine learning methods in remote sensing data analysis for mapping potential mineralization zones. Moreover, the workflow of using the combination of remote sensing data and machine learning methods for creating critical evidential maps in mineral prospectivity mapping is presented in Figure 3.

\begin{landscape}
\begin{table}[]
\centering
\caption{Sample studies focused on using machine learning methods in remote sensing data analysis for mapping potential mineralization zones.}
\resizebox{1.4\textwidth}{!}{%
\begin{tabular}{|l|l|l|l|l|l|}
\hline
\multicolumn{2}{|c|}{Machine Learning Method} & \multicolumn{1}{c|}{Platform/Sensor} & \multicolumn{1}{c|}{Target Feature} & \multicolumn{1}{c|}{Commodity} & \multicolumn{1}{c|}{Sample Study} \\ \hline
\multirow{11}{*}{Dimensionality Reduction} & \multirow{7}{*}{Principal Component Analysis} & UAV & Structures & Genreal exploration & \cite{Zhang2020} \\ \cline{3-6}
 &  & UAV & Structures & General exploration & \cite{Wang2020} \\ \cline{3-6}
 &  & Sentinel-2 & Lithological units, alteration types, minerals & Cu, Au & \cite{Abdolmaleki2020} \\ \cline{3-6}
 &  & WorldView-3, ASTER, Sentinel-2, Landsat 8 & Structures, alteration types & Pb, Zn & \cite{Sekandari2020} \\ \cline{3-6}
 &  & ASTER & Structures, alteration types & Au & \cite{Sheikhrahimi2019} \\ \cline{3-6}
 &  & ASTER & Alteration types & Au-Cu, Ag and/or Pb-Zn & \cite{Noori2019} \\ \cline{3-6}
 &  & ASTER & Structures & General exploration & \cite{ElAtillah2019} \\ \cline{2-6}
 & \multirow{2}{*}{Minimum Noise Fraction} & Landsat 8 & Lithological units, alteration types, minerals & Au & \cite{Traore2020} \\ \cline{3-6}
 &  & UAV & Lithological units, alteration types & General exploration & \cite{Lorenz2021} \\ \cline{2-6}
 & Independent Component Analysis & ASTER & Minerals & Au & \cite{TakodjouWambo2020} \\ \hline
\multirow{19}{*}{Classification} & \multirow{3}{*}{Minimum Distance} & ASTER, Sentinel-2, Landsat 8 & Minerals & General exploration & \cite{Gemusse2019} \\ \cline{3-6}
 &  & AVIRIS, WorldView-3 & Minerals & General exploration & \cite{Kruse2013} \\ \cline{3-6}
 &  & Hyperion, ASTER, Landsat 8 & Lithological units & General exploration & \cite{Pan2019} \\ \cline{2-6}
 & \multirow{6}{*}{Support Vector Machines} & ASTER & Lithological units & Cr & \cite{Othman2014} \\ \cline{3-6}
 &  & ASTER & Alteration types, minerals & Au & \cite{Xu2019} \\ \cline{3-6}
 &  & Hyperion & Alteration types, minerals & Au & \cite{Wang2010a} \\ \cline{3-6}
 &  & Sentinel-2 & Lithological units, alteration types, minerals & Cu, Au & \cite{Abdolmaleki2020} \\ \cline{3-6}
 &  & Sentinel-2 & Lithological units & Li & \cite{Cardoso-Fernandes2020a} \\ \cline{3-6}
 &  & UAV & Lithological units, alteration types & General exploration & \cite{Lorenz2021} \\ \cline{2-6}
 & \multirow{4}{*}{Simple Neural Networks} & Landsat 5 & Minerals & Au & \cite{Rigol-Sanchez2003} \\ \cline{3-6}
 &  & Hyperion, Landat 5 & Lithological units & General exploration & \cite{Leverington2010} \\ \cline{3-6}
 &  & Landsat 7 & Alteration types, minerals & Mo, Pb, Zn, Ag & \cite{Wang2010} \\ \cline{3-6}
 &  & Landsat 7 & Structures & General exploration & \cite{Borisova2014} \\ \cline{2-6}
 & \multirow{6}{*}{Random Forest} & Sentinel-2 & Lithological units & Li & \cite{Cardoso-Fernandes2019} \\ \cline{3-6}
 &  & Landsat 5 & Lithological units & Au & \cite{Kuhn2018} \\ \cline{3-6}
 &  & Landsat 7 & Lithological units & General exploration & \cite{Cracknell2014} \\ \cline{3-6}
 &  & ASTER, Sentinel-2 & Lithological units & Rare metals & \cite{Wang2020a} \\ \cline{3-6}
 &  & Sentinel-2, PALSAR & Lithological units & General exploration & \cite{Bachri2020} \\ \cline{3-6}
 &  & TOPSAR & Lithological units & General exploration & \cite{Radford2018} \\ \hline
\multirow{5}{*}{Clustering} & \multirow{3}{*}{K-means} & AVIRIS & Minerals & General exploration & \cite{Ren2019} \\ \cline{3-6}
 &  & HyMap & Lithological units, minerals & General exploration & \cite{Ishidoshiro2016} \\ \cline{3-6}
 &  & ASTER, Sentinel-2, Landsat 8, Landsat 7 & Lithological units & General exploration & \cite{ElAtillah2019} \\ \cline{2-6}
 & \multirow{2}{*}{ISODATA} & ASTER, Sentinel-2, Landsat 8, Landsat 7 & Lithological units & General exploration & \cite{ElAtillah2019} \\ \cline{3-6}
 &  & Hyperion, Landsat 8 & Lithological units, minerals & Fe & \cite{Ducart2016} \\ \hline
\multirow{5}{*}{Regression} & \multirow{2}{*}{Multi-Linear Regression} & Hypeion, Landsat 7 & Minerals & Cu, Au, Ag & \cite{Hoang2017} \\ \cline{3-6}
 &  & Landsat 5 & Alteration types, minerals & General exploration & \cite{Yetkin2004} \\ \cline{2-6}
 & \multirow{2}{*}{Multivariate Regression} & ASTER & Minerals & Fe & \cite{Mansouri2018} \\ \cline{3-6}
 &  & HyMap & Minerals & General exploration & \cite{Li2018} \\ \cline{2-6}
 & Logistic Regression & ASTER, Landsat 7 & Alteration types, structures & General exploration & \cite{Lin2020} \\ \hline
\multirow{3}{*}{Deep Learning} & \multirow{3}{*}{Convolutional Neural Network} & UAV & Lithological units & General exploration & \cite{Sang2020} \\ \cline{3-6}
 &  & Landsat 7, Landsat 5 & Lithological units & General exploration & \cite{Latifovic2018} \\ \cline{3-6}
 &  & AVIRIS & Minerals & General exploration & \cite{Zhao2020} \\ \hline
\end{tabular}%
}
\label{table2}
\end{table}
\end{landscape}

\begin{figure*}
  \centering
  \includegraphics[width=\linewidth]{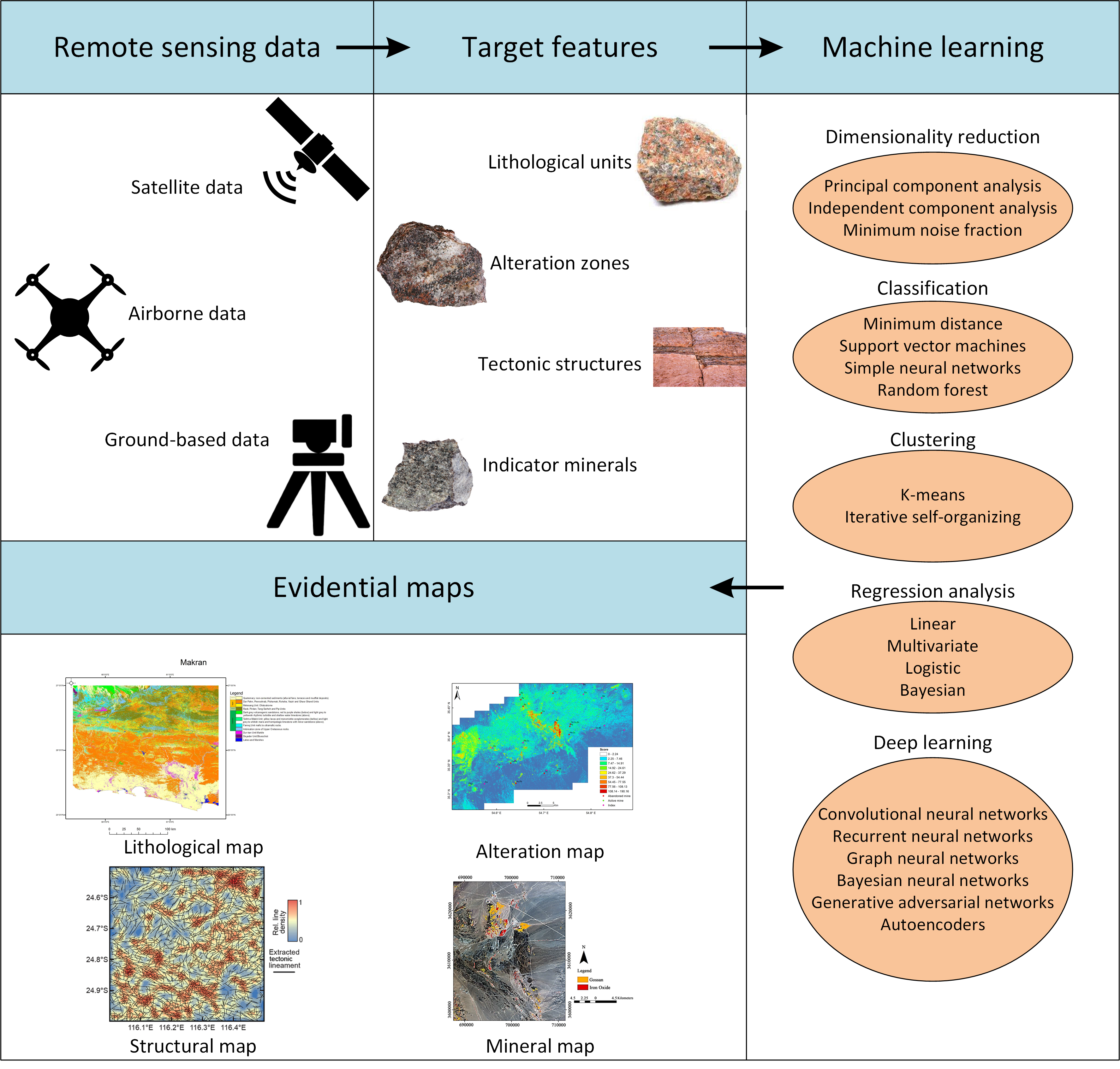}
  \caption{Workflow of using the combination of remote sensing data and machine learning methods for creating evidential maps.}
  \label{fig3}
\end{figure*}

\subsection{Dimensionality reduction techniques}
\label{sec5-1}
Dimensionality reduction techniques such as PCA \citep{Wold1987}, ICA \citep{Comon1994}, and MNF \citep{Nielsen2011} are multivariate statistical approaches that transform a collection of correlated input variables into uncorrelated or independent components and have been popular for processing remote sensing data \citep{Shirmard2020}. \cite{Traore2020} used Landsat 8 satellite images for lithological and alteration mapping by applying PCA and MNF for alluvial gold exploration. Clay and carbonate minerals, iron oxides, ferrous silicates, and lithological units were recognized and mapped in the study area. \cite{Abdolmaleki2020} used PCA for classifying and assessing Sentinel-2 data to explore iron oxide copper gold (IOCG) mineralization. \cite{TakodjouWambo2020} applied PCA and ICA on a regional scale to extract spectral details related to vegetation, iron oxide/hydroxide minerals, Al-OH and Fe-Mg-OH minerals, carbonate group, and silicification using OLI data. \citep{Sekandari2020} adopted and implemented PCA to prospect for Zn-Pb mineralization in Kerman, Iran using different remote sensing data, which includes Landsat 8, Sentinel-2, ASTER, and WV-3.\\
\cite{Sheikhrahimi2019} used ASTER data with PCA for mapping hydrothermal alteration minerals and to better discriminate structural features associated with orogenic gold occurrences in the Sanandaj Sirjan zone, Iran. In this region, PCA was used for image transformation in order to delineate lithological units and alteration minerals, which provided a fast and cost-efficient means to start a comprehensive geological exploration program. Propylitic, phyllic, argillic, and advanced argillic alteration and silicification zones are typically associated with Au-Cu, Ag, and/or Pb-Zn mineralization in the Toroud–Chahshirin magmatic belt, north of Iran. \cite{Noori2019} implemented selective PCA and related methods to map hydrothermal alteration zones. Furthermore, comprehensive fieldwork and laboratory research with methods such as X-ray diffraction (XRD), petrographic study, and spectroscopy were done to validate output maps resulted from remote sensing data processing. The findings suggest a range of high potential epithermal polymetallic vein mineralization and demonstrated potential for the method in other metallic provinces and semi-arid regions worldwide.\\
\cite{ElAtillah2019} applied PCA to extract lithological units and structures in Bou Azzer-El Graara inlier, Anti-Atlas Central, Morocco. They used a false-color composite image obtained by applying PCA on ASTER data to map lineaments. This case yielded a satisfying performance (74\%), especially after removing the lines corresponding to objects other than faults such as paths, borders, boundaries between geological formations, hill summits, and shadows. \cite{Shirmard2020} applied PCA, ICA, MNF and compared their efficiency for mapping different alteration types using ASTER data. As shown in Figure 4, they demonstrated that these dimensionality reduction techniques can be implemented jointly in remote sensing data processing for mapping those hydrothermal alteration zones related to epithermal Cu-Au deposits. PCA was used in identifying structural target features such as cracks \citep{Zhang2020} and faults \citep{Wang2020} using UAV images that can be useful in mineral exploration. Recently, PCA and MNF methods were considered in three separate datasets describing spectral imaging characteristics in geoscience and mineral exploration. The datasets were real-life case studies and obtained using innovative spectral imaging methods such as UAS-borne or small-angle terrestrial imaging and sensors at various spectral resolutions in VNIR, SWIR, and longwave infrared (LWIR) \citep{Lorenz2021}.

\begin{figure}
  \centering
  \frame{\includegraphics[width=\linewidth]{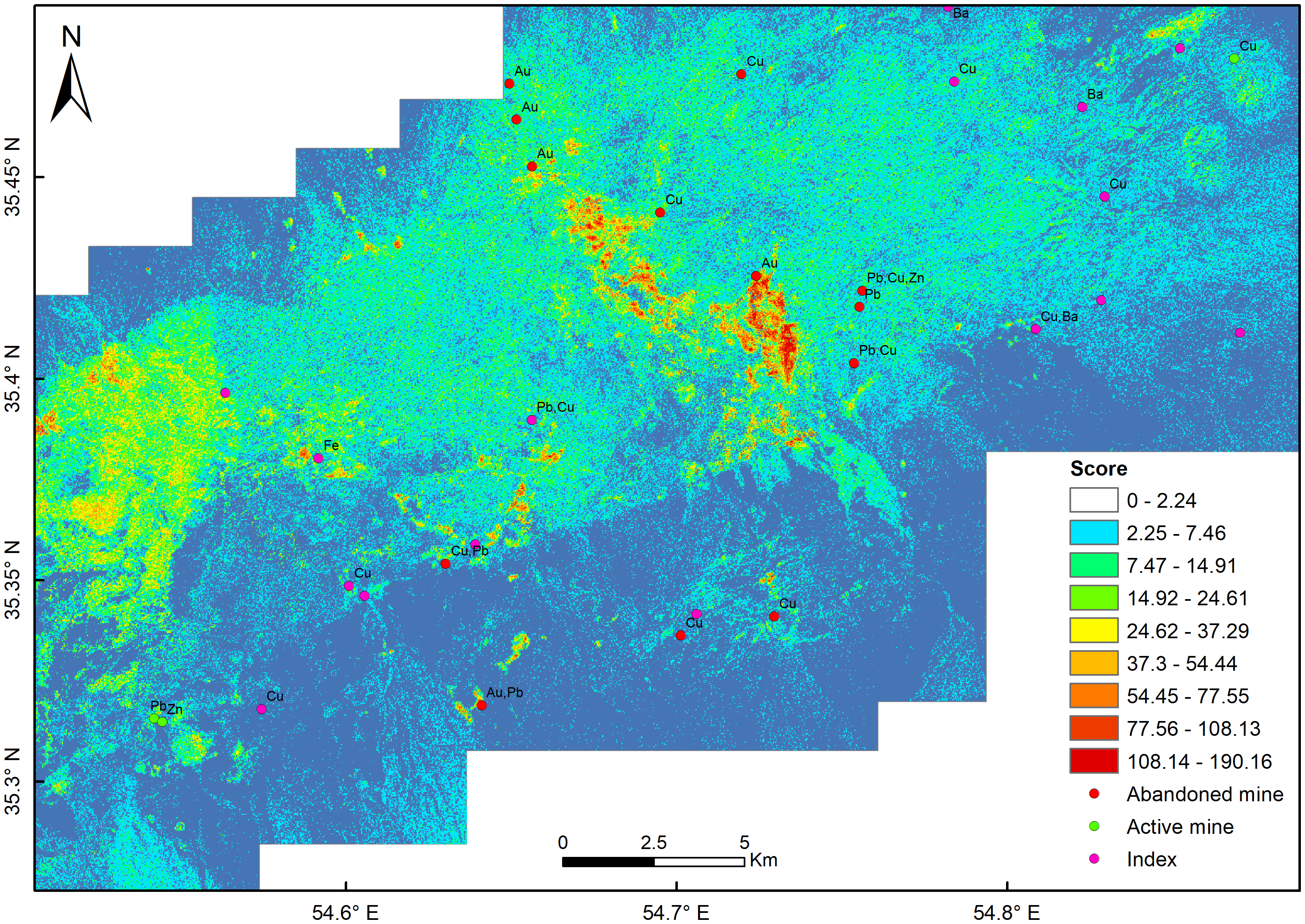}}
  \caption{Known mineral occurrences overlaid on the classified map generated by integrating selected components of dimensionality reduction techniques \citep{Shirmard2020}.}
  \label{fig4}
\end{figure}

\subsection{Classification}
\label{sec5-2}
We discuss some of the key methods used for classification problems in the scope of our applications that consider remote sensing for mineral exploration. Although some of the methods listed (such as neural networks and random forests) can also be used for regression and prediction, our focus is classification.

\subsubsection{Minimum distance classification}
\label{sec5-2-1}
The minimum distance (MD) classifier is a supervised learning method that utilizes the mean vector of each class in multi-dimensional space and measures the Euclidean distance of each class from each image pixel vector to the mean vector \citep{Richards2006}. MD has not been broadly used in remote sensing data analysis, particularly in mineral exploration. \cite{Kruse2013} defined the regions of interest to extract the endmember spectra for mineral mapping and then applied a minimum distance classification on 86-band AVIRIS SWIR data with a 7.5 m spatial resolution and simulated WV-3 SWIR data. In a recent study using Hyperion, ASTER, and OLI data, lithological units (from Udaipur district, state of Rajasthan in western India) were mapped by four supervised classifiers that include MD, SAM, SID, and support vector machine \citep{Pal2020}. \cite{Gemusse2019} used Sentinel-2, ASTER, and OLI data in a study aimed at providing land use classification maps and evaluating their accuracy. In another study, methods such as MD and SAM were used to localize pegmatites \citep{Gemusse2019}. An MD classifier is usually applied due to its flexibility and efficacy; however, it does have the downside of poor classification accuracy. In more recent work, the weighted MD classifier based on relative offset boosted the accuracy \citep{Wang2019}.

\subsubsection{Support vector machines}
\label{sec5-2-2}
support vector machine (SVM) is a supervised machine learning method used for classification and regression problems which over the years has proven to be robust and effective \citep{Cortes1995}. SVM projects the dataset into another feature space in which the dimensionality is smaller than that of the input space, which makes the process of classification much simpler \citep{Tahmasebi2020}. SVM has remarkably found its application in mineral exploration, particularly for processing remote sensing data with key developments discussed here-forth.\\
\cite{Othman2014} demonstrated the ability of SVM in targeting chromite deposits by classifying the lithological units of the Mawat ophiolite complex in northeastern Iraq utilizing ASTER data. In this study, multiple types of surface reflectance, at-sensor temperature, texture, and geomorphic parameters were jointly processed to create layers with maximum precision in classification. The findings led to the discovery of a new chromite-bearing site spanning over 0.3 $km^2$. The collected samples revealed that the chromite was coarsely crystalline within dunite host rocks. \cite{Xu2019} showed that mapping alteration zones using remote sensing data and SVM was significant for metallogenic prediction of gold deposits. In this study, remote sensing data were used to circle favorable metallogenic areas and find new gold occurrences. The study found that alteration maps are highly reliable and could play a crucial role in discovering metallic minerals. The favorable areas of gold deposition can be outlined according to alteration zones and the findings of field inspection.\\
In classification and detailed extraction of rocks and minerals, \cite{Wang2010a} used the SVM classifier for alteration mapping by processing two Hyperion images of a study area in Beiya, northwest of Yunnan, China. The results indicated that alteration zones were reliably delineated using Hyperion data. Mineralogical and lithological knowledge derived from Hyperion data was fairly similar to the geological map and prior study findings in a gold deposit. \cite{Abdolmaleki2020} applied SVM in a study for the classification and evaluation of Sentinel-2. They created a mineral prospectivity map by combining remote sensing, geological, geophysical, and geochemical data. The potential map of IOCG mineralization indicated that it is in good alignment with previous geological field studies. \cite{Gasmi2016} used the VNIR and SWIR spectral bands of ASTER data, PCA, false-color composite images, and SVM for discriminating lithological units. The outcomes were consistent with the field survey and the geological map already released.\\
\cite{Cardoso-Fernandes2020a} proposed a methodology based on the semi-automatization of SVMs to map lithium-bearing pegmatites and detected known Li-pegmatite units as well as other Li-exploration target zones. It was also revealed that class inequality negatively affected SVM accuracy since known Li-pegmatites were not detected. Recently, the applications of remote sensing data obtained from UAVs with SVM were reviewed in mining from exploration to reclamation \citep{Park2020}. In an innovative study, \cite{Lorenz2021} implemented the SVM classification method for mapping mineralization zones using the data obtained by a drone.

\subsubsection{Artificial neural networks}
\label{sec5-2-3}
Artificial neural networks, also known as neural networks are machine learning methods inspired from biological neural systems \citep{Rumelhart1986,Lecun1988,Hornik1989}. The rapid uptake of neural network approaches in remote sensing data analysis has been mainly due to their ability to learn complex patterns and taking into account nonlinear and complex relationship between explanatory and dependent variables \citep{Lek1999}. Simple neural networks, also known as multi-layer perceptron have been applied for analyzing remote sensing data with promising results and some key studies are discussed as follows. \cite{Rigol-Sanchez2003} used a simple neural network model to discriminate areas of high mineralization potential using remote sensing data and known mineral occurrences in the Rodalquilar gold region, southeast Spain. A potential map of gold mineralization was effectively provided suggesting that both historically identified and unknown potentially mineralized areas can be recognized. These initial findings indicate that neural networks can be considered a robust spatial data modeling method for mineral exploration.\\
\cite{Leverington2010} examined the efficiency of Landsat 5 and Hyperion data for discriminating lithological groups in an area and attempted to show the usefulness of broadband and hyperspectral datasets. In this study, TM data were categorized using a neural network algorithm, and then both TM and Hyperion data were unmixed linearly using ground truth spectra. \cite{Wang2010} applied probabilistic neural networks to integrate multi-mineral anomalies caused by geological information (geology, geophysics, geochemistry, and remote sensing) and to provide a 1:25000 potential map of Molybdenum and polymetallic Pb-Zn-Ag mineralization in Luanchuan, province of Henan, China. Neural networks are an efficient tool for mapping lineaments which provide key features for mineral exploration. \cite{Borisova2014} compared the results of using neural networks for mapping lineaments with the results obtained from the visual analysis of satellite imagery and geological maps. Recent contributions in the field of neural networks have been through deep learning methods, which focus on large and complex neural network architectures for multimedia and big data-related problems \citep{Schmidhuber2015}. There is a wide range of emerging research topics in the area of neural networks, particularly in deep learning, which is reviewed later (Section 5.5).

\subsubsection{Random forest}
\label{sec5-2-4}
A decision tree is a commonly used machine learning method for classification and regression problems by constructing an inverted tree with a root node, internal nodes, and leaf nodes \citep{Quinlan1986}. The algorithm is non-parametric and can efficiently deal with large, complicated datasets without imposing a complicated parametric structure \citep{Song2015,Quinlan1987}. The major problem of decision trees is over-fitting on the training dataset and their inability to generalise over complex datasets \citep{Utgoff1989}. A random forest is an ensemble machine learning method based on bagging that creates an ensemble of decision tree-based classifiers that addresses the limitations of decision trees \citep{Breiman1996}. The random forest ensemble consists of several trees systematically constructed by arbitrarily chosen subgroups of training data from which the trees constructed independently \citep{TinKamHo1998,Belgiu2016}. It is important to note that the final decision is made using all the trees in the ensemble (forest). A voting procedure is used to make a decision using the ensemble of trees for classification problems. In regression tasks, the mean prediction of the individual trees from the ensemble is used to make a decision.\\
\cite{Chung2020} applied random forests to minimize the number of variables for the model-building to pick the best representative bands using hyperspectral remote sensing data for the classification of magnesite and gangue-related minerals, including dolomite, calcite, and talc. \cite{Cardoso-Fernandes2019} used random forest with SVMs in application to lithological mapping using the same level 1-C Sentinel-2 images acquired in October 2017. The outcrops were mapped using both methods that correctly identified Li-bearing pegmatites in three open-pit mines.\\
\cite{Kuhn2018} used the random forest classifier to characterize the lithology of a largely unexplored region neighboring a historically important gold mine, using a combination of geophysical and remote sensing data. Given minimal data, the authors found the random forest can be an essential additional tool available to geoscientists in a greenfield of orogenic gold mineralization. In another study \citep{Cracknell2014}, five machine learning algorithms (naive-Bayes, K-nearest neighbor, random forest, SVM, and simple neural networks) were compared in terms of the efficiency in extracting different lithological units using Landsat 7, and spatially controlled remotely sensed geophysical data. \cite{Wang2020a} integrated multisource and multi-sensor remote sensing data and applied random forest to discriminate critical lithological units through exploring rare metals that promoted computing efficiency and classification accuracy.\\
Random forests have a number of applications in remote sensing data analysis for the extraction of target features critical for mineral exploration. \cite{Cracknell2013} compared random forests and SVMs for the inference of the spatial distribution of lithology from combined airborne geophysical and multispectral satellite data in a dynamic, folded, and high-grade metamorphic terrane. \cite{Bachri2020} combined the spectral, textural, and geomorphic information of Sentinel-2 and PALSAR and mapped lithological features using random forests. Mineral exploration and geological mapping in a region with steep topography, dense vegetation, and limited outcrop are challenging. SAR can potentially penetrate vegetation canopies and assist geological mapping in such environments. \cite{Radford2018} applied random forests to classify rock units using airborne polarimetric terrain observation by progressive scan (TOPSAR) and geophysical data. In another study, \cite{Belgiu2016} reviewed the applications of random forests in remote sensing data analysis.

\subsection{Clustering}
K-means clustering (KMC) can be used for separating $N$ data points into $K$ clusters \citep{Lloyd1982}, where each data point is categorized into a cluster with the smallest discrepancy among its value and the mean value of the cluster. In order to classify the representative data known as cluster centers, KMC can handle a dataset with high dimensionality \citep{Tang2019}. As a standard clustering algorithm, KMC is widely used to process hyperspectral images for recognizing objects. However, the alignment of data points and cluster centers with the standard KMC algorithm can become very complicated due to mixed pixels. In order to classify different types of minerals using the AVIRIS hyperspectral image of the Cuprite mining area, \cite{Ren2019} suggested an improved KMC algorithm. They utilized three approaches to pick the initial cluster centers and spectral information separation instead of the Euclidean distance to determine the similarity. Finally, they found that the enhanced KMC algorithm could get stronger clustering results and greater mineral mapping precision than the conventional algorithm by comparing the clustering results with the mineral distribution map of the study area and the United States Geological Survey (USGS) mineral spectral library. \cite{ElAtillah2019} generated a lithological map using KMC in a study area using ASTER, Landsat 7, Landsat 8, and Sentinel-2 data. The accuracy of the results was tested by comparison with field data and geological maps of the study area. In an unaltered rock area of Cuprite, Nevada, USA, KMC results were combined with the results obtained from spectral unmixing and full-pixel classification to produce a single distribution map of rocks and minerals. In this study, KMC was applied to airborne hyperspectral HyMap images, verifying that the result of mapping was consistent with both available geological maps and the outcome of the field survey \citep{Ishidoshiro2016}.\\
We note that KMC is a canonical clustering method and there has been major progress in this area for large and more complex datasets. The applications that used KMC can be further improved with these methods. Agglomerative clustering is the most common type of hierarchical clustering used to group objects in clusters based on their similarity \citep{Olson1995,Johnson1967}. It is also known as AGNES (Agglomerative Nesting) where the algorithm starts by treating each object as a singleton cluster. The pairs of clusters are successively merged until all clusters have been merged into one big cluster containing all objects. The result is a tree-based representation of the objects, named dendrogram \citep{Navarro1997}. Density-based spatial clustering of applications with noise (DBSCAN) on the other hand has been a popular clustering algorithm that uses local density estimation to identify clusters of arbitrary shapes not possible with the KMC \citep{Ester1996}. Spectral clustering transforms a clustering problem into a graph partitioning problem and has been one of the more recent addition to the family of clustering methods \citep{Biernacki2000}. The aim of spectral clustering is to identify sub-graphs (representing communities) based on the connections between nodes \citep{VonLuxburg2007}.\\
An unsupervised classification approach typically applied to satellite images is the iterative self-organizing data processing technique (ISODATA) \citep{Karimi2004}. In a multi-dimensional attribute space, the spectral reflectance from several bands has been used to evaluate clusters. Multispectral and hyperspectral images have been widely classified using ISODATA \citep{Sydow1977}. The number of features used for classification can be expanded by incorporating other bands or external data measured on a continuous scale, such as digital elevation maps or geophysical data.\\
\cite{ElAtillah2019} carried out lithological cartography using ISODATA and KMC methods. An idea of the mineralogy of the study area was provided by the assembly of lithological, structural, and hydrothermal alteration data derived from ASTER, ETM+, OLI, and Sentinel-2 data. The quality of findings was assessed and compared with field data and geological maps of the study area. In a study, \cite{Ducart2016} employed an ISODATA model to a data collection consisting of Hyperion VNIR 74 bands, LIDAR-derived digital terrain model, gamma-ray spectroscopy, gravimetry, and OLI-derived normalized difference vegetation index (NDVI) data. The selection of these data for the classification was based on their own importance in outcropping iron ore identification.

\subsection{Regression analysis}
\label{sec5-4}
Regression is a methodology for predictive analysis that can calculate the relative effect of many factors statistically and explain logically how values depend on predictor variables \citep{Sudaryatno2020}. Regression analysis as a branch of machine learning and statistics which can be effectively used in remote sensing for predicting target zones. \cite{Holloway2018} applied logistic, linear, multinomial regression, and boosted regression trees in remote sensing data analysis. \cite{Yetkin2004} used multi-linear regression analysis for mapping an alteration trend (in Melendiz Volcanic Complex, Turkey), where some mineral occurrences were extracted to detect alteration zones such as potassic, phyllic (sericitic), propylitic, argillic, and silicification. Moreover, the youngest volcanic complex in Hasandag was discovered to be poorly altered, and the volcanic complexes of Keciboyduran, Melendiz, and Tepek\"oy were found to be strongly altered. The alteration zones involved buried faults that may provide evidence for the source of alteration.\\
\cite{Mansouri2018} used multivariate regression to construct a statistical model to map iron outcrops where the accuracy of the model was verified by a map of iron outcrops and geological exploration. Based on field observation, the mineralization of iron has occurred in contact with limestone and intrusive rocks. \cite{Hoang2017} established a novel tool for translating ETM+ imagery into pseudo-Hyperion imagery using their band reflectance data correlations. This approach can be applied for mapping metallic minerals such as gold, silver, and copper. \cite{Lin2020} built evidential variables using geological, remote sensing, and geochemical data. They designed a conjugate gradient logistic regression (CG-LR) model based on these evidential variables to predict exploration goals in the study area. Finally, geological, remote sensing and geochemical data were efficiently integrated into the CG-LR model to predict skarn deposits. \cite{Li2018} presented a multivariate regression model based on hyperspectral imaging to quantitatively analyze and predict the device capability and optimize device parameters for detecting minerals and geological survey.\\
We note that uncertainty quantification in predictions is important and there lies uncertainty due to model parameters and data collection due to a wide range of issues such as noise, sparse datasets, and limitations of sensors used for data collection \citep{Gahegan2000}. Although not much has been done in terms of uncertainty quantification, there is scope for using Bayesian regression analysis \citep{Fernandez2000} with Bayesian linear and logistic regression \citep{Genkin2007} in the scope of remote sensing \citep{Ruiz2014,Storvik2005} and mineral exploration \citep{Porwal2006,Rendu1976}.

\subsection{Deep learning}
\label{sec5-5}
There exist only a few studies with the application of deep learning in processing remote sensing data for mineral exploration. Deep learning methods are neural network-based machine learning methods that can be used for supervised and unsupervised learning \citep{Guo2016}. Moreover, they can be used for semi-supervised learning, which refers to a learning problem that involves a small portion of labeled examples and a large number of unlabeled examples from which a model must learn and make predictions on new examples \citep{Zhu2009}. Deep learning methods have the ability to model complex and large datasets and are loosely grouped into feedforward and recurrent neural network architectures \citep{Shrestha2019}. Among a variety of deep learning methods, convolutional neural networks (CNNs) have gained more attention in remote sensing due to their success in image processing \citep{Lecun1998}. CNNs belong to a class of deep feedforward neural networks that have been successfully applied to image processing tasks due to their ability to provide automatic feature extraction using convolutional and pooling layers \citep{Fu2019}. The development of a CNN model varies for different datasets and tasks, and the optimal number of a convolutional and pooling layer is often experimentally determined in multimedia applications. Furthermore, pre-trained CNN architectures are widely shared openly that address the limitations of computational resource needs for large datasets and models \citep{Tahmasebi2020,Alom2018}.\\
\cite{Latifovic2018} reviewed the ability of deep neural networks to assist mapping of geological target features by offering an initial objective layer of surface materials that experts can change to accelerate the production of maps and increase the accuracy between mapped areas. The CNN was evaluated for predicting surficial geological classes under two sampling scenarios. In the first scenario, a CNN used the samples obtained from the field that needed to be mapped and in the second, a CNN trained in one field is then extended to areas where the samples available were not included in CNN model training. The study was carried out in five areas using aerial imagery, Landsat reflectance, and high-resolution digital elevation data. Finally, the fusion spatial-spectral function was primarily derived in a model by stacking local spatial characteristics obtained by a CNN-based model and spectral data.\\
Recurrent neural networks are deep learning methods that have the ability to model temporal data and dynamical systems \citep{Elman1990,Werbos1990}. The long-short term memory (LSTM) network \citep{Hochreiter1997} is a prominent recurrent neural network that has been applied to a wide range of spatiotemporal problems, including remote sensing; however, not much has been done in the scope of mineral exploration. There exists a number of combinations of LSTMs with other deep learning methods, such as LSTM-CNN combination for  image datasets with temporal features such as real-time video processing \citep{Xia2020}. Convolutional LSTM has been developed based on features from CNN which has been useful for spatiotemporal datasets \citep{Shi2015}. These have been used for remote sensing applications \citep{Li2020a,Kwak2019,Boulila2021} and have potential in the scope of mineral exploration.\\
\cite{Zhao2020} applied the CNN method on the hyperspectral data obtained in the Nevada mining region by the AVIRIS. The hierarchical spatial-spectral feature extraction with long short-term memory (HSS-LSTM) classification achieved accuracy of 94.70 percent which exceeded the most widely used methods. \cite{Sang2020} proposed a deep learning framework for high-resolution target feature mapping using UAV data. The UAV collected high-resolution images incorporated groundwork study to support lithological mapping. The framework provides an automated mapping mechanism based on the basic simple linear iterative clustering-convolutional neural network (SLIC-CNN). The CNN was used to classify the structure of the image and to validate the lithological distribution, while SLIC was used to outline the rock mass boundary. The fusion and mapping results were explained by the mode and expert decision process.\\
Recently, \cite{Li2020} proposed a new deep learning-based multi-label remote sensing image scene classification (MLRSSC) framework by integrating CNN and graph neural network (GNN). The multi-layer integration graph attention network model addresses MLRSSC, where graph attention network has been used to exploit the spatio-topological relationships of the scene graph completely. Extensive tests on two public MLRSSC datasets demonstrated that CNN-GNN combination achieved superior efficiency. Furthermore, \cite{Gao2021} presented a novel remote sensing scene classification approach based on a high-order graph convolutional network (H-GCN). The experimental reports on test functions indicated the viability and usefulness of the proposed approach for the classification of remote sensing scenes. These novel methods are effective in analyzing remote sensing data and can be used for mineral exploration purposes.

\section{Discussion: challenges and future prospective}
\label{sec6}
Mineral exploration is the process of discovering economically productive quantities of minerals that involves a chain of events that ideally leads to a mineable resource. The exploration methods taken to find new mineral deposits can differ based on the type of deposit, the location of the study area, the presence of infrastructure, and the presence and nature of existing geological knowledge available in an area. Mineral exploration needs theoretical knowledge about how and why a mineral exists in nature at a certain location in the Earth's crust in order to provide an exploration plan. Mineral discovery aims to locate an economic deposit at the lowest possible cost and in the shortest possible period. The success rate of exploration and the return on investment are low, because exploration is an incredibly risky sector. In recent years, due to the decline of the discovery success rate of mineral exploration and the increasing demand for critical metals, geologists are encouraged to apply new data types and approaches to identify new mineral deposits. With the introduction of remote sensing data, supercomputers, and modeling based on machine learning methods, there is more potential for new resource discovery although complications and challenges exist. At present, much of the groundbreaking activities in the field of mineral discovery rely on three areas: digitization and the use of artificial intelligence, remote sensing and geophysical technologies development, and new approaches to exploration through cover \citep{Gonzalez-Alvarez2020}.\\
The challenges of remote sensing given big data and complex machine learning methods face our era. Big data refers to a series of datasets so large and diverse that conventional algorithms and models for data analysis are infeasible. The challenges are in acquisition, preparation, searching, sharing, transition, analysis, and interpretation of the data. The complexity of machine learning methods in terms of the number of model parameters and model architecture has been increasing to address different data types. Therefore, when using remote sensing to understand geo-processes, it is critical to consider multi-source, multi-scale, high-dimensional, dynamic state, isomer, and non-linear characteristics of remote sensing data. The characteristics are fundamental assumptions and objectives when evaluating remote sensing big data and extracting information from them \citep{Liu2015}.\\
Satellite, airborne, and ground-based remote sensing data are typically very large and complicated datasets that take much time to decode without any assurance that hidden information could be detected. Valuable information can be decoded from such dynamic signals with the help of machine learning methods. However, emerging machine learning methods have not been applied in some branches of science. Recent deep learning methods could be applied in mineral exploration using remote sensing data. CNNs are ideal for automating feature extraction and capturing complex relationships in image-based data which is highly applicable for different types of remote sensing data for problems in geosciences. CNN has been designed to concentrate on salient characteristics and spatial dependence in images that can reflect valuable characteristics such as nearby regions in the case of remote sensing data. Experimental findings suggest that spectral and spatial knowledge can be exploited entirely by CNNs. In this context, \cite{Chen2017} proposed a hyperspectral image (HSI) classification approach based on CNN where the fully convolutional kernels can be learned from the data automatically by clustering without knowing the number of clusters. The CNN approach achieved improved classification performance which has been useful in identifying and discriminating target features for mineral exploration such as lithological units, alteration types, structures, and indicator minerals.\\
Although machine learning methods have been widely used for remote sensing data analysis, there remains a gap in uncertainty quantification in predictions. In general, assessing the uncertainty in remote sensing-based model applications is challenging. The model validation is usually carried out via comparison with ground truth or alternative information that presumably represents the ground truth. Different approaches have been developed to address the validation problem, which results in a large variety of potential models. A number of studies reviewed remote sensing data validation methods and documented their similarities and differences \citep{Wang2016,Loew2017,Wu2019}. As mentioned earlier, Bayesian inference provides a principled approach to uncertainty quantification in model parameters. Recently, there has been tremendous progress in the area of Bayesian neural networks and Bayesian deep learning \citep{Wang2020b,Shi2017}; however, these methods have not been much used for remote sensing, and their application for mineral exploration is absent. These methods have the potential to provide the meaningful interpretation of remote sensing data given challenges such as noise in data, sparse datasets, and missing data. The quantification of uncertainty using Bayesian inference can be used to project the uncertainty associated with model parameters and data. Recently, \cite{Chandra2020} proposed a Bayesian neural network framework that takes into account multiple data sources using Markov Chain Monte Carlo (MCMC) sampling which can be extended in the field of remote sensing and mineral exploration given multiple sources of data and information. Bayesian framework driven by MCMC sampling has been used for geoscientific models where different sources of information have been incorporated for inference of geophysical parameters in landscape evolution models \citep{Chandra2019}. The approach has also been used in 3D geological and geophysical data fusion for mineral prospectivity \citep{Olierook2021}. Bayesian framework has also been used for spatio-temporal reconstruction of lithologies in a paleo-climatology study for the last 250 million years \citep{Chandra2021}. Therefore, the Bayesian approach has good potential to integrate remote sensing data to explore mineral deposits.\\
As an alternative approach, transfer learning can be applied in remote sensing data processing for detecting ore deposits. Transfer learning is inspired by the field of educational psychology and refers to the transfer of learning that happens when learning in one condition increase (positive transfer) or decrease (negative transfer) the performance \citep{Baldwin1988}. There are challenges in deep learning methods given several sources of knowledge and data that needs to be integrated for developing a robust model. The autoencoder is a prominent unsupervised machine learning method which aims to compress data and reconstruct it accurately \citep{Wang2014}. Remote sensing data needs to be compressed for machine learning methods and the use of autoencoders and its variants such as deep autoencoders that feature CNN \citep{Pu2016} can be useful for mineral exploration.\\
Generative adversarial networks (GANs) are prominent deep learning methods used for generating images and data \citep{Goodfellow2014}. GANs are more recently becoming very popular for addressing the issue of class-imbalanced problems \citep{Ali-Gombe2019,Shamsolmoali2020}, which are key challenges in geoscience and remote sensing. Hence, they have the potential to be used for remote sensing \citep{Jiang2019} and mineral exploration to generate data for class imbalanced problems or to reconstruct data in cases of missing regions in remote sensing data. Finally, the use of spatio-temporal tectonic data analysis to augment machine learning models for mineral exploration can also be explored for understanding the original tectonic environments in which ore deposits may have formed. Recently, \citep{DiazRodriguez2021} used machine learning methods with data generated from a plate tectonic model that captures the time-dependence of subduction zone evolution in the context of porphyry Cu systems for the last 100 million years. Such spatio-temporal machine learning approaches, placing ore deposits in a plate tectonic and plate boundary evolution context, have the potential to significantly improve our understanding of the geological niche environments that give rise to particular ore deposits in space and time.\\
The computational cost of machine learning algorithms increases rapidly with the increase in size and quality of remote sensing data. The trade-off between the result accuracy and the computational effort should always be taken into account. In general, data-driven techniques may pose the risk of over-fitting and problem dependence which requires additional research. The results of previous studies indicate that there is no universal approach to solve all problems. Thus, the selection of the appropriate technique is based on specific applications and experiment conditions. Recently developed machine learning algorithms enable geologists to deal with high-complexity and high-dimensional problems or data. However, questions remain with regard to the efficiency of novel algorithms concerning the computational costs of conducting data analysis of high-dimensional data, handling outliers, and providing more insight into the analysis of black-box models. In summary, machine learning methods enable scientists to analyze complex datasets and solve challenging big data problems. Despite their complexity and the lack of clear and reliable guidelines for model settings and the interpretation of results, machine learning methods have been widely distributed and implemented. Nevertheless, to pursue such state-of-the-art methods, geologists must be motivated by a deeper understanding of problems in the area of mineral exploration.

\section{Conclusions}
\label{sec7}
We reviewed the implementation and adaptation of some popular and recently established machine learning methods for remote sensing data processing and investigated their applications for exploring different ore deposits. Remote sensing datasets have provided a new data resource to overcome problems associated with mapping geological features from field data alone. As a data-driven classification or prediction tool, neural networks have been widely applied in remote sensing data processing as well as a large number of research areas ranging from engineering and environmental science to physics and astronomy. Dimensionality reduction techniques can transform high-dimensional problems into a low-dimensional space and potentially mine special features from remote sensing data for mineral exploration. Recent advancements in deep learning methods have the potential to deal with large and complex remote sensing data with features in processing spectral and ground truth measurements against noise and uncertainties. Deep learning methods can be very effective in identifying target features and mineral discovery using remote sensing data. Advanced deep learning methods can improve the mapping of geological target features for both small and large-scale studies as the success rate of mineral exploration in the face of increasing demand for critical metals. The limitations of different machine learning methods and their specific requirements are the key obstacles that exploration geologists have been facing. In general, using advanced analytics in mineral exploration is important for achieving sustainable development goals in the mining industry. These techniques can help geoscientists limit the negative impact of mineral prospecting activities on the ecology, environment, and climate with efficient and effective cost-cutting solutions.


\section*{Acknowledgments}
We would like to express our gratitude to Prof. Gregoire Mariethoz for his constructive comments that helped to improve this paper. We also thank Prof. Jing M. Chen and Prof. Joseph Awange for handling this paper and anonymous reviewers for their insightful comments and suggestions.

\bibliographystyle{elsarticle-harv}
\bibliography{References}

\begin{thebibliography}{234}
\expandafter\ifx\csname natexlab\endcsname\relax\def\natexlab#1{#1}\fi
\providecommand{\url}[1]{\texttt{#1}}
\providecommand{\href}[2]{#2}
\providecommand{\path}[1]{#1}
\providecommand{\DOIprefix}{doi:}
\providecommand{\ArXivprefix}{arXiv:}
\providecommand{\URLprefix}{URL: }
\providecommand{\Pubmedprefix}{pmid:}
\providecommand{\doi}[1]{\href{http://dx.doi.org/#1}{\path{#1}}}
\providecommand{\Pubmed}[1]{\href{pmid:#1}{\path{#1}}}
\providecommand{\bibinfo}[2]{#2}
\ifx\xfnm\relax \def\xfnm[#1]{\unskip,\space#1}\fi
\bibitem[{Abdolmaleki et~al.(2020)Abdolmaleki, Rasmussen and
  Pal}]{Abdolmaleki2020}
\bibinfo{author}{Abdolmaleki, M.}, \bibinfo{author}{Rasmussen, T.},
  \bibinfo{author}{Pal, M.}, \bibinfo{year}{2020}.
\newblock \bibinfo{title}{{Exploration of IOCG mineralizations using
  integration of space-borne remote sensing data with airborne geophysical
  data}}.
\newblock \bibinfo{journal}{ISPRS - International Archives of the
  Photogrammetry, Remote Sensing and Spatial Information Sciences}
  \bibinfo{volume}{XLIII-B3-2}, \bibinfo{pages}{9--16}.
\newblock \DOIprefix\doi{10.5194/isprs-archives-XLIII-B3-2020-9-2020}.
\bibitem[{Agar and Coulter(2007)}]{Agar2007}
\bibinfo{author}{Agar, B.}, \bibinfo{author}{Coulter, D.},
  \bibinfo{year}{2007}.
\newblock \bibinfo{title}{{Remote sensing for mineral exploration – A decade
  perspective 1997-2007}}, in: \bibinfo{booktitle}{Fifth Decennial
  International Conference on Mineral Exploration}, pp.
  \bibinfo{pages}{109--136}.
\bibitem[{Agar(1994)}]{Agar1994}
\bibinfo{author}{Agar, R.A.}, \bibinfo{year}{1994}.
\newblock \bibinfo{title}{{Geoscan airborne multi-spectral scanners as
  exploration tools for Western Australian diamond and gold deposits}}.
\newblock \bibinfo{journal}{ASEG Extended Abstracts} \bibinfo{volume}{1994},
  \bibinfo{pages}{433--448}.
\newblock \DOIprefix\doi{10.1071/ASEGSpec07_33}.
\bibitem[{Ahmadirouhani et~al.(2018)Ahmadirouhani, Karimpour, Rahimi,
  Malekzadeh-Shafaroudi, Beiranvand~Pour and Pradhan}]{Ahmadirouhani2018}
\bibinfo{author}{Ahmadirouhani, R.}, \bibinfo{author}{Karimpour, M.H.},
  \bibinfo{author}{Rahimi, B.}, \bibinfo{author}{Malekzadeh-Shafaroudi, A.},
  \bibinfo{author}{Beiranvand~Pour, A.}, \bibinfo{author}{Pradhan, B.},
  \bibinfo{year}{2018}.
\newblock \bibinfo{title}{Integration of spot-5 and aster satellite data for
  structural tracing and hydrothermal alteration mineral mapping: implications
  for cu–au prospecting}.
\newblock \bibinfo{journal}{International Journal of Image and Data Fusion}
  \bibinfo{volume}{9}, \bibinfo{pages}{237--262}.
\newblock \DOIprefix\doi{10.1080/19479832.2018.1469548}.
\bibitem[{Al-Nahmi et~al.(2017)Al-Nahmi, Saddiqi, Hilali, Rhinane, Baidder,
  El~arabi and Khanbari}]{Al-Nahmi2017}
\bibinfo{author}{Al-Nahmi, F.}, \bibinfo{author}{Saddiqi, O.},
  \bibinfo{author}{Hilali, A.}, \bibinfo{author}{Rhinane, H.},
  \bibinfo{author}{Baidder, L.}, \bibinfo{author}{El~arabi, H.},
  \bibinfo{author}{Khanbari, K.}, \bibinfo{year}{2017}.
\newblock \bibinfo{title}{{Application of remote sensing in geological mapping,
  case study Al Maghrabah Area - Hajjah region, Yemen}}.
\newblock \bibinfo{journal}{ISPRS Annals of Photogrammetry, Remote Sensing and
  Spatial Information Sciences} \bibinfo{volume}{IV-4/W4},
  \bibinfo{pages}{63--71}.
\newblock \DOIprefix\doi{10.5194/isprs-annals-IV-4-W4-63-2017}.
\bibitem[{Ali and Beiranvand~Pour(2014)}]{Ali2014}
\bibinfo{author}{Ali, A.}, \bibinfo{author}{Beiranvand~Pour, A.},
  \bibinfo{year}{2014}.
\newblock \bibinfo{title}{{Lithological mapping and hydrothermal alteration
  using Landsat 8 data: A case study in Ariab mining district, Red Sea Hills,
  Sudan}}.
\newblock \bibinfo{journal}{International Journal of Basic and Applied
  Sciences} \bibinfo{volume}{3}.
\newblock \DOIprefix\doi{10.14419/ijbas.v3i3.2821}.
\bibitem[{Ali et~al.(2015)Ali, Greifeneder, Stamenkovic, Neumann and
  Notarnicola}]{Ali2015}
\bibinfo{author}{Ali, I.}, \bibinfo{author}{Greifeneder, F.},
  \bibinfo{author}{Stamenkovic, J.}, \bibinfo{author}{Neumann, M.},
  \bibinfo{author}{Notarnicola, C.}, \bibinfo{year}{2015}.
\newblock \bibinfo{title}{{Review of machine learning approaches for biomass
  and soil moisture retrievals from remote sensing data}}.
\newblock \bibinfo{journal}{Remote Sensing} \bibinfo{volume}{7},
  \bibinfo{pages}{16398--16421}.
\newblock \DOIprefix\doi{10.3390/rs71215841}.
\bibitem[{Ali-Gombe and Elyan(2019)}]{Ali-Gombe2019}
\bibinfo{author}{Ali-Gombe, A.}, \bibinfo{author}{Elyan, E.},
  \bibinfo{year}{2019}.
\newblock \bibinfo{title}{Mfc-gan: Class-imbalanced dataset classification
  using multiple fake class generative adversarial network}.
\newblock \bibinfo{journal}{Neurocomputing} \bibinfo{volume}{361},
  \bibinfo{pages}{212--221}.
\newblock \DOIprefix\doi{10.1016/j.neucom.2019.06.043}.
\bibitem[{Alom et~al.(2018)Alom, Taha, Yakopcic, Westberg, Sidike, Nasrin,
  Esesn, Awwal and Asari}]{Alom2018}
\bibinfo{author}{Alom, M.Z.}, \bibinfo{author}{Taha, T.M.},
  \bibinfo{author}{Yakopcic, C.}, \bibinfo{author}{Westberg, S.},
  \bibinfo{author}{Sidike, P.}, \bibinfo{author}{Nasrin, M.S.},
  \bibinfo{author}{Esesn, B.C.V.}, \bibinfo{author}{Awwal, A.A.S.},
  \bibinfo{author}{Asari, V.K.}, \bibinfo{year}{2018}.
\newblock \bibinfo{title}{The history began from alexnet: A comprehensive
  survey on deep learning approaches}.
\newblock \href{http://arxiv.org/abs/1803.01164}{{\tt arXiv:1803.01164}}.
\bibitem[{Amer et~al.(2010)Amer, Kusky and Ghulam}]{Amer2010}
\bibinfo{author}{Amer, R.}, \bibinfo{author}{Kusky, T.},
  \bibinfo{author}{Ghulam, A.}, \bibinfo{year}{2010}.
\newblock \bibinfo{title}{{Lithological mapping in the Central Eastern Desert
  of Egypt using ASTER data}}.
\newblock \bibinfo{journal}{Journal of African Earth Sciences}
  \bibinfo{volume}{56}, \bibinfo{pages}{75--82}.
\newblock \DOIprefix\doi{10.1016/j.jafrearsci.2009.06.004}.
\bibitem[{Asadzadeh and {de Souza Filho}(2016)}]{Asadzadeh2016}
\bibinfo{author}{Asadzadeh, S.}, \bibinfo{author}{{de Souza Filho}, C.R.},
  \bibinfo{year}{2016}.
\newblock \bibinfo{title}{{A review on spectral processing methods for
  geological remote sensing}}.
\newblock \bibinfo{journal}{International Journal of Applied Earth Observation
  and Geoinformation} \bibinfo{volume}{47}, \bibinfo{pages}{69--90}.
\newblock \DOIprefix\doi{10.1016/j.jag.2015.12.004}.
\bibitem[{Asokan et~al.(2020)Asokan, Anitha, Ciobanu, Gabor, Naaji and
  Hemanth}]{Asokan2020}
\bibinfo{author}{Asokan, A.}, \bibinfo{author}{Anitha, J.},
  \bibinfo{author}{Ciobanu, M.}, \bibinfo{author}{Gabor, A.},
  \bibinfo{author}{Naaji, A.}, \bibinfo{author}{Hemanth, D.J.},
  \bibinfo{year}{2020}.
\newblock \bibinfo{title}{{Image processing techniques for analysis of
  satellite images for historical maps classification—an overview}}.
\newblock \bibinfo{journal}{Applied Sciences} \bibinfo{volume}{10},
  \bibinfo{pages}{4207}.
\newblock \DOIprefix\doi{10.3390/app10124207}.
\bibitem[{Awad et~al.(2018)Awad, Amer, L{\'{o}}pez-Galindo, El-Rahmany,
  {Garc{\'{i}}a del Moral} and Viseras}]{Awad2018}
\bibinfo{author}{Awad, M.E.}, \bibinfo{author}{Amer, R.},
  \bibinfo{author}{L{\'{o}}pez-Galindo, A.}, \bibinfo{author}{El-Rahmany,
  M.M.}, \bibinfo{author}{{Garc{\'{i}}a del Moral}, L.F.},
  \bibinfo{author}{Viseras, C.}, \bibinfo{year}{2018}.
\newblock \bibinfo{title}{{Hyperspectral remote sensing for mapping and
  detection of Egyptian kaolin quality}}.
\newblock \bibinfo{journal}{Applied Clay Science} \bibinfo{volume}{160},
  \bibinfo{pages}{249--262}.
\newblock \DOIprefix\doi{10.1016/j.clay.2018.02.042}.
\bibitem[{Babbar and Rathee(2019)}]{Babbar2019}
\bibinfo{author}{Babbar, J.}, \bibinfo{author}{Rathee, N.},
  \bibinfo{year}{2019}.
\newblock \bibinfo{title}{{Satellite image analysis: A review}}, in:
  \bibinfo{booktitle}{IEEE International Conference on Electrical, Computer and
  Communication Technologies)}, pp. \bibinfo{pages}{1--6}.
\newblock \DOIprefix\doi{10.1109/ICECCT.2019.8869481}.
\bibitem[{Bachri et~al.(2020)Bachri, Hakdaoui, Raji and
  Benbouziane}]{Bachri2020}
\bibinfo{author}{Bachri, I.}, \bibinfo{author}{Hakdaoui, M.},
  \bibinfo{author}{Raji, M.}, \bibinfo{author}{Benbouziane, A.},
  \bibinfo{year}{2020}.
\newblock \bibinfo{title}{{Geological mapping using random forests applied to
  remote sensing data: A demonstration study from Msaidira-Souk Al Had, Sidi
  Ifni inlier (Western Anti-Atlas, Morocco)}}, in: \bibinfo{booktitle}{IEEE
  International conference of Moroccan Geomatics (Morgeo)},
  \bibinfo{publisher}{IEEE}. pp. \bibinfo{pages}{1--5}.
\newblock \DOIprefix\doi{10.1109/Morgeo49228.2020.9121888}.
\bibitem[{Bachri et~al.(2019)Bachri, Hakdaoui, Raji, Teodoro and
  Benbouziane}]{Bachri2019}
\bibinfo{author}{Bachri, I.}, \bibinfo{author}{Hakdaoui, M.},
  \bibinfo{author}{Raji, M.}, \bibinfo{author}{Teodoro, A.C.},
  \bibinfo{author}{Benbouziane, A.}, \bibinfo{year}{2019}.
\newblock \bibinfo{title}{{Machine learning algorithms for automatic
  lithological mapping using remote sensing data: A case study from Souk Arbaa
  Sahel, Sidi Ifni Inlier, Western Anti-Atlas, Morocco}}.
\newblock \bibinfo{journal}{ISPRS International Journal of Geo-Information}
  \bibinfo{volume}{8}, \bibinfo{pages}{248}.
\newblock \DOIprefix\doi{10.3390/ijgi8060248}.
\bibitem[{Bailey et~al.(2012)Bailey, Whitmeyer and De~Paor}]{Bailey2012}
\bibinfo{author}{Bailey, J.}, \bibinfo{author}{Whitmeyer, S.},
  \bibinfo{author}{De~Paor, D.}, \bibinfo{year}{2012}.
\newblock \bibinfo{title}{Introduction: The application of google geo tools to
  geoscience education and research}.
\newblock \bibinfo{journal}{Google Earth and Virtual Visualizations in
  Geoscience Education and Research: Geological Society of America Special
  Paper} \bibinfo{volume}{492}, \bibinfo{pages}{7--19}.
\newblock \DOIprefix\doi{10.1130/2012.2492(00)}.
\bibitem[{Baldwin and Ford(1988)}]{Baldwin1988}
\bibinfo{author}{Baldwin, T.T.}, \bibinfo{author}{Ford, J.K.},
  \bibinfo{year}{1988}.
\newblock \bibinfo{title}{{Transfer of training: A review and directions for
  future research}}.
\newblock \bibinfo{journal}{Personnel Psychology} \bibinfo{volume}{41},
  \bibinfo{pages}{63--105}.
\newblock \DOIprefix\doi{10.1111/j.1744-6570.1988.tb00632.x}.
\bibitem[{Banskota et~al.(2014)Banskota, Kayastha, Falkowski, Wulder, Froese
  and White}]{Banskota2014}
\bibinfo{author}{Banskota, A.}, \bibinfo{author}{Kayastha, N.},
  \bibinfo{author}{Falkowski, M.J.}, \bibinfo{author}{Wulder, M.A.},
  \bibinfo{author}{Froese, R.E.}, \bibinfo{author}{White, J.C.},
  \bibinfo{year}{2014}.
\newblock \bibinfo{title}{{Forest monitoring using Landsat time series data: A
  review}}.
\newblock \bibinfo{journal}{Canadian Journal of Remote Sensing}
  \bibinfo{volume}{40}, \bibinfo{pages}{362--384}.
\newblock \DOIprefix\doi{10.1080/07038992.2014.987376}.
\bibitem[{Bartalev et~al.(2014)Bartalev, Egorov, Loupian and
  Khvostikov}]{Bartalev2014}
\bibinfo{author}{Bartalev, S.}, \bibinfo{author}{Egorov, V.},
  \bibinfo{author}{Loupian, E.}, \bibinfo{author}{Khvostikov, S.},
  \bibinfo{year}{2014}.
\newblock \bibinfo{title}{A new locally-adaptive classification method lagma
  for large-scale land cover mapping using remote-sensing data}.
\newblock \bibinfo{journal}{Remote Sensing Letters} \bibinfo{volume}{5},
  \bibinfo{pages}{55--64}.
\newblock \DOIprefix\doi{10.1080/2150704X.2013.870675}.
\bibitem[{Beiranvand~Pour and Hashim(2014)}]{Pour2014}
\bibinfo{author}{Beiranvand~Pour, A.}, \bibinfo{author}{Hashim, M.},
  \bibinfo{year}{2014}.
\newblock \bibinfo{title}{{ASTER, ALI and Hyperion sensors data for
  lithological mapping and ore minerals exploration}}.
\newblock \bibinfo{journal}{SpringerPlus} \bibinfo{volume}{3}.
\newblock \DOIprefix\doi{10.1186/2193-1801-3-130}.
\bibitem[{Beiranvand~Pour and Hashim(2016)}]{Pour2016a}
\bibinfo{author}{Beiranvand~Pour, A.}, \bibinfo{author}{Hashim, M.},
  \bibinfo{year}{2016}.
\newblock \bibinfo{title}{Application of satellite remote sensing data for
  geological mapping in antarctic peninsula}, in: \bibinfo{booktitle}{37th
  Asian Conference on Remote Sensing, ACRS 2016}, pp.
  \bibinfo{pages}{171--177}.
\bibitem[{Beiranvand~Pour et~al.(2019a)Beiranvand~Pour, Hashim, Hong and
  Park}]{Pour2019}
\bibinfo{author}{Beiranvand~Pour, A.}, \bibinfo{author}{Hashim, M.},
  \bibinfo{author}{Hong, J.K.}, \bibinfo{author}{Park, Y.},
  \bibinfo{year}{2019}a.
\newblock \bibinfo{title}{{Lithological and alteration mineral mapping in
  poorly exposed lithologies using Landsat-8 and ASTER satellite data:
  North-eastern Graham Land, Antarctic Peninsula}}.
\newblock \bibinfo{journal}{Ore Geology Reviews} \bibinfo{volume}{108},
  \bibinfo{pages}{112--133}.
\newblock \DOIprefix\doi{10.1016/j.oregeorev.2017.07.018}.
\bibitem[{Beiranvand~Pour et~al.(2016)Beiranvand~Pour, Hashim, Makoundi and
  Zaw}]{Pour2016}
\bibinfo{author}{Beiranvand~Pour, A.}, \bibinfo{author}{Hashim, M.},
  \bibinfo{author}{Makoundi, C.}, \bibinfo{author}{Zaw, K.},
  \bibinfo{year}{2016}.
\newblock \bibinfo{title}{{Structural mapping of the Bentong-Raub Suture Zone
  using PALSAR remote sensing data, Peninsular Malaysia: Implications for
  sediment-hosted/orogenic gold mineral systems exploration}}.
\newblock \bibinfo{journal}{Resource Geology} \bibinfo{volume}{66},
  \bibinfo{pages}{368--385}.
\newblock \DOIprefix\doi{10.1111/rge.12105}.
\bibitem[{Beiranvand~Pour et~al.(2019b)Beiranvand~Pour, Park, Park, Hong,
  Muslim, L{\"{a}}ufer, Crispini, Pradhan, Zoheir, Rahmani, Hashim and
  Hossain}]{Pour2019a}
\bibinfo{author}{Beiranvand~Pour, A.}, \bibinfo{author}{Park, T.Y.},
  \bibinfo{author}{Park, Y.}, \bibinfo{author}{Hong, J.},
  \bibinfo{author}{Muslim, A.}, \bibinfo{author}{L{\"{a}}ufer, A.},
  \bibinfo{author}{Crispini, L.}, \bibinfo{author}{Pradhan, B.},
  \bibinfo{author}{Zoheir, B.}, \bibinfo{author}{Rahmani, O.},
  \bibinfo{author}{Hashim, M.}, \bibinfo{author}{Hossain, M.},
  \bibinfo{year}{2019}b.
\newblock \bibinfo{title}{Landsat-8, advanced spaceborne thermal emission and
  reflection radiometer, and worldview-3 multispectral satellite imagery for
  prospecting copper-gold mineralization in the northeastern inglefield mobile
  belt (imb), northwest greenland}.
\newblock \bibinfo{journal}{Remote Sensing} \bibinfo{volume}{11},
  \bibinfo{pages}{2430}.
\newblock \DOIprefix\doi{10.3390/rs11202430}.
\bibitem[{Beiranvand~Pour et~al.(2018)Beiranvand~Pour, Park, Park, Hong,
  Hashim, Woo and Ayoobi}]{Pour2018}
\bibinfo{author}{Beiranvand~Pour, A.}, \bibinfo{author}{Park, Y.},
  \bibinfo{author}{Park, T.Y.S.}, \bibinfo{author}{Hong, J.K.},
  \bibinfo{author}{Hashim, M.}, \bibinfo{author}{Woo, J.},
  \bibinfo{author}{Ayoobi, I.}, \bibinfo{year}{2018}.
\newblock \bibinfo{title}{{Regional geology mapping using satellite-based
  remote sensing approach in northern Victoria Land, Antarctica}}.
\newblock \bibinfo{journal}{Polar Science} \bibinfo{volume}{16},
  \bibinfo{pages}{23--46}.
\newblock \DOIprefix\doi{10.1016/j.polar.2018.02.004}.
\bibitem[{Belgiu and Drăguţ(2016)}]{Belgiu2016}
\bibinfo{author}{Belgiu, M.}, \bibinfo{author}{Drăguţ, L.},
  \bibinfo{year}{2016}.
\newblock \bibinfo{title}{{Random forest in remote sensing: A review of
  applications and future directions}}.
\newblock \bibinfo{journal}{ISPRS Journal of Photogrammetry and Remote Sensing}
  \bibinfo{volume}{114}, \bibinfo{pages}{24--31}.
\newblock \DOIprefix\doi{10.1016/j.isprsjprs.2016.01.011}.
\bibitem[{Benuwa et~al.(2016)Benuwa, Zhan, Ghansah, Wornyo and {Banaseka
  Kataka}}]{Benuwa2016}
\bibinfo{author}{Benuwa, B.B.}, \bibinfo{author}{Zhan, Y.Z.},
  \bibinfo{author}{Ghansah, B.}, \bibinfo{author}{Wornyo, D.K.},
  \bibinfo{author}{{Banaseka Kataka}, F.}, \bibinfo{year}{2016}.
\newblock \bibinfo{title}{{A review of deep machine learning}}.
\newblock \bibinfo{journal}{International Journal of Engineering Research in
  Africa} \bibinfo{volume}{24}, \bibinfo{pages}{124--136}.
\newblock \DOIprefix\doi{10.4028/www.scientific.net/JERA.24.124}.
\bibitem[{Biernacki et~al.(2000)Biernacki, Celeux and Govaert}]{Biernacki2000}
\bibinfo{author}{Biernacki, C.}, \bibinfo{author}{Celeux, G.},
  \bibinfo{author}{Govaert, G.}, \bibinfo{year}{2000}.
\newblock \bibinfo{title}{Assessing a mixture model for clustering with the
  integrated completed likelihood}.
\newblock \bibinfo{journal}{IEEE Transactions on Pattern Analysis and Machine
  Intelligence} \bibinfo{volume}{22}, \bibinfo{pages}{719--725}.
\newblock \DOIprefix\doi{10.1109/34.865189}.
\bibitem[{Bishta(2018)}]{Bishta2018}
\bibinfo{author}{Bishta, A.}, \bibinfo{year}{2018}.
\newblock \bibinfo{title}{Assessment of the reliability of supervised
  classifications of landsat-7, aster, and spot-5 multispectral data in rock
  unit discriminations of jabal daf-wadi fatima area, saudi arabia}.
\newblock \bibinfo{journal}{Arabian Journal of Geosciences}
  \bibinfo{volume}{11}.
\newblock \DOIprefix\doi{10.1007/s12517-018-4093-2}.
\bibitem[{Bishta and Sonbul(2021)}]{Bishta2021}
\bibinfo{author}{Bishta, A.}, \bibinfo{author}{Sonbul, A.},
  \bibinfo{year}{2021}.
\newblock \bibinfo{title}{Rock unit discriminations using image processing
  technique of ablah area, arabian shield, saudi arabia}.
\newblock \bibinfo{journal}{Journal of the Indian Society of Remote Sensing}
  \DOIprefix\doi{10.1007/s12524-021-01370-1}.
\bibitem[{Black et~al.(2016)Black, Riley, Ferrier, Fleming and
  Fretwell}]{Black2016}
\bibinfo{author}{Black, M.}, \bibinfo{author}{Riley, T.R.},
  \bibinfo{author}{Ferrier, G.}, \bibinfo{author}{Fleming, A.H.},
  \bibinfo{author}{Fretwell, P.T.}, \bibinfo{year}{2016}.
\newblock \bibinfo{title}{{Automated lithological mapping using airborne
  hyperspectral thermal infrared data: A case study from Anchorage Island,
  Antarctica}}.
\newblock \bibinfo{journal}{Remote Sensing of Environment}
  \bibinfo{volume}{176}, \bibinfo{pages}{225--241}.
\newblock \DOIprefix\doi{10.1016/j.rse.2016.01.022}.
\bibitem[{Bolouki et~al.(2020)Bolouki, Ramazi, Maghsoudi, {Beiranvand Pour} and
  Sohrabi}]{Bolouki2020}
\bibinfo{author}{Bolouki, S.M.}, \bibinfo{author}{Ramazi, H.R.},
  \bibinfo{author}{Maghsoudi, A.}, \bibinfo{author}{{Beiranvand Pour}, A.},
  \bibinfo{author}{Sohrabi, G.}, \bibinfo{year}{2020}.
\newblock \bibinfo{title}{{A remote sensing-based application of Bayesian
  networks for epithermal gold potential mapping in Ahar-Arasbaran Area, NW
  Iran}}.
\newblock \bibinfo{journal}{Remote Sensing} \bibinfo{volume}{12},
  \bibinfo{pages}{105}.
\newblock \DOIprefix\doi{10.3390/rs12010105}.
\bibitem[{Booysen et~al.(2020)Booysen, Jackisch, Lorenz, Zimmermann, Kirsch,
  Nex and Gloaguen}]{Booysen2020}
\bibinfo{author}{Booysen, R.}, \bibinfo{author}{Jackisch, R.},
  \bibinfo{author}{Lorenz, S.}, \bibinfo{author}{Zimmermann, R.},
  \bibinfo{author}{Kirsch, M.}, \bibinfo{author}{Nex, P.A.},
  \bibinfo{author}{Gloaguen, R.}, \bibinfo{year}{2020}.
\newblock \bibinfo{title}{{Detection of REEs with lightweight UAV-based
  hyperspectral imaging}}.
\newblock \bibinfo{journal}{Scientific Reports} \bibinfo{volume}{10},
  \bibinfo{pages}{17450}.
\newblock \DOIprefix\doi{10.1038/s41598-020-74422-0}.
\bibitem[{Borisova et~al.(2014)Borisova, Jelev, Atanassov, Koprinkova-Hristova
  and Alexiev}]{Borisova2014}
\bibinfo{author}{Borisova, D.}, \bibinfo{author}{Jelev, G.},
  \bibinfo{author}{Atanassov, V.}, \bibinfo{author}{Koprinkova-Hristova, P.},
  \bibinfo{author}{Alexiev, K.}, \bibinfo{year}{2014}.
\newblock \bibinfo{title}{{Algorithms for lineaments detection in processing of
  multispectral images}}, in: \bibinfo{editor}{Michel, U.},
  \bibinfo{editor}{Schulz, K.} (Eds.), \bibinfo{booktitle}{Earth Resources and
  Environmental Remote Sensing/GIS Applications V}, p. \bibinfo{pages}{92451L}.
\newblock \DOIprefix\doi{10.1117/12.2067245}.
\bibitem[{Boulila et~al.(2021)Boulila, Ghandorh, Khan, Ahmed and
  Ahmad}]{Boulila2021}
\bibinfo{author}{Boulila, W.}, \bibinfo{author}{Ghandorh, H.},
  \bibinfo{author}{Khan, M.A.}, \bibinfo{author}{Ahmed, F.},
  \bibinfo{author}{Ahmad, J.}, \bibinfo{year}{2021}.
\newblock \bibinfo{title}{A novel cnn-lstm-based approach to predict urban
  expansion}.
\newblock \bibinfo{journal}{Ecological Informatics} \bibinfo{volume}{64},
  \bibinfo{pages}{101325}.
\newblock \DOIprefix\doi{10.1016/j.ecoinf.2021.101325}.
\bibitem[{Breiman(1996)}]{Breiman1996}
\bibinfo{author}{Breiman, L.}, \bibinfo{year}{1996}.
\newblock \bibinfo{title}{{Bagging predictors}}.
\newblock \bibinfo{journal}{Machine Learning} \bibinfo{volume}{24},
  \bibinfo{pages}{123--140}.
\newblock \DOIprefix\doi{10.1007/BF00058655}.
\bibitem[{Brimhall et~al.(2005)Brimhall, Dilles and Proffett}]{Brimhall2005}
\bibinfo{author}{Brimhall, G.H.}, \bibinfo{author}{Dilles, J.H.},
  \bibinfo{author}{Proffett, J.M.}, \bibinfo{year}{2005}.
\newblock \bibinfo{title}{{The Role of Geologic Mapping in Mineral
  Exploration}}.
\newblock \DOIprefix\doi{10.5382/SP.12.11}.
\bibitem[{Buddenbaum et~al.(2012)Buddenbaum, Stern, Stellmes, Stoffels,
  Pueschel, Hill and Werner}]{Buddenbaum2012}
\bibinfo{author}{Buddenbaum, H.}, \bibinfo{author}{Stern, O.},
  \bibinfo{author}{Stellmes, M.}, \bibinfo{author}{Stoffels, J.},
  \bibinfo{author}{Pueschel, P.}, \bibinfo{author}{Hill, J.},
  \bibinfo{author}{Werner, W.}, \bibinfo{year}{2012}.
\newblock \bibinfo{title}{Field imaging spectroscopy of beech seedlings under
  dryness stress}.
\newblock \bibinfo{journal}{Remote Sensing} \bibinfo{volume}{4},
  \bibinfo{pages}{3721--3740}.
\newblock \DOIprefix\doi{10.3390/rs4123721}.
\bibitem[{Caggiano et~al.(2018)Caggiano, Angelone, Napolitano, Nele and
  Teti}]{Caggiano2018}
\bibinfo{author}{Caggiano, A.}, \bibinfo{author}{Angelone, R.},
  \bibinfo{author}{Napolitano, F.}, \bibinfo{author}{Nele, L.},
  \bibinfo{author}{Teti, R.}, \bibinfo{year}{2018}.
\newblock \bibinfo{title}{{Dimensionality reduction of sensorial features by
  principal component analysis for ANN machine learning in tool condition
  monitoring of CFRP drilling}}.
\newblock \bibinfo{journal}{Procedia CIRP} \bibinfo{volume}{78},
  \bibinfo{pages}{307--312}.
\newblock \DOIprefix\doi{10.1016/j.procir.2018.09.072}.
\bibitem[{{\c{C}}ığşar and {\"{U}}nal(2019)}]{Cgsar2019}
\bibinfo{author}{{\c{C}}ığşar, B.}, \bibinfo{author}{{\"{U}}nal, D.},
  \bibinfo{year}{2019}.
\newblock \bibinfo{title}{{Comparison of data mining classification algorithms
  determining the default risk}}.
\newblock \bibinfo{journal}{Scientific Programming} \bibinfo{volume}{2019},
  \bibinfo{pages}{1--8}.
\newblock \DOIprefix\doi{10.1155/2019/8706505}.
\bibitem[{Cardoso-Fernandes et~al.(2020a)Cardoso-Fernandes, Teodoro, Lima,
  Perrotta and Roda-Robles}]{Cardoso-Fernandes2020}
\bibinfo{author}{Cardoso-Fernandes, J.}, \bibinfo{author}{Teodoro, A.C.},
  \bibinfo{author}{Lima, A.}, \bibinfo{author}{Perrotta, M.},
  \bibinfo{author}{Roda-Robles, E.}, \bibinfo{year}{2020}a.
\newblock \bibinfo{title}{{Detecting lithium (Li) mineralizations from space:
  Current research and future perspectives}}.
\newblock \bibinfo{journal}{Applied Sciences} \bibinfo{volume}{10},
  \bibinfo{pages}{1785}.
\newblock \DOIprefix\doi{10.3390/app10051785}.
\bibitem[{Cardoso-Fernandes et~al.(2020b)Cardoso-Fernandes, Teodoro, Lima and
  Roda-Robles}]{Cardoso-Fernandes2020a}
\bibinfo{author}{Cardoso-Fernandes, J.}, \bibinfo{author}{Teodoro, A.C.},
  \bibinfo{author}{Lima, A.}, \bibinfo{author}{Roda-Robles, E.},
  \bibinfo{year}{2020}b.
\newblock \bibinfo{title}{{Semi-automatization of support vector machines to
  map lithium (Li) bearing pegmatites}}.
\newblock \bibinfo{journal}{Remote Sensing} \bibinfo{volume}{12},
  \bibinfo{pages}{2319}.
\newblock \DOIprefix\doi{10.3390/rs12142319}.
\bibitem[{Cardoso-Fernandes et~al.(2019)Cardoso-Fernandes, Teodoro, Lima and
  Roda-Robles}]{Cardoso-Fernandes2019}
\bibinfo{author}{Cardoso-Fernandes, J.}, \bibinfo{author}{Teodoro, A.C.M.},
  \bibinfo{author}{Lima, A.}, \bibinfo{author}{Roda-Robles, E.},
  \bibinfo{year}{2019}.
\newblock \bibinfo{title}{{Evaluating the performance of support vector
  machines (SVMs) and random forest (RF) in Li-pegmatite mapping: Preliminary
  results}}, in: \bibinfo{editor}{Schulz, K.}, \bibinfo{editor}{Nikolakopoulos,
  K.G.}, \bibinfo{editor}{Michel, U.} (Eds.), \bibinfo{booktitle}{Earth
  Resources and Environmental Remote Sensing/GIS Applications X},
  p.~\bibinfo{pages}{26}.
\newblock \DOIprefix\doi{10.1117/12.2532577}.
\bibitem[{Chakouri et~al.(2020)Chakouri, El~Harti, Lhissou, El~Hachimi and
  Jellouli}]{Chakouri2020}
\bibinfo{author}{Chakouri, M.}, \bibinfo{author}{El~Harti, A.},
  \bibinfo{author}{Lhissou, R.}, \bibinfo{author}{El~Hachimi, J.},
  \bibinfo{author}{Jellouli, A.}, \bibinfo{year}{2020}.
\newblock \bibinfo{title}{Geological and mineralogical mapping in moroccan
  central jebilet using multispectral and hyperspectral satellite data and
  machine learning}.
\newblock \bibinfo{journal}{International Journal of Advanced Trends in
  Computer Science and Engineering} \bibinfo{volume}{9},
  \bibinfo{pages}{5772--5783}.
\newblock \DOIprefix\doi{10.30534/ijatcse/2020/234942020}.
\bibitem[{Chandra et~al.(2019)Chandra, Azam, M{\"{u}}ller, Salles and
  Cripps}]{Chandra2019}
\bibinfo{author}{Chandra, R.}, \bibinfo{author}{Azam, D.},
  \bibinfo{author}{M{\"{u}}ller, R.D.}, \bibinfo{author}{Salles, T.},
  \bibinfo{author}{Cripps, S.}, \bibinfo{year}{2019}.
\newblock \bibinfo{title}{Bayeslands: A bayesian inference approach for
  parameter uncertainty quantification in badlands}.
\newblock \bibinfo{journal}{Computers {\&} Geosciences} \bibinfo{volume}{131},
  \bibinfo{pages}{89--101}.
\newblock \DOIprefix\doi{10.1016/j.cageo.2019.06.012}.
\bibitem[{Chandra et~al.(2021)Chandra, Cripps, Butterworth and
  M{\"{u}}ller}]{Chandra2021}
\bibinfo{author}{Chandra, R.}, \bibinfo{author}{Cripps, S.},
  \bibinfo{author}{Butterworth, N.}, \bibinfo{author}{M{\"{u}}ller, R.D.},
  \bibinfo{year}{2021}.
\newblock \bibinfo{title}{Precipitation reconstruction from climate-sensitive
  lithologies using bayesian machine learning}.
\newblock \bibinfo{journal}{Environmental Modelling {\&} Software}
  \bibinfo{volume}{139}, \bibinfo{pages}{105002}.
\newblock \DOIprefix\doi{10.1016/j.envsoft.2021.105002}.
\bibitem[{Chandra and Kapoor(2020)}]{Chandra2020}
\bibinfo{author}{Chandra, R.}, \bibinfo{author}{Kapoor, A.},
  \bibinfo{year}{2020}.
\newblock \bibinfo{title}{{Bayesian neural multi-source transfer learning}}.
\newblock \bibinfo{journal}{Neurocomputing} \bibinfo{volume}{378},
  \bibinfo{pages}{54--64}.
\newblock \DOIprefix\doi{10.1016/j.neucom.2019.10.042}.
\bibitem[{Cheng(1999)}]{Cheng1999}
\bibinfo{author}{Cheng, Q.}, \bibinfo{year}{1999}.
\newblock \bibinfo{title}{Spatial and scaling modelling for geochemical anomaly
  separation}.
\newblock \bibinfo{journal}{Journal of Geochemical Exploration}
  \bibinfo{volume}{65}, \bibinfo{pages}{175--194}.
\newblock \DOIprefix\doi{10.1016/S0375-6742(99)00028-X}.
\bibitem[{Cheng(2007)}]{Cheng2007}
\bibinfo{author}{Cheng, Q.}, \bibinfo{year}{2007}.
\newblock \bibinfo{title}{Mapping singularities with stream sediment
  geochemical data for prediction of undiscovered mineral deposits in gejiu,
  yunnan province, china}.
\newblock \bibinfo{journal}{Ore Geology Reviews} \bibinfo{volume}{32},
  \bibinfo{pages}{314--324}.
\newblock \DOIprefix\doi{10.1016/j.oregeorev.2006.10.002}.
\bibitem[{Chinkaka(2019)}]{Chinkaka2019}
\bibinfo{author}{Chinkaka, E.}, \bibinfo{year}{2019}.
\newblock \bibinfo{title}{{Integrating WorldView-3, ASTER and aeromagnetic data
  for lineament structural interpretation and tectonic evolution of the Haib
  Area, Namibia}}.
\newblock Ph.D. thesis. University of Twente.
\bibitem[{Chung et~al.(2020)Chung, Yu, Wang, Kim, Lee, Koh and Lee}]{Chung2020}
\bibinfo{author}{Chung, B.}, \bibinfo{author}{Yu, J.}, \bibinfo{author}{Wang,
  L.}, \bibinfo{author}{Kim, N.H.}, \bibinfo{author}{Lee, B.H.},
  \bibinfo{author}{Koh, S.}, \bibinfo{author}{Lee, S.}, \bibinfo{year}{2020}.
\newblock \bibinfo{title}{{Detection of magnesite and associated gangue
  minerals using hyperspectral remote sensing—A laboratory approach}}.
\newblock \bibinfo{journal}{Remote Sensing} \bibinfo{volume}{12},
  \bibinfo{pages}{1325}.
\newblock \DOIprefix\doi{10.3390/rs12081325}.
\bibitem[{Colomina and Molina(2014)}]{Colomina2014}
\bibinfo{author}{Colomina, I.}, \bibinfo{author}{Molina, P.},
  \bibinfo{year}{2014}.
\newblock \bibinfo{title}{{Unmanned aerial systems for photogrammetry and
  remote sensing: A review}}.
\newblock \bibinfo{journal}{ISPRS Journal of Photogrammetry and Remote Sensing}
  \bibinfo{volume}{92}, \bibinfo{pages}{79--97}.
\newblock \DOIprefix\doi{10.1016/j.isprsjprs.2014.02.013}.
\bibitem[{Comon(1994)}]{Comon1994}
\bibinfo{author}{Comon, P.}, \bibinfo{year}{1994}.
\newblock \bibinfo{title}{Independent component analysis, a new concept?}
\newblock \bibinfo{journal}{Signal Processing} \bibinfo{volume}{36},
  \bibinfo{pages}{287--314}.
\newblock \DOIprefix\doi{10.1016/0165-1684(94)90029-9}.
\bibitem[{Cortes and Vapnik(1995)}]{Cortes1995}
\bibinfo{author}{Cortes, C.}, \bibinfo{author}{Vapnik, V.},
  \bibinfo{year}{1995}.
\newblock \bibinfo{title}{{Support-vector networks}}.
\newblock \bibinfo{journal}{Machine Learning} \bibinfo{volume}{20},
  \bibinfo{pages}{273--297}.
\newblock \DOIprefix\doi{10.1007/BF00994018}.
\bibitem[{Cracknell and Reading(2013)}]{Cracknell2013}
\bibinfo{author}{Cracknell, M.J.}, \bibinfo{author}{Reading, A.M.},
  \bibinfo{year}{2013}.
\newblock \bibinfo{title}{{The upside of uncertainty: Identification of
  lithology contact zones from airborne geophysics and satellite data using
  random forests and support vector machines}}.
\newblock \bibinfo{journal}{Geophysics} \bibinfo{volume}{78},
  \bibinfo{pages}{WB113--WB126}.
\newblock \DOIprefix\doi{10.1190/geo2012-0411.1}.
\bibitem[{Cracknell and Reading(2014)}]{Cracknell2014}
\bibinfo{author}{Cracknell, M.J.}, \bibinfo{author}{Reading, A.M.},
  \bibinfo{year}{2014}.
\newblock \bibinfo{title}{{Geological mapping using remote sensing data: A
  comparison of five machine learning algorithms, their response to variations
  in the spatial distribution of training data and the use of explicit spatial
  information}}.
\newblock \bibinfo{journal}{Computers {\&} Geosciences} \bibinfo{volume}{63},
  \bibinfo{pages}{22--33}.
\newblock \DOIprefix\doi{10.1016/j.cageo.2013.10.008}.
\bibitem[{Dai et~al.(2017)Dai, Wang, Dai, Liu and Wu}]{Dai2017}
\bibinfo{author}{Dai, J.}, \bibinfo{author}{Wang, D.}, \bibinfo{author}{Dai,
  H.}, \bibinfo{author}{Liu, L.}, \bibinfo{author}{Wu, Y.},
  \bibinfo{year}{2017}.
\newblock \bibinfo{title}{Geological mapping and ore-prospecting study using
  remote sensing technology in jiajika area of western sichuan province}.
\newblock \bibinfo{journal}{Geology in China} \bibinfo{volume}{44},
  \bibinfo{pages}{389--398}.
\bibitem[{Diaz-Rodriguez et~al.(2021)Diaz-Rodriguez, {Dietmar M{\"{u}}ller} and
  Chandra}]{DiazRodriguez2021}
\bibinfo{author}{Diaz-Rodriguez, J.}, \bibinfo{author}{{Dietmar M{\"{u}}ller},
  R.}, \bibinfo{author}{Chandra, R.}, \bibinfo{year}{2021}.
\newblock \bibinfo{title}{Predicting the emplacement of cordilleran porphyry
  copper systems using a spatio-temporal machine learning model}.
\newblock \bibinfo{journal}{Ore Geology Reviews} ,
  \bibinfo{pages}{104300}\DOIprefix\doi{10.1016/j.oregeorev.2021.104300}.
\bibitem[{Dietterich(2002)}]{Dietterich2002}
\bibinfo{author}{Dietterich, T.}, \bibinfo{year}{2002}.
\newblock \bibinfo{title}{{Ensemble Learning, The Handbook of Brain Theory and
  Neural Networks}}.
\newblock \bibinfo{publisher}{MIT Press}.
\bibitem[{Ding et~al.(2017)Ding, Li, Xia, Wei, Zhang and Zhang}]{Chen2017}
\bibinfo{author}{Ding, C.}, \bibinfo{author}{Li, Y.}, \bibinfo{author}{Xia,
  Y.}, \bibinfo{author}{Wei, W.}, \bibinfo{author}{Zhang, L.},
  \bibinfo{author}{Zhang, Y.}, \bibinfo{year}{2017}.
\newblock \bibinfo{title}{{Convolutional neural networks based hyperspectral
  image classification method with adaptive kernels}}.
\newblock \bibinfo{journal}{Remote Sensing} \bibinfo{volume}{9},
  \bibinfo{pages}{618}.
\newblock \DOIprefix\doi{10.3390/rs9060618}.
\bibitem[{Drusch et~al.(2012)Drusch, {Del Bello}, Carlier, Colin, Fernandez,
  Gascon, Hoersch, Isola, Laberinti, Martimort, Meygret, Spoto, Sy, Marchese
  and Bargellini}]{Drusch2012}
\bibinfo{author}{Drusch, M.}, \bibinfo{author}{{Del Bello}, U.},
  \bibinfo{author}{Carlier, S.}, \bibinfo{author}{Colin, O.},
  \bibinfo{author}{Fernandez, V.}, \bibinfo{author}{Gascon, F.},
  \bibinfo{author}{Hoersch, B.}, \bibinfo{author}{Isola, C.},
  \bibinfo{author}{Laberinti, P.}, \bibinfo{author}{Martimort, P.},
  \bibinfo{author}{Meygret, A.}, \bibinfo{author}{Spoto, F.},
  \bibinfo{author}{Sy, O.}, \bibinfo{author}{Marchese, F.},
  \bibinfo{author}{Bargellini, P.}, \bibinfo{year}{2012}.
\newblock \bibinfo{title}{{Sentinel-2: ESA's optical high-resolution mission
  for GMES operational services}}.
\newblock \bibinfo{journal}{Remote Sensing of Environment}
  \bibinfo{volume}{120}, \bibinfo{pages}{25--36}.
\newblock \DOIprefix\doi{10.1016/j.rse.2011.11.026}.
\bibitem[{Ducart et~al.(2016)Ducart, Silva, Toledo and de~Assis}]{Ducart2016}
\bibinfo{author}{Ducart, D.F.}, \bibinfo{author}{Silva, A.M.},
  \bibinfo{author}{Toledo, C.L.B.}, \bibinfo{author}{de~Assis, L.M.},
  \bibinfo{year}{2016}.
\newblock \bibinfo{title}{{Mapping iron oxides with Landsat-8/OLI and
  EO-1/Hyperion imagery from the Serra Norte iron deposits in the Caraj{\'{a}}s
  Mineral Province, Brazil}}.
\newblock \bibinfo{journal}{Brazilian Journal of Geology} \bibinfo{volume}{46},
  \bibinfo{pages}{331--349}.
\newblock \DOIprefix\doi{10.1590/2317-4889201620160023}.
\bibitem[{{El Atillah} et~al.(2019){El Atillah}, {El Morjani} and
  Souhassou}]{ElAtillah2019}
\bibinfo{author}{{El Atillah}, A.}, \bibinfo{author}{{El Morjani}, Z.E.A.},
  \bibinfo{author}{Souhassou, M.}, \bibinfo{year}{2019}.
\newblock \bibinfo{title}{{Use of the Sentinel-2A multispectral image for
  litho-structural and alteration mapping in Al Glo'a Map Sheet (1/50,000) (Bou
  Azzer–El Graara Inlier, Central Anti-Atlas, Morocco)}}.
\newblock \bibinfo{journal}{Artificial Satellites} \bibinfo{volume}{54},
  \bibinfo{pages}{73--96}.
\newblock \DOIprefix\doi{10.2478/arsa-2019-0007}.
\bibitem[{Elman(1990)}]{Elman1990}
\bibinfo{author}{Elman, J.L.}, \bibinfo{year}{1990}.
\newblock \bibinfo{title}{Finding structure in time}.
\newblock \bibinfo{journal}{Cognitive Science} \bibinfo{volume}{14},
  \bibinfo{pages}{179--211}.
\newblock \DOIprefix\doi{10.1207/s15516709cog1402\_1}.
\bibitem[{Ester et~al.(1996)Ester, Kriegel, Sander and Xu}]{Ester1996}
\bibinfo{author}{Ester, M.}, \bibinfo{author}{Kriegel, H.P.},
  \bibinfo{author}{Sander, J.}, \bibinfo{author}{Xu, X.}, \bibinfo{year}{1996}.
\newblock \bibinfo{title}{A density-based algorithm for discovering clusters in
  large spatial databases with noise}, in: \bibinfo{booktitle}{Second
  International Conference on Knowledge Discovery and Data Mining}, pp.
  \bibinfo{pages}{226–--231}.
\bibitem[{Farahbakhsh et~al.(2020a)Farahbakhsh, Chandra, Olierook, Scalzo,
  Clark, Reddy and M{\"{u}}ller}]{Farahbakhsh2020a}
\bibinfo{author}{Farahbakhsh, E.}, \bibinfo{author}{Chandra, R.},
  \bibinfo{author}{Olierook, H.K.}, \bibinfo{author}{Scalzo, R.},
  \bibinfo{author}{Clark, C.}, \bibinfo{author}{Reddy, S.M.},
  \bibinfo{author}{M{\"{u}}ller, R.D.}, \bibinfo{year}{2020}a.
\newblock \bibinfo{title}{{Computer vision-based framework for extracting
  tectonic lineaments from optical remote sensing data}}.
\newblock \bibinfo{journal}{International Journal of Remote Sensing}
  \bibinfo{volume}{41}, \bibinfo{pages}{1760--1787}.
\newblock \DOIprefix\doi{10.1080/01431161.2019.1674462}.
\bibitem[{Farahbakhsh et~al.(2020b)Farahbakhsh, Hezarkhani, Eslamkish, Bahroudi
  and Chandra}]{Farahbakhsh2020}
\bibinfo{author}{Farahbakhsh, E.}, \bibinfo{author}{Hezarkhani, A.},
  \bibinfo{author}{Eslamkish, T.}, \bibinfo{author}{Bahroudi, A.},
  \bibinfo{author}{Chandra, R.}, \bibinfo{year}{2020}b.
\newblock \bibinfo{title}{Three-dimensional weights of evidence modelling of a
  deep-seated porphyry cu deposit}.
\newblock \bibinfo{journal}{Geochemistry: Exploration, Environment, Analysis}
  \bibinfo{volume}{20}, \bibinfo{pages}{480--495}.
\newblock \DOIprefix\doi{10.1144/geochem2020-038}.
\bibitem[{Farahbakhsh et~al.(2016)Farahbakhsh, Shirmard, Bahroudi and
  Eslamkish}]{Farahbakhsh2016}
\bibinfo{author}{Farahbakhsh, E.}, \bibinfo{author}{Shirmard, H.},
  \bibinfo{author}{Bahroudi, A.}, \bibinfo{author}{Eslamkish, T.},
  \bibinfo{year}{2016}.
\newblock \bibinfo{title}{{Fusing ASTER and QuickBird-2 satellite data for
  detailed investigation of porphyry copper deposits using PCA; case study of
  Naysian deposit, Iran}}.
\newblock \bibinfo{journal}{Journal of the Indian Society of Remote Sensing}
  \bibinfo{volume}{44}, \bibinfo{pages}{525--537}.
\newblock \DOIprefix\doi{10.1007/s12524-015-0516-7}.
\bibitem[{Fernandez and Steel(2000)}]{Fernandez2000}
\bibinfo{author}{Fernandez, C.}, \bibinfo{author}{Steel, M.F.},
  \bibinfo{year}{2000}.
\newblock \bibinfo{title}{Bayesian regression analysis with scale mixtures of
  normals}.
\newblock \bibinfo{journal}{Econometric Theory} \bibinfo{volume}{16},
  \bibinfo{pages}{80--101}.
\bibitem[{Ferrier et~al.(2002)Ferrier, White, Griffiths, Bryant and
  Stefouli}]{Ferrier2002}
\bibinfo{author}{Ferrier, G.}, \bibinfo{author}{White, K.},
  \bibinfo{author}{Griffiths, G.}, \bibinfo{author}{Bryant, R.},
  \bibinfo{author}{Stefouli, M.}, \bibinfo{year}{2002}.
\newblock \bibinfo{title}{The mapping of hydrothermal alteration zones on the
  island of lesvos, greece using an integrated remote sensing dataset}.
\newblock \bibinfo{journal}{International Journal of Remote Sensing}
  \bibinfo{volume}{23}, \bibinfo{pages}{341--356}.
\newblock \DOIprefix\doi{10.1080/01431160010003857}.
\bibitem[{Fisher et~al.(2012)Fisher, Amos, Bookhagen, Burbank and
  Godard}]{Fisher2012}
\bibinfo{author}{Fisher, G.B.}, \bibinfo{author}{Amos, C.B.},
  \bibinfo{author}{Bookhagen, B.}, \bibinfo{author}{Burbank, D.W.},
  \bibinfo{author}{Godard, V.}, \bibinfo{year}{2012}.
\newblock \bibinfo{title}{Channel widths, landslides, faults, and beyond: The
  new world order of high-spatial resolution google earth imagery in the study
  of earth surface processes}.
\newblock \bibinfo{journal}{Google Earth and Virtual Visualizations in
  Geoscience Education and Research: Geological Society of America Special
  Paper} \bibinfo{volume}{492}, \bibinfo{pages}{1--22}.
\newblock \DOIprefix\doi{10.1130/2012.2492(01)}.
\bibitem[{Fu and Aldrich(2019)}]{Fu2019}
\bibinfo{author}{Fu, Y.}, \bibinfo{author}{Aldrich, C.}, \bibinfo{year}{2019}.
\newblock \bibinfo{title}{{Flotation froth image recognition with convolutional
  neural networks}}.
\newblock \bibinfo{journal}{Minerals Engineering} \bibinfo{volume}{132},
  \bibinfo{pages}{183--190}.
\newblock \DOIprefix\doi{10.1016/j.mineng.2018.12.011}.
\bibitem[{Gad and Kusky(2007)}]{Gad2007}
\bibinfo{author}{Gad, S.}, \bibinfo{author}{Kusky, T.}, \bibinfo{year}{2007}.
\newblock \bibinfo{title}{{ASTER spectral ratioing for lithological mapping in
  the Arabian–Nubian shield, the Neoproterozoic Wadi Kid area, Sinai,
  Egypt}}.
\newblock \bibinfo{journal}{Gondwana Research} \bibinfo{volume}{11},
  \bibinfo{pages}{326--335}.
\newblock \DOIprefix\doi{10.1016/j.gr.2006.02.010}.
\bibitem[{Gahegan and Ehlers(2000)}]{Gahegan2000}
\bibinfo{author}{Gahegan, M.}, \bibinfo{author}{Ehlers, M.},
  \bibinfo{year}{2000}.
\newblock \bibinfo{title}{A framework for the modelling of uncertainty between
  remote sensing and geographic information systems}.
\newblock \bibinfo{journal}{ISPRS Journal of Photogrammetry and Remote Sensing}
  \bibinfo{volume}{55}, \bibinfo{pages}{176--188}.
\newblock \DOIprefix\doi{10.1016/S0924-2716(00)00018-6}.
\bibitem[{Gao et~al.(2021)Gao, Shi, Li and Wang}]{Gao2021}
\bibinfo{author}{Gao, Y.}, \bibinfo{author}{Shi, J.}, \bibinfo{author}{Li, J.},
  \bibinfo{author}{Wang, R.}, \bibinfo{year}{2021}.
\newblock \bibinfo{title}{{Remote sensing scene classification based on
  high-order graph convolutional network}}.
\newblock \bibinfo{journal}{European Journal of Remote Sensing} ,
  \bibinfo{pages}{1--15}\DOIprefix\doi{10.1080/22797254.2020.1868273}.
\bibitem[{Gasmi et~al.(2016)Gasmi, Gomez, Zouari, Masse and Ducrot}]{Gasmi2016}
\bibinfo{author}{Gasmi, A.}, \bibinfo{author}{Gomez, C.},
  \bibinfo{author}{Zouari, H.}, \bibinfo{author}{Masse, A.},
  \bibinfo{author}{Ducrot, D.}, \bibinfo{year}{2016}.
\newblock \bibinfo{title}{{PCA and SVM as geo-computational methods for
  geological mapping in the southern of Tunisia, using ASTER remote sensing
  data set}}.
\newblock \bibinfo{journal}{Arabian Journal of Geosciences}
  \bibinfo{volume}{9}, \bibinfo{pages}{753}.
\newblock \DOIprefix\doi{10.1007/s12517-016-2791-1}.
\bibitem[{Gemusse et~al.(2019)Gemusse, Lima and Teodoro}]{Gemusse2019}
\bibinfo{author}{Gemusse, U.}, \bibinfo{author}{Lima, A.},
  \bibinfo{author}{Teodoro, A.C.M.}, \bibinfo{year}{2019}.
\newblock \bibinfo{title}{{Comparing different techniques of satellite imagery
  classification to mineral mapping pegmatite of Muiane and Naipa:
  Mozambique}}, in: \bibinfo{editor}{Schulz, K.},
  \bibinfo{editor}{Nikolakopoulos, K.G.}, \bibinfo{editor}{Michel, U.} (Eds.),
  \bibinfo{booktitle}{Earth Resources and Environmental Remote Sensing/GIS
  Applications X}, p.~\bibinfo{pages}{49}.
\newblock \DOIprefix\doi{10.1117/12.2532570}.
\bibitem[{Genkin et~al.(2007)Genkin, Lewis and Madigan}]{Genkin2007}
\bibinfo{author}{Genkin, A.}, \bibinfo{author}{Lewis, D.D.},
  \bibinfo{author}{Madigan, D.}, \bibinfo{year}{2007}.
\newblock \bibinfo{title}{Large-scale bayesian logistic regression for text
  categorization}.
\newblock \bibinfo{journal}{Technometrics} \bibinfo{volume}{49},
  \bibinfo{pages}{291--304}.
\newblock \DOIprefix\doi{10.1198/004017007000000245}.
\bibitem[{Gewali et~al.(2018)Gewali, Monteiro and Saber}]{Gewali2018}
\bibinfo{author}{Gewali, U.B.}, \bibinfo{author}{Monteiro, S.T.},
  \bibinfo{author}{Saber, E.}, \bibinfo{year}{2018}.
\newblock \bibinfo{title}{{Machine learning based hyperspectral image analysis:
  A survey}} .
\bibitem[{Ghulam et~al.(2010)Ghulam, Amer, Candidate and Kusky}]{Ghulam2010}
\bibinfo{author}{Ghulam, A.}, \bibinfo{author}{Amer, R.},
  \bibinfo{author}{Candidate}, \bibinfo{author}{Kusky, T.},
  \bibinfo{year}{2010}.
\newblock \bibinfo{title}{{Mineral exploration and alteration zone mapping in
  Eastern Desert of Egypt using ASTER Data}}, in: \bibinfo{booktitle}{ASPRS
  Annual Conference}.
\bibitem[{Goetz and Rowan(1981)}]{Goetz1981}
\bibinfo{author}{Goetz, A.}, \bibinfo{author}{Rowan, L.}, \bibinfo{year}{1981}.
\newblock \bibinfo{title}{{Geologic remote sensing}}.
\newblock \bibinfo{journal}{Science} \bibinfo{volume}{211},
  \bibinfo{pages}{781--791}.
\newblock \DOIprefix\doi{10.1126/science.211.4484.781}.
\bibitem[{Gonzalez-Alvarez et~al.(2020)Gonzalez-Alvarez, Goncalves and
  Carranza}]{Gonzalez-Alvarez2020}
\bibinfo{author}{Gonzalez-Alvarez, I.}, \bibinfo{author}{Goncalves, M.},
  \bibinfo{author}{Carranza, E.}, \bibinfo{year}{2020}.
\newblock \bibinfo{title}{{Introduction to the special issue challenges for
  mineral exploration in the 21st century: Targeting mineral deposits under
  cover}}.
\newblock \bibinfo{journal}{Ore Geology Reviews} \bibinfo{volume}{126},
  \bibinfo{pages}{103785}.
\newblock \DOIprefix\doi{10.1016/j.oregeorev.2020.103785}.
\bibitem[{Good et~al.(2018)Good, Mallare, Payne, Wachs, Wamsley, Fasnacht, Bode
  and Padilla}]{Good2018}
\bibinfo{author}{Good, W.S.}, \bibinfo{author}{Mallare, B.},
  \bibinfo{author}{Payne, Z.}, \bibinfo{author}{Wachs, J.},
  \bibinfo{author}{Wamsley, C.}, \bibinfo{author}{Fasnacht, J.},
  \bibinfo{author}{Bode, R.}, \bibinfo{author}{Padilla, S.},
  \bibinfo{year}{2018}.
\newblock \bibinfo{title}{Demonstration of persistent, high resolution remote
  sensing from an advanced stratollite platform}, in: \bibinfo{booktitle}{AIAA
  Information Systems-AIAA Infotech at Aerospace}.
\newblock \DOIprefix\doi{10.2514/6.2018-0193}.
\bibitem[{Goodfellow et~al.(2014)Goodfellow, Pouget-Abadie, Mirza, Xu,
  Warde-Farley, Ozair, Courville and Bengio}]{Goodfellow2014}
\bibinfo{author}{Goodfellow, I.J.}, \bibinfo{author}{Pouget-Abadie, J.},
  \bibinfo{author}{Mirza, M.}, \bibinfo{author}{Xu, B.},
  \bibinfo{author}{Warde-Farley, D.}, \bibinfo{author}{Ozair, S.},
  \bibinfo{author}{Courville, A.}, \bibinfo{author}{Bengio, Y.},
  \bibinfo{year}{2014}.
\newblock \bibinfo{title}{{Generative Adversarial Networks}}.
\newblock \bibinfo{journal}{arXiv} \URLprefix
  \url{https://arxiv.org/abs/1406.2661}.
\bibitem[{Grebby et~al.(2012)Grebby, Cunningham, Naden and Tansey}]{Grebby2012}
\bibinfo{author}{Grebby, S.}, \bibinfo{author}{Cunningham, D.},
  \bibinfo{author}{Naden, J.}, \bibinfo{author}{Tansey, K.},
  \bibinfo{year}{2012}.
\newblock \bibinfo{title}{{Application of airborne LiDAR data and airborne
  multispectral imagery to structural mapping of the upper section of the
  Troodos ophiolite, Cyprus}}.
\newblock \bibinfo{journal}{International Journal of Earth Sciences}
  \bibinfo{volume}{101}, \bibinfo{pages}{1645--1660}.
\newblock \DOIprefix\doi{10.1007/s00531-011-0742-3}.
\bibitem[{Guan et~al.(2014)Guan, Yuan, Lee, Najeebullah and Rasel}]{Guan2014}
\bibinfo{author}{Guan, D.}, \bibinfo{author}{Yuan, W.}, \bibinfo{author}{Lee,
  Y.K.}, \bibinfo{author}{Najeebullah, K.}, \bibinfo{author}{Rasel, M.K.},
  \bibinfo{year}{2014}.
\newblock \bibinfo{title}{{A review of ensemble learning based feature
  selection}}.
\newblock \bibinfo{journal}{IETE Technical Review} \bibinfo{volume}{31},
  \bibinfo{pages}{190--198}.
\newblock \DOIprefix\doi{10.1080/02564602.2014.906859}.
\bibitem[{Guo et~al.(2016)Guo, Liu, Oerlemans, Lao, Wu and Lew}]{Guo2016}
\bibinfo{author}{Guo, Y.}, \bibinfo{author}{Liu, Y.},
  \bibinfo{author}{Oerlemans, A.}, \bibinfo{author}{Lao, S.},
  \bibinfo{author}{Wu, S.}, \bibinfo{author}{Lew, M.S.}, \bibinfo{year}{2016}.
\newblock \bibinfo{title}{Deep learning for visual understanding: A review}.
\newblock \bibinfo{journal}{Neurocomputing} \bibinfo{volume}{187},
  \bibinfo{pages}{27--48}.
\newblock \DOIprefix\doi{10.1016/j.neucom.2015.09.116}.
\bibitem[{Hamimi et~al.(2020)Hamimi, Hagag, Kamh and El-Araby}]{Hamimi2020}
\bibinfo{author}{Hamimi, Z.}, \bibinfo{author}{Hagag, W.},
  \bibinfo{author}{Kamh, S.}, \bibinfo{author}{El-Araby, A.},
  \bibinfo{year}{2020}.
\newblock \bibinfo{title}{{Application of remote-sensing techniques in
  geological and structural mapping of Atalla Shear Zone and Environs, Central
  Eastern Desert, Egypt}}.
\newblock \bibinfo{journal}{Arabian Journal of Geosciences}
  \bibinfo{volume}{13}, \bibinfo{pages}{414}.
\newblock \DOIprefix\doi{10.1007/s12517-020-05324-8}.
\bibitem[{Hamlin et~al.(2011)Hamlin, Green, Mouroulis, Eastwood, Wilson, Dudik
  and Paine}]{Hamlin2011}
\bibinfo{author}{Hamlin, L.}, \bibinfo{author}{Green, R.},
  \bibinfo{author}{Mouroulis, P.}, \bibinfo{author}{Eastwood, M.},
  \bibinfo{author}{Wilson, D.}, \bibinfo{author}{Dudik, M.},
  \bibinfo{author}{Paine, C.}, \bibinfo{year}{2011}.
\newblock \bibinfo{title}{{Imaging spectrometer science measurements for
  terrestrial ecology: AVIRIS and new developments}}, in:
  \bibinfo{booktitle}{Aerospace Conference}, pp. \bibinfo{pages}{1--7}.
\newblock \DOIprefix\doi{10.1109/AERO.2011.5747395}.
\bibitem[{Harbi and Madani(2014)}]{Harbi2014}
\bibinfo{author}{Harbi, H.}, \bibinfo{author}{Madani, A.},
  \bibinfo{year}{2014}.
\newblock \bibinfo{title}{Utilization of spot 5 data for mapping gold
  mineralized diorite–tonalite intrusion, bulghah gold mine area, saudi
  arabia}.
\newblock \bibinfo{journal}{Arabian Journal of Geosciences}
  \bibinfo{volume}{7}, \bibinfo{pages}{3829--3838}.
\newblock \DOIprefix\doi{10.1007/s12517-013-1035-x}.
\bibitem[{Harris et~al.(2011)Harris, Wickert, Lynds, Behnia, Rainbird, Grunsky,
  McGregor and Schetselaar}]{Harris2011}
\bibinfo{author}{Harris, J.}, \bibinfo{author}{Wickert, L.},
  \bibinfo{author}{Lynds, T.}, \bibinfo{author}{Behnia, P.},
  \bibinfo{author}{Rainbird, R.}, \bibinfo{author}{Grunsky, E.},
  \bibinfo{author}{McGregor, R.}, \bibinfo{author}{Schetselaar, E.},
  \bibinfo{year}{2011}.
\newblock \bibinfo{title}{Remote predictive mapping 3. optical remote sensing -
  a review for remote predictive geological mapping in northern canada}.
\newblock \bibinfo{journal}{Geoscience Canada} \bibinfo{volume}{38},
  \bibinfo{pages}{49--84}.
\bibitem[{Harvey and Fotopoulos(2016)}]{Harvey2016}
\bibinfo{author}{Harvey, A.}, \bibinfo{author}{Fotopoulos, G.},
  \bibinfo{year}{2016}.
\newblock \bibinfo{title}{Geological mapping using machine learning
  algorithms}.
\newblock \bibinfo{journal}{International Archives of the Photogrammetry,
  Remote Sensing and Spatial Information Sciences - ISPRS Archives}
  \bibinfo{volume}{41}, \bibinfo{pages}{423--430}.
\newblock \DOIprefix\doi{10.5194/isprsarchives-XLI-B8-423-2016}.
\bibitem[{Heincke et~al.(2019)Heincke, Jackisch, Saartenoja, Salmirinne, Rapp,
  Zimmermann, Pirttij{\"{a}}rvi, {Vest S{\"{o}}rensen}, Gloaguen, Ek,
  Bergstr{\"{o}}m, Karinen, Salehi, Madriz and Middleton}]{Heincke2019}
\bibinfo{author}{Heincke, B.}, \bibinfo{author}{Jackisch, R.},
  \bibinfo{author}{Saartenoja, A.}, \bibinfo{author}{Salmirinne, H.},
  \bibinfo{author}{Rapp, S.}, \bibinfo{author}{Zimmermann, R.},
  \bibinfo{author}{Pirttij{\"{a}}rvi, M.}, \bibinfo{author}{{Vest
  S{\"{o}}rensen}, E.}, \bibinfo{author}{Gloaguen, R.}, \bibinfo{author}{Ek,
  L.}, \bibinfo{author}{Bergstr{\"{o}}m, J.}, \bibinfo{author}{Karinen, A.},
  \bibinfo{author}{Salehi, S.}, \bibinfo{author}{Madriz, Y.},
  \bibinfo{author}{Middleton, M.}, \bibinfo{year}{2019}.
\newblock \bibinfo{title}{{Developing multi-sensor drones for geological
  mapping and mineral exploration: Setup and first results from the MULSEDRO
  project}}.
\newblock \bibinfo{journal}{Geological Survey of Denmark and Greenland
  Bulletin} \bibinfo{volume}{43}.
\newblock \DOIprefix\doi{10.34194/GEUSB-201943-03-02}.
\bibitem[{Heinrich and Candela(2014)}]{Heinrich2014}
\bibinfo{author}{Heinrich, C.}, \bibinfo{author}{Candela, P.},
  \bibinfo{year}{2014}.
\newblock \bibinfo{title}{{Fluids and ore formation in the Earth's crust}}, in:
  \bibinfo{booktitle}{Treatise on Geochemistry}. \bibinfo{publisher}{Elsevier},
  pp. \bibinfo{pages}{1--28}.
\newblock \DOIprefix\doi{10.1016/B978-0-08-095975-7.01101-3}.
\bibitem[{Ho(1998)}]{TinKamHo1998}
\bibinfo{author}{Ho, T.K.}, \bibinfo{year}{1998}.
\newblock \bibinfo{title}{{The random subspace method for constructing decision
  forests}}.
\newblock \bibinfo{journal}{IEEE Transactions on Pattern Analysis and Machine
  Intelligence} \bibinfo{volume}{20}, \bibinfo{pages}{832--844}.
\newblock \DOIprefix\doi{10.1109/34.709601}.
\bibitem[{Hoang and Koike(2017)}]{Hoang2017}
\bibinfo{author}{Hoang, N.T.}, \bibinfo{author}{Koike, K.},
  \bibinfo{year}{2017}.
\newblock \bibinfo{title}{{Transformation of Landsat imagery into
  pseudo-hyperspectral imagery by a multiple regression-based model with
  application to metal deposit-related minerals mapping}}.
\newblock \bibinfo{journal}{ISPRS Journal of Photogrammetry and Remote Sensing}
  \bibinfo{volume}{133}, \bibinfo{pages}{157--173}.
\newblock \DOIprefix\doi{10.1016/j.isprsjprs.2017.09.016}.
\bibitem[{Hochreiter and Schmidhuber(1997)}]{Hochreiter1997}
\bibinfo{author}{Hochreiter, S.}, \bibinfo{author}{Schmidhuber, J.},
  \bibinfo{year}{1997}.
\newblock \bibinfo{title}{Long short-term memory}.
\newblock \bibinfo{journal}{Neural Computation} \bibinfo{volume}{9},
  \bibinfo{pages}{1735--1780}.
\newblock \DOIprefix\doi{10.1162/neco.1997.9.8.1735}.
\bibitem[{Holloway and Mengersen(2018)}]{Holloway2018}
\bibinfo{author}{Holloway, J.}, \bibinfo{author}{Mengersen, K.},
  \bibinfo{year}{2018}.
\newblock \bibinfo{title}{{Statistical machine learning methods and remote
  sensing for sustainable development goals: A review}}.
\newblock \bibinfo{journal}{Remote Sensing} \bibinfo{volume}{10},
  \bibinfo{pages}{1365}.
\newblock \DOIprefix\doi{10.3390/rs10091365}.
\bibitem[{Hornik et~al.(1989)Hornik, Stinchcombe and White}]{Hornik1989}
\bibinfo{author}{Hornik, K.}, \bibinfo{author}{Stinchcombe, M.},
  \bibinfo{author}{White, H.}, \bibinfo{year}{1989}.
\newblock \bibinfo{title}{Multilayer feedforward networks are universal
  approximators}.
\newblock \bibinfo{journal}{Neural Networks} \bibinfo{volume}{2},
  \bibinfo{pages}{359--366}.
\newblock \DOIprefix\doi{10.1016/0893-6080(89)90020-8}.
\bibitem[{Ibrahim et~al.(2017)Ibrahim, Hariri, Abdullatif, Makkawi and
  Elzain}]{Ibrahim2017}
\bibinfo{author}{Ibrahim, M.I.}, \bibinfo{author}{Hariri, M.M.},
  \bibinfo{author}{Abdullatif, O.M.}, \bibinfo{author}{Makkawi, M.H.},
  \bibinfo{author}{Elzain, H.}, \bibinfo{year}{2017}.
\newblock \bibinfo{title}{Fractures system within qusaiba shale outcrop and its
  relationship to the lithological properties, qasim area, central saudi
  arabia}.
\newblock \bibinfo{journal}{Journal of African Earth Sciences}
  \bibinfo{volume}{133}, \bibinfo{pages}{104--122}.
\newblock \DOIprefix\doi{10.1016/j.jafrearsci.2017.05.011}.
\bibitem[{Iris et~al.(2019)Iris, Kroupnik, Lisle and Wierus}]{Iris2019}
\bibinfo{author}{Iris, S.}, \bibinfo{author}{Kroupnik, G.},
  \bibinfo{author}{Lisle, D.D.}, \bibinfo{author}{Wierus, M.},
  \bibinfo{year}{2019}.
\newblock \bibinfo{title}{{Radarsat constellation mission}}, in:
  \bibinfo{booktitle}{IEEE International Geoscience and Remote Sensing
  Symposium}, pp. \bibinfo{pages}{5749--5751}.
\newblock \DOIprefix\doi{10.1109/IGARSS.2019.8898976}.
\bibitem[{Ishidoshiro et~al.(2016)Ishidoshiro, Yamaguchi, Noda, Asano, Kondo,
  Kawakami, Mitsuishi and Nakamura}]{Ishidoshiro2016}
\bibinfo{author}{Ishidoshiro, N.}, \bibinfo{author}{Yamaguchi, Y.},
  \bibinfo{author}{Noda, S.}, \bibinfo{author}{Asano, Y.},
  \bibinfo{author}{Kondo, T.}, \bibinfo{author}{Kawakami, Y.},
  \bibinfo{author}{Mitsuishi, M.}, \bibinfo{author}{Nakamura, H.},
  \bibinfo{year}{2016}.
\newblock \bibinfo{title}{{Geological mapping by combining spectral unmixing
  and cluster analysis for hyperspectral data}}.
\newblock \bibinfo{journal}{ISPRS - International Archives of the
  Photogrammetry, Remote Sensing and Spatial Information Sciences}
  \bibinfo{volume}{XLI-B8}, \bibinfo{pages}{431--435}.
\newblock \DOIprefix\doi{10.5194/isprsarchives-XLI-B8-431-2016}.
\bibitem[{Jackisch et~al.(2020)Jackisch, Lorenz, Kirsch, Zimmermann, Tusa,
  Pirttij{\"a}rvi, Saartenoja, Ugalde, Madriz, Savolainen
  et~al.}]{Jackisch2020}
\bibinfo{author}{Jackisch, R.}, \bibinfo{author}{Lorenz, S.},
  \bibinfo{author}{Kirsch, M.}, \bibinfo{author}{Zimmermann, R.},
  \bibinfo{author}{Tusa, L.}, \bibinfo{author}{Pirttij{\"a}rvi, M.},
  \bibinfo{author}{Saartenoja, A.}, \bibinfo{author}{Ugalde, H.},
  \bibinfo{author}{Madriz, Y.}, \bibinfo{author}{Savolainen, M.}, et~al.,
  \bibinfo{year}{2020}.
\newblock \bibinfo{title}{Integrated geological and geophysical mapping of a
  carbonatite-hosting outcrop in siilinj{\"a}rvi, finland, using unmanned
  aerial systems}.
\newblock \bibinfo{journal}{Remote Sensing} \bibinfo{volume}{12},
  \bibinfo{pages}{2998}.
\newblock \DOIprefix\doi{10.3390/rs12182998}.
\bibitem[{Jakob et~al.(2017)Jakob, Zimmermann and Gloaguen}]{Jakob2017}
\bibinfo{author}{Jakob, S.}, \bibinfo{author}{Zimmermann, R.},
  \bibinfo{author}{Gloaguen, R.}, \bibinfo{year}{2017}.
\newblock \bibinfo{title}{{The need for accurate geometric and radiometric
  corrections of drone-borne hyperspectral data for mineral exploration:
  MEPHySTo—A toolbox for pre-processing drone-borne hyperspectral data}}.
\newblock \bibinfo{journal}{Remote Sensing} \bibinfo{volume}{9},
  \bibinfo{pages}{88}.
\newblock \DOIprefix\doi{10.3390/rs9010088}.
\bibitem[{Jiang et~al.(2019)Jiang, Wang, Yi, Wang, Lu and Jiang}]{Jiang2019}
\bibinfo{author}{Jiang, K.}, \bibinfo{author}{Wang, Z.}, \bibinfo{author}{Yi,
  P.}, \bibinfo{author}{Wang, G.}, \bibinfo{author}{Lu, T.},
  \bibinfo{author}{Jiang, J.}, \bibinfo{year}{2019}.
\newblock \bibinfo{title}{Edge-enhanced gan for remote sensing image
  superresolution}.
\newblock \bibinfo{journal}{IEEE Transactions on Geoscience and Remote Sensing}
  \bibinfo{volume}{57}, \bibinfo{pages}{5799--5812}.
\newblock \DOIprefix\doi{10.1109/TGRS.2019.2902431}.
\bibitem[{Johnson(1967)}]{Johnson1967}
\bibinfo{author}{Johnson, S.C.}, \bibinfo{year}{1967}.
\newblock \bibinfo{title}{Hierarchical clustering schemes}.
\newblock \bibinfo{journal}{Psychometrika} \bibinfo{volume}{32},
  \bibinfo{pages}{241--254}.
\newblock \DOIprefix\doi{10.1007/BF02289588}.
\bibitem[{Jung et~al.(2015)Jung, Vohland and Thiele-Bruhn}]{Jung2015}
\bibinfo{author}{Jung, A.}, \bibinfo{author}{Vohland, M.},
  \bibinfo{author}{Thiele-Bruhn, S.}, \bibinfo{year}{2015}.
\newblock \bibinfo{title}{Use of a portable camera for proximal soil sensing
  with hyperspectral image data}.
\newblock \bibinfo{journal}{Remote Sensing} \bibinfo{volume}{7},
  \bibinfo{pages}{11434--11448}.
\newblock \DOIprefix\doi{10.3390/rs70911434}.
\bibitem[{Karimi and Peng(2004)}]{Karimi2004}
\bibinfo{author}{Karimi, H.A.}, \bibinfo{author}{Peng, J.},
  \bibinfo{year}{2004}.
\newblock \bibinfo{title}{Using maximum likelihood (ml) and maximum a prior
  probability (map) in iterative self‐organizing data (isodata)}.
\newblock \bibinfo{journal}{Geocarto International} \bibinfo{volume}{19},
  \bibinfo{pages}{29--36}.
\newblock \DOIprefix\doi{10.1080/10106040408542296}.
\bibitem[{Kotsiantis(2007)}]{Kotsiantis2007}
\bibinfo{author}{Kotsiantis, S.}, \bibinfo{year}{2007}.
\newblock \bibinfo{title}{{Supervised machine learning: A review of
  classification techniques}}.
\newblock \bibinfo{journal}{Informatica (Slovenia)} \bibinfo{volume}{31},
  \bibinfo{pages}{249--268}.
\bibitem[{Kratt et~al.(2010)Kratt, Calvin and Coolbaugh}]{Kratt2010}
\bibinfo{author}{Kratt, C.}, \bibinfo{author}{Calvin, W.M.},
  \bibinfo{author}{Coolbaugh, M.F.}, \bibinfo{year}{2010}.
\newblock \bibinfo{title}{{Mineral mapping in the Pyramid Lake basin:
  Hydrothermal alteration, chemical precipitates and geothermal energy
  potential}}.
\newblock \bibinfo{journal}{Remote Sensing of Environment}
  \bibinfo{volume}{114}, \bibinfo{pages}{2297--2304}.
\newblock \DOIprefix\doi{10.1016/j.rse.2010.05.006}.
\bibitem[{Krupnik et~al.(2016)Krupnik, Khan, Okyay, Hartzell and
  Zhou}]{Krupnik2016}
\bibinfo{author}{Krupnik, D.}, \bibinfo{author}{Khan, S.},
  \bibinfo{author}{Okyay, U.}, \bibinfo{author}{Hartzell, P.},
  \bibinfo{author}{Zhou, H.W.}, \bibinfo{year}{2016}.
\newblock \bibinfo{title}{Study of upper albian rudist buildups in the edwards
  formation using ground-based hyperspectral imaging and terrestrial laser
  scanning}.
\newblock \bibinfo{journal}{Sedimentary Geology} \bibinfo{volume}{345},
  \bibinfo{pages}{154--167}.
\newblock \DOIprefix\doi{10.1016/j.sedgeo.2016.09.008}.
\bibitem[{Kruse and Perry(2013)}]{Kruse2013}
\bibinfo{author}{Kruse, F.}, \bibinfo{author}{Perry, S.}, \bibinfo{year}{2013}.
\newblock \bibinfo{title}{{Mineral mapping using simulated Worldview-3
  short-wave-infrared imagery}}.
\newblock \bibinfo{journal}{Remote Sensing} \bibinfo{volume}{5},
  \bibinfo{pages}{2688--2703}.
\newblock \DOIprefix\doi{10.3390/rs5062688}.
\bibitem[{Kruse et~al.(2015)Kruse, Baugh and Perry}]{Kruse2015}
\bibinfo{author}{Kruse, F.A.}, \bibinfo{author}{Baugh, W.M.},
  \bibinfo{author}{Perry, S.L.}, \bibinfo{year}{2015}.
\newblock \bibinfo{title}{{Validation of DigitalGlobe WorldView-3 Earth imaging
  satellite shortwave infrared bands for mineral mapping}}.
\newblock \bibinfo{journal}{Journal of Applied Remote Sensing}
  \bibinfo{volume}{9}, \bibinfo{pages}{096044}.
\newblock \DOIprefix\doi{10.1117/1.JRS.9.096044}.
\bibitem[{Kuhn et~al.(2018)Kuhn, Cracknell and Reading}]{Kuhn2018}
\bibinfo{author}{Kuhn, S.}, \bibinfo{author}{Cracknell, M.J.},
  \bibinfo{author}{Reading, A.M.}, \bibinfo{year}{2018}.
\newblock \bibinfo{title}{{Lithologic mapping using random forests applied to
  geophysical and remote-sensing data: A demonstration study from the Eastern
  Goldfields of Australia}}.
\newblock \bibinfo{journal}{GEOPHYSICS} \bibinfo{volume}{83},
  \bibinfo{pages}{B183--B193}.
\newblock \DOIprefix\doi{10.1190/geo2017-0590.1}.
\bibitem[{Kwak et~al.(2019)Kwak, Park, Park, Lee, Na, Ahn and Park}]{Kwak2019}
\bibinfo{author}{Kwak, G.H.}, \bibinfo{author}{Park, M.G.},
  \bibinfo{author}{Park, C.W.}, \bibinfo{author}{Lee, K.D.},
  \bibinfo{author}{Na, S.I.}, \bibinfo{author}{Ahn, H.Y.},
  \bibinfo{author}{Park, N.W.}, \bibinfo{year}{2019}.
\newblock \bibinfo{title}{Combining 2d cnn and bidirectional lstm to consider
  spatio-temporal features in crop classification}.
\newblock \bibinfo{journal}{Korean Journal of Remote Sensing}
  \bibinfo{volume}{35}, \bibinfo{pages}{681--692}.
\newblock \DOIprefix\doi{10.7780/kjrs.2019.35.5.1.5}.
\bibitem[{Latifovic et~al.(2018)Latifovic, Pouliot and
  Campbell}]{Latifovic2018}
\bibinfo{author}{Latifovic, R.}, \bibinfo{author}{Pouliot, D.},
  \bibinfo{author}{Campbell, J.}, \bibinfo{year}{2018}.
\newblock \bibinfo{title}{{Assessment of convolution neural networks for
  surficial geology mapping in the South Rae geological region, Northwest
  Territories, Canada}}.
\newblock \bibinfo{journal}{Remote Sensing} \bibinfo{volume}{10},
  \bibinfo{pages}{307}.
\newblock \DOIprefix\doi{10.3390/rs10020307}.
\bibitem[{Leach et~al.(2005)Leach, Sangster, Kelley, Large, Garven, Allen,
  Gutzmer and Walters}]{Leach2005}
\bibinfo{author}{Leach, D.}, \bibinfo{author}{Sangster, D.},
  \bibinfo{author}{Kelley, K.}, \bibinfo{author}{Large, R.},
  \bibinfo{author}{Garven, G.}, \bibinfo{author}{Allen, C.},
  \bibinfo{author}{Gutzmer, J.}, \bibinfo{author}{Walters, S.},
  \bibinfo{year}{2005}.
\newblock \bibinfo{title}{{Sediment-hosted lead-zinc deposits: A global
  perspective}}.
\newblock \bibinfo{journal}{Economic Geology} \bibinfo{volume}{100th Anni},
  \bibinfo{pages}{561--607}.
\bibitem[{LeCun et~al.(1988)LeCun, Touresky, Hinton and Sejnowski}]{Lecun1988}
\bibinfo{author}{LeCun, Y.}, \bibinfo{author}{Touresky, D.},
  \bibinfo{author}{Hinton, G.}, \bibinfo{author}{Sejnowski, T.},
  \bibinfo{year}{1988}.
\newblock \bibinfo{title}{A theoretical framework for back-propagation}, in:
  \bibinfo{booktitle}{Connectionist Models Summer School}, pp.
  \bibinfo{pages}{21--28}.
\bibitem[{Lecun and Yoshua(1998)}]{Lecun1998}
\bibinfo{author}{Lecun, Y.}, \bibinfo{author}{Yoshua, B.},
  \bibinfo{year}{1998}.
\newblock \bibinfo{title}{{Convolutional networks for images, speech, and time
  series In: The Handbook of Brain Theory and Neural Networks.}}
\bibitem[{Lee et~al.(2020)Lee, Churnside, Mao, Wu and Zibordi}]{Lee2020}
\bibinfo{author}{Lee, Z.}, \bibinfo{author}{Churnside, J.},
  \bibinfo{author}{Mao, Z.}, \bibinfo{author}{Wu, S.},
  \bibinfo{author}{Zibordi, G.}, \bibinfo{year}{2020}.
\newblock \bibinfo{title}{Active and passive optical remote sensing of the
  aquatic environment: Introduction to the feature issue}.
\newblock \bibinfo{journal}{Applied Optics} \bibinfo{volume}{59},
  \bibinfo{pages}{APS1--APS2}.
\newblock \DOIprefix\doi{10.1364/AO.392549}.
\bibitem[{Lek and Guegan(1999)}]{Lek1999}
\bibinfo{author}{Lek, S.}, \bibinfo{author}{Guegan, J.}, \bibinfo{year}{1999}.
\newblock \bibinfo{title}{{Artificial neural networks as a tool in ecological
  modelling, an introduction}}.
\newblock \bibinfo{journal}{Ecological Modelling} \bibinfo{volume}{120 (2-3)},
  \bibinfo{pages}{65--73}.
\bibitem[{Leverington(2010)}]{Leverington2010}
\bibinfo{author}{Leverington, D.W.}, \bibinfo{year}{2010}.
\newblock \bibinfo{title}{{Discrimination of sedimentary lithologies using
  Hyperion and Landsat Thematic Mapper data: A case study at Melville Island,
  Canadian High Arctic}}.
\newblock \bibinfo{journal}{International Journal of Remote Sensing}
  \bibinfo{volume}{31}, \bibinfo{pages}{233--260}.
\newblock \DOIprefix\doi{10.1080/01431160902882637}.
\bibitem[{Leverington and Moon(2012)}]{Leverington2012}
\bibinfo{author}{Leverington, D.W.}, \bibinfo{author}{Moon, W.M.},
  \bibinfo{year}{2012}.
\newblock \bibinfo{title}{{Landsat-TM-based discrimination of lithological
  units associated with the Purtuniq ophiolite, Quebec, Canada}}.
\newblock \bibinfo{journal}{Remote Sensing} \bibinfo{volume}{4},
  \bibinfo{pages}{1208--1231}.
\newblock \DOIprefix\doi{10.3390/rs4051208}.
\bibitem[{Li et~al.(2020a)Li, Hu, Li, Li, Du and Plaza}]{Li2020a}
\bibinfo{author}{Li, H.C.}, \bibinfo{author}{Hu, W.S.}, \bibinfo{author}{Li,
  W.}, \bibinfo{author}{Li, J.}, \bibinfo{author}{Du, Q.},
  \bibinfo{author}{Plaza, A.}, \bibinfo{year}{2020}a.
\newblock \bibinfo{title}{A$^{3}${CLNN}: Spatial, spectral and multiscale
  attention convlstm neural network for multisource remote sensing data
  classification}.
\newblock \bibinfo{journal}{IEEE Transactions on Neural Networks and Learning
  Systems} , \bibinfo{pages}{1--15}\DOIprefix\doi{10.1109/TNNLS.2020.3028945}.
\bibitem[{Li et~al.(2018)Li, Huang, Zhao, Qiu, Geng, Jia and Wang}]{Li2018}
\bibinfo{author}{Li, N.}, \bibinfo{author}{Huang, X.}, \bibinfo{author}{Zhao,
  H.}, \bibinfo{author}{Qiu, X.}, \bibinfo{author}{Geng, R.},
  \bibinfo{author}{Jia, X.}, \bibinfo{author}{Wang, D.}, \bibinfo{year}{2018}.
\newblock \bibinfo{title}{{Multiparameter optimization for mineral mapping
  using hyperspectral imagery}}.
\newblock \bibinfo{journal}{IEEE Journal of Selected Topics in Applied Earth
  Observations and Remote Sensing} \bibinfo{volume}{11},
  \bibinfo{pages}{1348--1357}.
\newblock \DOIprefix\doi{10.1109/JSTARS.2018.2814617}.
\bibitem[{Li et~al.(2020b)Li, Chen, Zhang, Zhang and Chen}]{Li2020}
\bibinfo{author}{Li, Y.}, \bibinfo{author}{Chen, R.}, \bibinfo{author}{Zhang,
  Y.}, \bibinfo{author}{Zhang, M.}, \bibinfo{author}{Chen, L.},
  \bibinfo{year}{2020}b.
\newblock \bibinfo{title}{{Multi-label remote sensing image scene
  classification by combining a convolutional neural network and a graph neural
  network}}.
\newblock \bibinfo{journal}{Remote Sensing} \bibinfo{volume}{12},
  \bibinfo{pages}{4003}.
\newblock \DOIprefix\doi{10.3390/rs12234003}.
\bibitem[{Lin et~al.(2020)Lin, Chen and Lu}]{Lin2020}
\bibinfo{author}{Lin, N.}, \bibinfo{author}{Chen, Y.}, \bibinfo{author}{Lu,
  L.}, \bibinfo{year}{2020}.
\newblock \bibinfo{title}{{Mineral potential mapping using a conjugate gradient
  logistic regression model}}.
\newblock \bibinfo{journal}{Natural Resources Research} \bibinfo{volume}{29},
  \bibinfo{pages}{173--188}.
\newblock \DOIprefix\doi{10.1007/s11053-019-09509-1}.
\bibitem[{Liu(2015)}]{Liu2015}
\bibinfo{author}{Liu, P.}, \bibinfo{year}{2015}.
\newblock \bibinfo{title}{{A survey of remote-sensing big data}}.
\newblock \bibinfo{journal}{Frontiers in Environmental Science}
  \bibinfo{volume}{3}.
\newblock \DOIprefix\doi{10.3389/fenvs.2015.00045}.
\bibitem[{Lloyd(1982)}]{Lloyd1982}
\bibinfo{author}{Lloyd, S.}, \bibinfo{year}{1982}.
\newblock \bibinfo{title}{{Least squares quantization in PCM}}.
\newblock \bibinfo{journal}{IEEE Transactions on Information Theory}
  \bibinfo{volume}{28}, \bibinfo{pages}{129--137}.
\newblock \DOIprefix\doi{10.1109/TIT.1982.1056489}.
\bibitem[{Loew et~al.(2017)Loew, Bell, Brocca, Bulgin, Burdanowitz, Calbet,
  Donner, Ghent, Gruber, Kaminski, Kinzel, Klepp, Lambert, Schaepman-Strub,
  Schröder and Verhoelst}]{Loew2017}
\bibinfo{author}{Loew, A.}, \bibinfo{author}{Bell, W.},
  \bibinfo{author}{Brocca, L.}, \bibinfo{author}{Bulgin, C.E.},
  \bibinfo{author}{Burdanowitz, J.}, \bibinfo{author}{Calbet, X.},
  \bibinfo{author}{Donner, R.V.}, \bibinfo{author}{Ghent, D.},
  \bibinfo{author}{Gruber, A.}, \bibinfo{author}{Kaminski, T.},
  \bibinfo{author}{Kinzel, J.}, \bibinfo{author}{Klepp, C.},
  \bibinfo{author}{Lambert, J.C.}, \bibinfo{author}{Schaepman-Strub, G.},
  \bibinfo{author}{Schröder, M.}, \bibinfo{author}{Verhoelst, T.},
  \bibinfo{year}{2017}.
\newblock \bibinfo{title}{Validation practices for satellite-based earth
  observation data across communities}.
\newblock \bibinfo{journal}{Reviews of Geophysics} \bibinfo{volume}{55},
  \bibinfo{pages}{779--817}.
\newblock \DOIprefix\doi{10.1002/2017RG000562}.
\bibitem[{Lorenz et~al.(2021)Lorenz, Ghamisi, Kirsch, Jackisch, Rasti and
  Gloaguen}]{Lorenz2021}
\bibinfo{author}{Lorenz, S.}, \bibinfo{author}{Ghamisi, P.},
  \bibinfo{author}{Kirsch, M.}, \bibinfo{author}{Jackisch, R.},
  \bibinfo{author}{Rasti, B.}, \bibinfo{author}{Gloaguen, R.},
  \bibinfo{year}{2021}.
\newblock \bibinfo{title}{{Feature extraction for hyperspectral mineral domain
  mapping: A test of conventional and innovative methods}}.
\newblock \bibinfo{journal}{Remote Sensing of Environment}
  \bibinfo{volume}{252}, \bibinfo{pages}{112129}.
\newblock \DOIprefix\doi{10.1016/j.rse.2020.112129}.
\bibitem[{Lowell and Guilbert(1970)}]{Lowell1970}
\bibinfo{author}{Lowell, J.D.}, \bibinfo{author}{Guilbert, J.M.},
  \bibinfo{year}{1970}.
\newblock \bibinfo{title}{{Lateral and vertical alteration-mineralization
  zoning in porphyry ore deposits}}.
\newblock \bibinfo{journal}{Economic Geology} \bibinfo{volume}{65},
  \bibinfo{pages}{373--408}.
\newblock \DOIprefix\doi{10.2113/gsecongeo.65.4.373}.
\bibitem[{Ma et~al.(2017)Ma, Xiao, Qin, Chen, Hu, Li and Zhao}]{Ma2017}
\bibinfo{author}{Ma, J.}, \bibinfo{author}{Xiao, X.}, \bibinfo{author}{Qin,
  Y.}, \bibinfo{author}{Chen, B.}, \bibinfo{author}{Hu, Y.},
  \bibinfo{author}{Li, X.}, \bibinfo{author}{Zhao, B.}, \bibinfo{year}{2017}.
\newblock \bibinfo{title}{{Estimating aboveground biomass of broadleaf,
  needleleaf, and mixed forests in northeastern China through analysis of 25-m
  ALOS/PALSAR mosaic data}}.
\newblock \bibinfo{journal}{Forest Ecology and Management}
  \bibinfo{volume}{389}, \bibinfo{pages}{199--210}.
\newblock \DOIprefix\doi{10.1016/j.foreco.2016.12.020}.
\bibitem[{Mahanta and Maiti(2018)}]{Mahanta2018}
\bibinfo{author}{Mahanta, P.}, \bibinfo{author}{Maiti, S.},
  \bibinfo{year}{2018}.
\newblock \bibinfo{title}{{Regional scale demarcation of alteration zone using
  ASTER imageries in South Purulia Shear Zone, East India: Implication for
  mineral exploration in vegetated regions}}.
\newblock \bibinfo{journal}{Ore Geology Reviews} \bibinfo{volume}{102},
  \bibinfo{pages}{846--861}.
\newblock \DOIprefix\doi{10.1016/j.oregeorev.2018.07.028}.
\bibitem[{Mansouri et~al.(2018)Mansouri, Feizi, {Jafari Rad} and
  Arian}]{Mansouri2018}
\bibinfo{author}{Mansouri, E.}, \bibinfo{author}{Feizi, F.},
  \bibinfo{author}{{Jafari Rad}, A.}, \bibinfo{author}{Arian, M.},
  \bibinfo{year}{2018}.
\newblock \bibinfo{title}{{Remote-sensing data processing with the multivariate
  regression analysis method for iron mineral resource potential mapping: A
  case study in the Sarvian area, Central Iran}}.
\newblock \bibinfo{journal}{Solid Earth} \bibinfo{volume}{9},
  \bibinfo{pages}{373--384}.
\newblock \DOIprefix\doi{10.5194/se-9-373-2018}.
\bibitem[{Maxwell et~al.(2018)Maxwell, Warner and Fang}]{Maxwell2018}
\bibinfo{author}{Maxwell, A.E.}, \bibinfo{author}{Warner, T.A.},
  \bibinfo{author}{Fang, F.}, \bibinfo{year}{2018}.
\newblock \bibinfo{title}{{Implementation of machine-learning classification in
  remote sensing: An applied review}}.
\newblock \bibinfo{journal}{International Journal of Remote Sensing}
  \bibinfo{volume}{39}, \bibinfo{pages}{2784--2817}.
\newblock \DOIprefix\doi{10.1080/01431161.2018.1433343}.
\bibitem[{van~der Meer(1994)}]{VanderMeer1994}
\bibinfo{author}{van~der Meer, F.}, \bibinfo{year}{1994}.
\newblock \bibinfo{title}{Extraction of mineral absorption features from
  high-spectralresolution data using non-parametric geostatistical techniques}.
\newblock \bibinfo{journal}{International Journal of Remote Sensing}
  \bibinfo{volume}{15}, \bibinfo{pages}{2193--2214}.
\newblock \DOIprefix\doi{10.1080/01431169408954238}.
\bibitem[{van~der Meer et~al.(2012)van~der Meer, van~der Werff, van Ruitenbeek,
  Hecker, Bakker, Noomen, van~der Meijde, Carranza, de~Smeth and
  Woldai}]{VanderMeer2012}
\bibinfo{author}{van~der Meer, F.D.}, \bibinfo{author}{van~der Werff, H.M.},
  \bibinfo{author}{van Ruitenbeek, F.J.}, \bibinfo{author}{Hecker, C.A.},
  \bibinfo{author}{Bakker, W.H.}, \bibinfo{author}{Noomen, M.F.},
  \bibinfo{author}{van~der Meijde, M.}, \bibinfo{author}{Carranza, E.J.M.},
  \bibinfo{author}{de~Smeth, J.B.}, \bibinfo{author}{Woldai, T.},
  \bibinfo{year}{2012}.
\newblock \bibinfo{title}{{Multi- and hyperspectral geologic remote sensing: A
  review}}.
\newblock \bibinfo{journal}{International Journal of Applied Earth Observation
  and Geoinformation} \bibinfo{volume}{14}, \bibinfo{pages}{112--128}.
\newblock \DOIprefix\doi{10.1016/j.jag.2011.08.002}.
\bibitem[{Metelka et~al.(2018)Metelka, Baratoux, Jessell, Barth, Je{\v{z}}ek
  and Naba}]{Metelka2018}
\bibinfo{author}{Metelka, V.}, \bibinfo{author}{Baratoux, L.},
  \bibinfo{author}{Jessell, M.W.}, \bibinfo{author}{Barth, A.},
  \bibinfo{author}{Je{\v{z}}ek, J.}, \bibinfo{author}{Naba, S.},
  \bibinfo{year}{2018}.
\newblock \bibinfo{title}{{Automated regolith landform mapping using airborne
  geophysics and remote sensing data, Burkina Faso, West Africa}}.
\newblock \bibinfo{journal}{Remote Sensing of Environment}
  \bibinfo{volume}{204}, \bibinfo{pages}{964--978}.
\newblock \DOIprefix\doi{10.1016/j.rse.2017.08.004}.
\bibitem[{Micklethwaite et~al.(2010)Micklethwaite, Sheldon and
  Baker}]{Micklethwaite2010}
\bibinfo{author}{Micklethwaite, S.}, \bibinfo{author}{Sheldon, H.A.},
  \bibinfo{author}{Baker, T.}, \bibinfo{year}{2010}.
\newblock \bibinfo{title}{{Active fault and shear processes and their
  implications for mineral deposit formation and discovery}}.
\newblock \bibinfo{journal}{Journal of Structural Geology}
  \bibinfo{volume}{32}, \bibinfo{pages}{151--165}.
\newblock \DOIprefix\doi{10.1016/j.jsg.2009.10.009}.
\bibitem[{Navarro et~al.(1997)Navarro, Frenk and White}]{Navarro1997}
\bibinfo{author}{Navarro, J.F.}, \bibinfo{author}{Frenk, C.S.},
  \bibinfo{author}{White, S.D.}, \bibinfo{year}{1997}.
\newblock \bibinfo{title}{A universal density profile from hierarchical
  clustering}.
\newblock \bibinfo{journal}{The Astrophysical Journal} \bibinfo{volume}{490},
  \bibinfo{pages}{493--508}.
\newblock \DOIprefix\doi{10.1086/304888}.
\bibitem[{Nielsen(2011)}]{Nielsen2011}
\bibinfo{author}{Nielsen, A.A.}, \bibinfo{year}{2011}.
\newblock \bibinfo{title}{Kernel maximum autocorrelation factor and minimum
  noise fraction transformations}.
\newblock \bibinfo{journal}{IEEE Transactions on Image Processing}
  \bibinfo{volume}{20}, \bibinfo{pages}{612--624}.
\newblock \DOIprefix\doi{10.1109/TIP.2010.2076296}.
\bibitem[{Ninomiya and Fu(2019)}]{Ninomiya2019}
\bibinfo{author}{Ninomiya, Y.}, \bibinfo{author}{Fu, B.}, \bibinfo{year}{2019}.
\newblock \bibinfo{title}{{Thermal infrared multispectral remote sensing of
  lithology and mineralogy based on spectral properties of materials}}.
\newblock \bibinfo{journal}{Ore Geology Reviews} \bibinfo{volume}{108},
  \bibinfo{pages}{54--72}.
\newblock \DOIprefix\doi{10.1016/j.oregeorev.2018.03.012}.
\bibitem[{Ninomiya et~al.(2005)Ninomiya, Fu and Cudahy}]{Ninomiya2005}
\bibinfo{author}{Ninomiya, Y.}, \bibinfo{author}{Fu, B.},
  \bibinfo{author}{Cudahy, T.J.}, \bibinfo{year}{2005}.
\newblock \bibinfo{title}{{Detecting lithology with Advanced Spaceborne Thermal
  Emission and Reflection Radiometer (ASTER) multispectral thermal infrared
  "radiance-at-sensor" data}}.
\newblock \bibinfo{journal}{Remote Sensing of Environment}
  \bibinfo{volume}{99}, \bibinfo{pages}{127--139}.
\newblock \DOIprefix\doi{10.1016/j.rse.2005.06.009}.
\bibitem[{Noori et~al.(2019)Noori, Beiranvand~Pour, Askari, Taghipour, Pradhan,
  Lee and Honarmand}]{Noori2019}
\bibinfo{author}{Noori, L.}, \bibinfo{author}{Beiranvand~Pour, A.},
  \bibinfo{author}{Askari, G.}, \bibinfo{author}{Taghipour, N.},
  \bibinfo{author}{Pradhan, B.}, \bibinfo{author}{Lee, C.W.},
  \bibinfo{author}{Honarmand, M.}, \bibinfo{year}{2019}.
\newblock \bibinfo{title}{{Comparison of different algorithms to map
  hydrothermal alteration zones using ASTER remote sensing data for
  polymetallic vein-type ore exploration: Toroud–Chahshirin magmatic belt
  (TCMB), North Iran}}.
\newblock \bibinfo{journal}{Remote Sensing} \bibinfo{volume}{11},
  \bibinfo{pages}{495}.
\newblock \DOIprefix\doi{10.3390/rs11050495}.
\bibitem[{Olierook et~al.(2021)Olierook, Scalzo, Kohn, Chandra, Farahbakhsh,
  Clark, Reddy and M{\"{u}}ller}]{Olierook2021}
\bibinfo{author}{Olierook, H.K.}, \bibinfo{author}{Scalzo, R.},
  \bibinfo{author}{Kohn, D.}, \bibinfo{author}{Chandra, R.},
  \bibinfo{author}{Farahbakhsh, E.}, \bibinfo{author}{Clark, C.},
  \bibinfo{author}{Reddy, S.M.}, \bibinfo{author}{M{\"{u}}ller, R.D.},
  \bibinfo{year}{2021}.
\newblock \bibinfo{title}{{Bayesian geological and geophysical data fusion for
  the construction and uncertainty quantification of 3D geological models}}.
\newblock \bibinfo{journal}{Geoscience Frontiers} \bibinfo{volume}{12},
  \bibinfo{pages}{479--493}.
\newblock \DOIprefix\doi{10.1016/j.gsf.2020.04.015}.
\bibitem[{Olson(1995)}]{Olson1995}
\bibinfo{author}{Olson, C.F.}, \bibinfo{year}{1995}.
\newblock \bibinfo{title}{Parallel algorithms for hierarchical clustering}.
\newblock \bibinfo{journal}{Parallel Computing} \bibinfo{volume}{21},
  \bibinfo{pages}{1313--1325}.
\newblock \DOIprefix\doi{10.1016/0167-8191(95)00017-I}.
\bibitem[{Othman and Gloaguen(2014)}]{Othman2014}
\bibinfo{author}{Othman, A.}, \bibinfo{author}{Gloaguen, R.},
  \bibinfo{year}{2014}.
\newblock \bibinfo{title}{{Improving lithological mapping by SVM classification
  of spectral and morphological features: The discovery of a new chromite body
  in the Mawat ophiolite complex (Kurdistan, NE Iraq)}}.
\newblock \bibinfo{journal}{Remote Sensing} \bibinfo{volume}{6},
  \bibinfo{pages}{6867--6896}.
\newblock \DOIprefix\doi{10.3390/rs6086867}.
\bibitem[{Pal et~al.(2020)Pal, Rasmussen and Porwal}]{Pal2020}
\bibinfo{author}{Pal, M.}, \bibinfo{author}{Rasmussen, T.},
  \bibinfo{author}{Porwal, A.}, \bibinfo{year}{2020}.
\newblock \bibinfo{title}{{Optimized lithological mapping from multispectral
  and hyperspectral remote sensing images using fused multi-classifiers}}.
\newblock \bibinfo{journal}{Remote Sensing} \bibinfo{volume}{12},
  \bibinfo{pages}{177}.
\newblock \DOIprefix\doi{10.3390/rs12010177}.
\bibitem[{Pan et~al.(2019)Pan, Ma, Mei, Dai, Fan, Tian and Ma}]{Pan2019}
\bibinfo{author}{Pan, E.}, \bibinfo{author}{Ma, Y.}, \bibinfo{author}{Mei, X.},
  \bibinfo{author}{Dai, X.}, \bibinfo{author}{Fan, F.}, \bibinfo{author}{Tian,
  X.}, \bibinfo{author}{Ma, J.}, \bibinfo{year}{2019}.
\newblock \bibinfo{title}{{Spectral-spatial classification of hyperspectral
  image based on a joint attention network}}, in: \bibinfo{booktitle}{IEEE
  International Geoscience and Remote Sensing Symposium}, pp.
  \bibinfo{pages}{413--416}.
\newblock \DOIprefix\doi{10.1109/IGARSS.2019.8898758}.
\bibitem[{Park and Choi(2020)}]{Park2020}
\bibinfo{author}{Park, S.}, \bibinfo{author}{Choi, Y.}, \bibinfo{year}{2020}.
\newblock \bibinfo{title}{{Applications of unmanned aerial vehicles in mining
  from exploration to reclamation: A review}}.
\newblock \bibinfo{journal}{Minerals} \bibinfo{volume}{10},
  \bibinfo{pages}{663}.
\newblock \DOIprefix\doi{10.3390/min10080663}.
\bibitem[{Pearlman et~al.(2003)Pearlman, Barry, Segal, Shepanski, Beiso and
  Carman}]{Pearlman2003}
\bibinfo{author}{Pearlman, J.}, \bibinfo{author}{Barry, P.},
  \bibinfo{author}{Segal, C.}, \bibinfo{author}{Shepanski, J.},
  \bibinfo{author}{Beiso, D.}, \bibinfo{author}{Carman, S.},
  \bibinfo{year}{2003}.
\newblock \bibinfo{title}{{Hyperion, a space-based imaging spectrometer}}.
\newblock \bibinfo{journal}{IEEE Transactions on Geoscience and Remote Sensing}
  \bibinfo{volume}{41}, \bibinfo{pages}{1160--1173}.
\newblock \DOIprefix\doi{10.1109/TGRS.2003.815018}.
\bibitem[{Peng and Gao(2013)}]{Peng2013}
\bibinfo{author}{Peng, G.}, \bibinfo{author}{Gao, G.}, \bibinfo{year}{2013}.
\newblock \bibinfo{title}{Remote sensing prospecting of pegmatite deposits in
  the azubai region, xinjiang}.
\newblock \bibinfo{journal}{Geotectonica et Metallogenia} \bibinfo{volume}{37},
  \bibinfo{pages}{109--117}.
\bibitem[{Porwal et~al.(2006)Porwal, Carranza and Hale}]{Porwal2006}
\bibinfo{author}{Porwal, A.}, \bibinfo{author}{Carranza, E.},
  \bibinfo{author}{Hale, M.}, \bibinfo{year}{2006}.
\newblock \bibinfo{title}{Bayesian network classifiers for mineral potential
  mapping}.
\newblock \bibinfo{journal}{Computers {\&} Geosciences} \bibinfo{volume}{32},
  \bibinfo{pages}{1--16}.
\newblock \DOIprefix\doi{10.1016/j.cageo.2005.03.018}.
\bibitem[{Prost(2014)}]{Prost2014}
\bibinfo{author}{Prost, G.L.}, \bibinfo{year}{2014}.
\newblock \bibinfo{title}{Remote Sensing for Geoscientists: Image Analysis and
  Integration}.
\newblock \bibinfo{edition}{3rd} ed., \bibinfo{publisher}{CRC Press}.
\bibitem[{Pu et~al.(2016)Pu, Gan, Henao, Yuan, Li, Stevens and Carin}]{Pu2016}
\bibinfo{author}{Pu, Y.}, \bibinfo{author}{Gan, Z.}, \bibinfo{author}{Henao,
  R.}, \bibinfo{author}{Yuan, X.}, \bibinfo{author}{Li, C.},
  \bibinfo{author}{Stevens, A.}, \bibinfo{author}{Carin, L.},
  \bibinfo{year}{2016}.
\newblock \bibinfo{title}{Variational autoencoder for deep learning of images,
  labels and captions}.
\newblock \href{http://arxiv.org/abs/1609.08976}{{\tt arXiv:1609.08976}}.
\bibitem[{Quinlan(1986)}]{Quinlan1986}
\bibinfo{author}{Quinlan, J.R.}, \bibinfo{year}{1986}.
\newblock \bibinfo{title}{Induction of decision trees}.
\newblock \bibinfo{journal}{Machine learning} \bibinfo{volume}{1},
  \bibinfo{pages}{81--106}.
\newblock \DOIprefix\doi{10.1007/BF00116251}.
\bibitem[{Quinlan(1987)}]{Quinlan1987}
\bibinfo{author}{Quinlan, J.R.}, \bibinfo{year}{1987}.
\newblock \bibinfo{title}{Simplifying decision trees}.
\newblock \bibinfo{journal}{International Journal of Man-Machine Studies}
  \bibinfo{volume}{27}, \bibinfo{pages}{221--234}.
\newblock \DOIprefix\doi{10.1016/S0020-7373(87)80053-6}.
\bibitem[{Radford et~al.(2018)Radford, Cracknell, Roach and
  Cumming}]{Radford2018}
\bibinfo{author}{Radford, D.D.}, \bibinfo{author}{Cracknell, M.J.},
  \bibinfo{author}{Roach, M.J.}, \bibinfo{author}{Cumming, G.V.},
  \bibinfo{year}{2018}.
\newblock \bibinfo{title}{{Geological mapping in Western Tasmania using radar
  and random forests}}.
\newblock \bibinfo{journal}{IEEE Journal of Selected Topics in Applied Earth
  Observations and Remote Sensing} \bibinfo{volume}{11},
  \bibinfo{pages}{3075--3087}.
\newblock \DOIprefix\doi{10.1109/JSTARS.2018.2855207}.
\bibitem[{Raharimahefa and Kusky(2009)}]{Raharimahefa2009}
\bibinfo{author}{Raharimahefa, T.}, \bibinfo{author}{Kusky, T.M.},
  \bibinfo{year}{2009}.
\newblock \bibinfo{title}{{Structural and remote sensing analysis of the
  Betsimisaraka Suture in northeastern Madagascar}}.
\newblock \bibinfo{journal}{Gondwana Research} \bibinfo{volume}{15},
  \bibinfo{pages}{14--27}.
\newblock \DOIprefix\doi{10.1016/j.gr.2008.07.004}.
\bibitem[{{Rajan Girija} and Mayappan(2019)}]{RajanGirija2019}
\bibinfo{author}{{Rajan Girija}, R.}, \bibinfo{author}{Mayappan, S.},
  \bibinfo{year}{2019}.
\newblock \bibinfo{title}{{Mapping of mineral resources and lithological units:
  A review of remote sensing techniques}}.
\newblock \bibinfo{journal}{International Journal of Image and Data Fusion}
  \bibinfo{volume}{10}, \bibinfo{pages}{79--106}.
\newblock \DOIprefix\doi{10.1080/19479832.2019.1589585}.
\bibitem[{Rajesh(2004)}]{Rajesh2004}
\bibinfo{author}{Rajesh, H.}, \bibinfo{year}{2004}.
\newblock \bibinfo{title}{{Application of remote sensing and GIS in mineral
  resource mapping-An overview}}.
\newblock \bibinfo{journal}{Journal of Mineralogical and Petrological Sciences}
  \bibinfo{volume}{99}, \bibinfo{pages}{83--103}.
\newblock \DOIprefix\doi{10.2465/jmps.99.83}.
\bibitem[{Ren et~al.(2019)Ren, Sun, Zhai and Liu}]{Ren2019}
\bibinfo{author}{Ren, Z.}, \bibinfo{author}{Sun, L.}, \bibinfo{author}{Zhai,
  Q.}, \bibinfo{author}{Liu, X.}, \bibinfo{year}{2019}.
\newblock \bibinfo{title}{{Mineral mapping with hyperspectral image based on an
  improved k-means clustering algorithm}}, in: \bibinfo{booktitle}{IEEE
  International Geoscience and Remote Sensing Symposium}, pp.
  \bibinfo{pages}{2989--2992}.
\newblock \DOIprefix\doi{10.1109/IGARSS.2019.8899113}.
\bibitem[{Rendu(1976)}]{Rendu1976}
\bibinfo{author}{Rendu, J.}, \bibinfo{year}{1976}.
\newblock \bibinfo{title}{Bayesian decision theory applied to mineral
  exploration and mine valuation}, in: \bibinfo{booktitle}{Advanced
  Geostatistics in the Mining Industry}, \bibinfo{publisher}{Springer
  Netherlands}. pp. \bibinfo{pages}{435--445}.
\bibitem[{Rezaei et~al.(2020)Rezaei, Hassani, Moarefvand and
  Golmohammadi}]{Rezaei2020}
\bibinfo{author}{Rezaei, A.}, \bibinfo{author}{Hassani, H.},
  \bibinfo{author}{Moarefvand, P.}, \bibinfo{author}{Golmohammadi, A.},
  \bibinfo{year}{2020}.
\newblock \bibinfo{title}{{Lithological mapping in Sangan region in Northeast
  Iran using ASTER satellite data and image processing methods}}.
\newblock \bibinfo{journal}{Geology, Ecology, and Landscapes}
  \bibinfo{volume}{4}, \bibinfo{pages}{59--70}.
\newblock \DOIprefix\doi{10.1080/24749508.2019.1585657}.
\bibitem[{Richards(2003)}]{Richards2003}
\bibinfo{author}{Richards, J.}, \bibinfo{year}{2003}.
\newblock \bibinfo{title}{{Tectono-Magmatic Precursors for Porphyry Cu-(Mo-Au)
  Deposit Formation}}.
\newblock \bibinfo{journal}{Economic Geology} \bibinfo{volume}{98},
  \bibinfo{pages}{1515--1533}.
\newblock \DOIprefix\doi{10.2113/98.8.1515}.
\bibitem[{Richards and Jia(2006)}]{Richards2006}
\bibinfo{author}{Richards, J.A.}, \bibinfo{author}{Jia, X.},
  \bibinfo{year}{2006}.
\newblock \bibinfo{title}{{Remote Sensing Digital Image Analysis}}.
\newblock \bibinfo{edition}{4th} ed., \bibinfo{publisher}{Springer:
  Berlin/Heidelberg, Germany}.
\bibitem[{Richards(2011)}]{Richards2011}
\bibinfo{author}{Richards, J.P.}, \bibinfo{year}{2011}.
\newblock \bibinfo{title}{{Magmatic to hydrothermal metal fluxes in convergent
  and collided margins}}.
\newblock \bibinfo{journal}{Ore Geology Reviews} \bibinfo{volume}{40},
  \bibinfo{pages}{1--26}.
\newblock \DOIprefix\doi{10.1016/j.oregeorev.2011.05.006}.
\bibitem[{Rigol-Sanchez et~al.(2003)Rigol-Sanchez, Chica-Olmo and
  Abarca-Hernandez}]{Rigol-Sanchez2003}
\bibinfo{author}{Rigol-Sanchez, J.}, \bibinfo{author}{Chica-Olmo, M.},
  \bibinfo{author}{Abarca-Hernandez, F.}, \bibinfo{year}{2003}.
\newblock \bibinfo{title}{{Artificial neural networks as a tool for mineral
  potential mapping with GIS}}.
\newblock \bibinfo{journal}{International Journal of Remote Sensing}
  \bibinfo{volume}{24}, \bibinfo{pages}{1151--1156}.
\newblock \DOIprefix\doi{10.1080/0143116021000031791}.
\bibitem[{Ripley and Li(2018)}]{Ripley2018}
\bibinfo{author}{Ripley, E.M.}, \bibinfo{author}{Li, C.}, \bibinfo{year}{2018}.
\newblock \bibinfo{title}{{Metallic ore deposits associated with mafic to
  ultramafic igneous rocks}}, in: \bibinfo{booktitle}{Processes and Ore
  Deposits of Ultramafic-Mafic Magmas through Space and Time}.
  \bibinfo{publisher}{Elsevier}, pp. \bibinfo{pages}{79--111}.
\newblock \DOIprefix\doi{10.1016/B978-0-12-811159-8.00004-4}.
\bibitem[{Rokach(2010)}]{Rokach2010}
\bibinfo{author}{Rokach, L.}, \bibinfo{year}{2010}.
\newblock \bibinfo{title}{{Ensemble-based classifiers}}.
\newblock \bibinfo{journal}{Artificial Intelligence Review}
  \bibinfo{volume}{33}, \bibinfo{pages}{1--39}.
\newblock \DOIprefix\doi{10.1007/s10462-009-9124-7}.
\bibitem[{Rowan and Mars(2003)}]{Rowan2003}
\bibinfo{author}{Rowan, L.C.}, \bibinfo{author}{Mars, J.C.},
  \bibinfo{year}{2003}.
\newblock \bibinfo{title}{{Lithologic mapping in the Mountain Pass, California
  area using Advanced Spaceborne Thermal Emission and Reflection Radiometer
  (ASTER) data}}.
\newblock \bibinfo{journal}{Remote Sensing of Environment}
  \bibinfo{volume}{84}, \bibinfo{pages}{350--366}.
\newblock \DOIprefix\doi{10.1016/S0034-4257(02)00127-X}.
\bibitem[{Rowan et~al.(2006)Rowan, Schmidt and Mars}]{Rowan2006}
\bibinfo{author}{Rowan, L.C.}, \bibinfo{author}{Schmidt, R.G.},
  \bibinfo{author}{Mars, J.C.}, \bibinfo{year}{2006}.
\newblock \bibinfo{title}{{Distribution of hydrothermally altered rocks in the
  Reko Diq, Pakistan mineralized area based on spectral analysis of ASTER
  data}}.
\newblock \bibinfo{journal}{Remote Sensing of Environment}
  \bibinfo{volume}{104}, \bibinfo{pages}{74--87}.
\newblock \DOIprefix\doi{10.1016/j.rse.2006.05.014}.
\bibitem[{Ruiz et~al.(2014)Ruiz, Mateos, Camps-Valls, Molina and
  Katsaggelos}]{Ruiz2014}
\bibinfo{author}{Ruiz, P.}, \bibinfo{author}{Mateos, J.},
  \bibinfo{author}{Camps-Valls, G.}, \bibinfo{author}{Molina, R.},
  \bibinfo{author}{Katsaggelos, A.K.}, \bibinfo{year}{2014}.
\newblock \bibinfo{title}{Bayesian active remote sensing image classification}.
\newblock \bibinfo{journal}{IEEE Transactions on Geoscience and Remote Sensing}
  \bibinfo{volume}{52}, \bibinfo{pages}{2186--2196}.
\newblock \DOIprefix\doi{10.1109/TGRS.2013.2258468}.
\bibitem[{Rumelhart et~al.(1986)Rumelhart, Hinton and Williams}]{Rumelhart1986}
\bibinfo{author}{Rumelhart, D.E.}, \bibinfo{author}{Hinton, G.E.},
  \bibinfo{author}{Williams, R.J.}, \bibinfo{year}{1986}.
\newblock \bibinfo{title}{Learning representations by back-propagating errors}.
\newblock \bibinfo{journal}{nature} \bibinfo{volume}{323},
  \bibinfo{pages}{533--536}.
\newblock \DOIprefix\doi{10.1038/323533a0}.
\bibitem[{Sabins(1999)}]{Sabins1999}
\bibinfo{author}{Sabins, F.F.}, \bibinfo{year}{1999}.
\newblock \bibinfo{title}{{Remote sensing for mineral exploration}}.
\newblock \bibinfo{journal}{Ore Geology Reviews} \bibinfo{volume}{14},
  \bibinfo{pages}{157--183}.
\newblock \DOIprefix\doi{10.1016/S0169-1368(99)00007-4}.
\bibitem[{Sagi and Rokach(2018)}]{Sagi2018}
\bibinfo{author}{Sagi, O.}, \bibinfo{author}{Rokach, L.}, \bibinfo{year}{2018}.
\newblock \bibinfo{title}{{Ensemble learning: A survey}}.
\newblock \bibinfo{journal}{Wiley Interdisciplinary Reviews: Data Mining and
  Knowledge Discovery} \bibinfo{volume}{8}.
\newblock \DOIprefix\doi{10.1002/widm.1249}.
\bibitem[{Salazar and Coffman(2020)}]{Salazar2020}
\bibinfo{author}{Salazar, S.E.}, \bibinfo{author}{Coffman, R.A.},
  \bibinfo{year}{2020}.
\newblock \bibinfo{title}{{Validation of a ground-based telescope-assisted
  hyperspectral remote sensor for soil measurements}}.
\newblock \bibinfo{journal}{Journal of Applied Remote Sensing}
  \bibinfo{volume}{14}, \bibinfo{pages}{1--13}.
\newblock \DOIprefix\doi{10.1117/1.JRS.14.027503}.
\bibitem[{Sang et~al.(2020)Sang, Xue, Ran, Li, Liu and Liu}]{Sang2020}
\bibinfo{author}{Sang, X.}, \bibinfo{author}{Xue, L.}, \bibinfo{author}{Ran,
  X.}, \bibinfo{author}{Li, X.}, \bibinfo{author}{Liu, J.},
  \bibinfo{author}{Liu, Z.}, \bibinfo{year}{2020}.
\newblock \bibinfo{title}{{Intelligent high-resolution geological mapping based
  on SLIC-CNN}}.
\newblock \bibinfo{journal}{ISPRS International Journal of Geo-Information}
  \bibinfo{volume}{9}, \bibinfo{pages}{99}.
\newblock \DOIprefix\doi{10.3390/ijgi9020099}.
\bibitem[{Schmidhuber(2015)}]{Schmidhuber2015}
\bibinfo{author}{Schmidhuber, J.}, \bibinfo{year}{2015}.
\newblock \bibinfo{title}{{Deep learning in neural networks: An overview}}.
\newblock \bibinfo{journal}{Neural Networks} \bibinfo{volume}{61},
  \bibinfo{pages}{85--117}.
\newblock \DOIprefix\doi{10.1016/j.neunet.2014.09.003}.
\bibitem[{Sekandari et~al.(2020)Sekandari, Masoumi, {Beiranvand Pour}, {M
  Muslim}, Rahmani, Hashim, Zoheir, Pradhan, Misra and
  Aminpour}]{Sekandari2020}
\bibinfo{author}{Sekandari, M.}, \bibinfo{author}{Masoumi, I.},
  \bibinfo{author}{{Beiranvand Pour}, A.}, \bibinfo{author}{{M Muslim}, A.},
  \bibinfo{author}{Rahmani, O.}, \bibinfo{author}{Hashim, M.},
  \bibinfo{author}{Zoheir, B.}, \bibinfo{author}{Pradhan, B.},
  \bibinfo{author}{Misra, A.}, \bibinfo{author}{Aminpour, S.M.},
  \bibinfo{year}{2020}.
\newblock \bibinfo{title}{{Application of Landsat-8, Sentinel-2, ASTER and
  WorldView-3 spectral imagery for exploration of carbonate-hosted Pb-Zn
  deposits in the Central Iranian Terrane (CIT)}}.
\newblock \bibinfo{journal}{Remote Sensing} \bibinfo{volume}{12},
  \bibinfo{pages}{1239}.
\newblock \DOIprefix\doi{10.3390/rs12081239}.
\bibitem[{Shamsolmoali et~al.(2020)Shamsolmoali, Zareapoor, Shen, Sadka and
  Yang}]{Shamsolmoali2020}
\bibinfo{author}{Shamsolmoali, P.}, \bibinfo{author}{Zareapoor, M.},
  \bibinfo{author}{Shen, L.}, \bibinfo{author}{Sadka, A.H.},
  \bibinfo{author}{Yang, J.}, \bibinfo{year}{2020}.
\newblock \bibinfo{title}{Imbalanced data learning by minority class
  augmentation using capsule adversarial networks}.
\newblock \bibinfo{journal}{Neurocomputing}
  \DOIprefix\doi{10.1016/j.neucom.2020.01.119}.
\bibitem[{Sheikhrahimi et~al.(2019)Sheikhrahimi, Beiranvand~Pour, Pradhan and
  Zoheir}]{Sheikhrahimi2019}
\bibinfo{author}{Sheikhrahimi, A.}, \bibinfo{author}{Beiranvand~Pour, A.},
  \bibinfo{author}{Pradhan, B.}, \bibinfo{author}{Zoheir, B.},
  \bibinfo{year}{2019}.
\newblock \bibinfo{title}{{Mapping hydrothermal alteration zones and lineaments
  associated with orogenic gold mineralization using ASTER data: A case study
  from the Sanandaj-Sirjan Zone, Iran}}.
\newblock \bibinfo{journal}{Advances in Space Research} \bibinfo{volume}{63},
  \bibinfo{pages}{3315--3332}.
\newblock \DOIprefix\doi{10.1016/j.asr.2019.01.035}.
\bibitem[{Shi et~al.(2017)Shi, Chen, Zhu, Sun, Luo, Gu and Zhou}]{Shi2017}
\bibinfo{author}{Shi, J.}, \bibinfo{author}{Chen, J.}, \bibinfo{author}{Zhu,
  J.}, \bibinfo{author}{Sun, S.}, \bibinfo{author}{Luo, Y.},
  \bibinfo{author}{Gu, Y.}, \bibinfo{author}{Zhou, Y.}, \bibinfo{year}{2017}.
\newblock \bibinfo{title}{Zhusuan: A library for bayesian deep learning}.
\newblock \href{http://arxiv.org/abs/1709.05870}{{\tt arXiv:1709.05870}}.
\bibitem[{Shi et~al.(2015)Shi, Chen, Wang, Yeung, kin Wong and chun
  Woo}]{Shi2015}
\bibinfo{author}{Shi, X.}, \bibinfo{author}{Chen, Z.}, \bibinfo{author}{Wang,
  H.}, \bibinfo{author}{Yeung, D.Y.}, \bibinfo{author}{kin Wong, W.},
  \bibinfo{author}{chun Woo, W.}, \bibinfo{year}{2015}.
\newblock \bibinfo{title}{Convolutional lstm network: A machine learning
  approach for precipitation nowcasting}.
\newblock \href{http://arxiv.org/abs/1506.04214}{{\tt arXiv:1506.04214}}.
\bibitem[{Shirmard et~al.(2020)Shirmard, Farahbakhsh, {Beiranvand Pour},
  Muslim, M{\"{u}}ller and Chandra}]{Shirmard2020}
\bibinfo{author}{Shirmard, H.}, \bibinfo{author}{Farahbakhsh, E.},
  \bibinfo{author}{{Beiranvand Pour}, A.}, \bibinfo{author}{Muslim, A.M.},
  \bibinfo{author}{M{\"{u}}ller, R.D.}, \bibinfo{author}{Chandra, R.},
  \bibinfo{year}{2020}.
\newblock \bibinfo{title}{{Integration of selective dimensionality reduction
  techniques for mineral exploration using ASTER satellite data}}.
\newblock \bibinfo{journal}{Remote Sensing} \bibinfo{volume}{12},
  \bibinfo{pages}{1261}.
\newblock \DOIprefix\doi{10.3390/rs12081261}.
\bibitem[{Shrestha and Mahmood(2019)}]{Shrestha2019}
\bibinfo{author}{Shrestha, A.}, \bibinfo{author}{Mahmood, A.},
  \bibinfo{year}{2019}.
\newblock \bibinfo{title}{Review of deep learning algorithms and
  architectures}.
\newblock \bibinfo{journal}{IEEE Access} \bibinfo{volume}{7},
  \bibinfo{pages}{53040--53065}.
\newblock \DOIprefix\doi{10.1109/ACCESS.2019.2912200}.
\bibitem[{Sillitoe(2010)}]{Sillitoe2010}
\bibinfo{author}{Sillitoe, R.H.}, \bibinfo{year}{2010}.
\newblock \bibinfo{title}{{Porphyry copper systems}}.
\newblock \bibinfo{journal}{Economic Geology} \bibinfo{volume}{105},
  \bibinfo{pages}{3--4}.
\bibitem[{Song et~al.(2020)Song, Wang, Yao, Fu, Hao and You}]{Song2020}
\bibinfo{author}{Song, K.}, \bibinfo{author}{Wang, E.}, \bibinfo{author}{Yao,
  Y.}, \bibinfo{author}{Fu, J.}, \bibinfo{author}{Hao, D.},
  \bibinfo{author}{You, X.}, \bibinfo{year}{2020}.
\newblock \bibinfo{title}{Spectral alteration zonation based on close range
  hyspex-320 m imaging spectroscopy: A case study in the gongchangling
  high-grade iron ore deposit, liaoning province, ne china}.
\newblock \bibinfo{journal}{Applied Sciences} \bibinfo{volume}{10},
  \bibinfo{pages}{8369}.
\newblock \DOIprefix\doi{10.3390/app10238369}.
\bibitem[{Song and Ying(2015)}]{Song2015}
\bibinfo{author}{Song, Y.Y.}, \bibinfo{author}{Ying, L.}, \bibinfo{year}{2015}.
\newblock \bibinfo{title}{Decision tree methods: Applications for
  classification and prediction}.
\newblock \bibinfo{journal}{Shanghai Archives of Psychiatry}
  \bibinfo{volume}{27}, \bibinfo{pages}{130--135}.
\newblock \DOIprefix\doi{10.11919/j.issn.1002-0829.215044}.
\bibitem[{Storvik et~al.(2005)Storvik, Fjortoft and Solberg}]{Storvik2005}
\bibinfo{author}{Storvik, G.}, \bibinfo{author}{Fjortoft, R.},
  \bibinfo{author}{Solberg, A.}, \bibinfo{year}{2005}.
\newblock \bibinfo{title}{A bayesian approach to classification of
  multiresolution remote sensing data}.
\newblock \bibinfo{journal}{IEEE Transactions on Geoscience and Remote Sensing}
  \bibinfo{volume}{43}, \bibinfo{pages}{539--547}.
\newblock \DOIprefix\doi{10.1109/TGRS.2004.841395}.
\bibitem[{Sudaryatno et~al.(2020)Sudaryatno, Widayani, Wibowo, Pramono, Afifah,
  Meikasari and Firdaus}]{Sudaryatno2020}
\bibinfo{author}{Sudaryatno}, \bibinfo{author}{Widayani, P.},
  \bibinfo{author}{Wibowo, T.W.}, \bibinfo{author}{Pramono, B.A.S.},
  \bibinfo{author}{Afifah, Z.N.}, \bibinfo{author}{Meikasari, A.D.},
  \bibinfo{author}{Firdaus, M.R.}, \bibinfo{year}{2020}.
\newblock \bibinfo{title}{{Multiple linear regression analysis of remote
  sensing data for determining vulnerability factors of landslide in
  Purworejo}}.
\newblock \bibinfo{journal}{IOP Conference Series: Earth and Environmental
  Science} \bibinfo{volume}{500}, \bibinfo{pages}{012046}.
\newblock \DOIprefix\doi{10.1088/1755-1315/500/1/012046}.
\bibitem[{Sun and Scanlon(2019)}]{Sun2019}
\bibinfo{author}{Sun, A.Y.}, \bibinfo{author}{Scanlon, B.R.},
  \bibinfo{year}{2019}.
\newblock \bibinfo{title}{{How can big data and machine learning benefit
  environment and water management: A survey of methods, applications, and
  future directions}}.
\newblock \bibinfo{journal}{Environmental Research Letters}
  \bibinfo{volume}{14}, \bibinfo{pages}{73001}.
\newblock \DOIprefix\doi{10.1088/1748-9326/ab1b7d}.
\bibitem[{Sydow(1977)}]{Sydow1977}
\bibinfo{author}{Sydow, A.}, \bibinfo{year}{1977}.
\newblock \bibinfo{title}{{Pattern Recognition Principles}}.
\newblock \bibinfo{journal}{Journal of Applied Mathematics and Mechanics}
  \bibinfo{volume}{57}, \bibinfo{pages}{353--354}.
\newblock \DOIprefix\doi{10.1002/zamm.19770570626}.
\bibitem[{Tagnon et~al.(2020)Tagnon, Assoma, Mangoua, Douagui, Kouam{\'{e}} and
  Savan{\'{e}}}]{Tagnon2020}
\bibinfo{author}{Tagnon, B.O.}, \bibinfo{author}{Assoma, V.T.},
  \bibinfo{author}{Mangoua, J.M.O.}, \bibinfo{author}{Douagui, A.G.},
  \bibinfo{author}{Kouam{\'{e}}, F.K.}, \bibinfo{author}{Savan{\'{e}}, I.},
  \bibinfo{year}{2020}.
\newblock \bibinfo{title}{{Contribution of SAR/RADARSAT-1 and ASAR/ENVISAT
  images to geological structural mapping and assessment of lineaments density
  in Divo-Oume Area (C{\^{o}}te d'Ivoire)}}.
\newblock \bibinfo{journal}{The Egyptian Journal of Remote Sensing and Space
  Science} \bibinfo{volume}{23}, \bibinfo{pages}{231--241}.
\newblock \DOIprefix\doi{10.1016/j.ejrs.2018.12.001}.
\bibitem[{Tahmasebi et~al.(2020)Tahmasebi, Kamrava, Bai and
  Sahimi}]{Tahmasebi2020}
\bibinfo{author}{Tahmasebi, P.}, \bibinfo{author}{Kamrava, S.},
  \bibinfo{author}{Bai, T.}, \bibinfo{author}{Sahimi, M.},
  \bibinfo{year}{2020}.
\newblock \bibinfo{title}{{Machine learning in geo- and environmental sciences:
  From small to large scale}}.
\newblock \bibinfo{journal}{Advances in Water Resources} \bibinfo{volume}{142},
  \bibinfo{pages}{103619}.
\newblock \DOIprefix\doi{10.1016/j.advwatres.2020.103619}.
\bibitem[{{Takodjou Wambo} et~al.(2020){Takodjou Wambo}, Beiranvand~Pour,
  Ganno, Asimow, Zoheir, Salles, Nzenti, Pradhan and
  Muslim}]{TakodjouWambo2020}
\bibinfo{author}{{Takodjou Wambo}, J.D.}, \bibinfo{author}{Beiranvand~Pour,
  A.}, \bibinfo{author}{Ganno, S.}, \bibinfo{author}{Asimow, P.D.},
  \bibinfo{author}{Zoheir, B.}, \bibinfo{author}{Salles, R.d.R.},
  \bibinfo{author}{Nzenti, J.P.}, \bibinfo{author}{Pradhan, B.},
  \bibinfo{author}{Muslim, A.M.}, \bibinfo{year}{2020}.
\newblock \bibinfo{title}{{Identifying high potential zones of gold
  mineralization in a sub-tropical region using Landsat-8 and ASTER remote
  sensing data: A case study of the Ngoura-Colomines goldfield, eastern
  Cameroon}}.
\newblock \bibinfo{journal}{Ore Geology Reviews} \bibinfo{volume}{122},
  \bibinfo{pages}{103530}.
\newblock \DOIprefix\doi{10.1016/j.oregeorev.2020.103530}.
\bibitem[{Tang et~al.(2019)Tang, Chen, Zhao, Huang and Luo}]{Tang2019}
\bibinfo{author}{Tang, T.}, \bibinfo{author}{Chen, S.}, \bibinfo{author}{Zhao,
  M.}, \bibinfo{author}{Huang, W.}, \bibinfo{author}{Luo, J.},
  \bibinfo{year}{2019}.
\newblock \bibinfo{title}{Very large-scale data classification based on k-means
  clustering and multi-kernel svm}.
\newblock \bibinfo{journal}{Soft Computing} \bibinfo{volume}{23},
  \bibinfo{pages}{3793--3801}.
\newblock \DOIprefix\doi{10.1007/s00500-018-3041-0}.
\bibitem[{Testa et~al.(2018)Testa, Villanueva, Cooke and Zhang}]{Testa2018}
\bibinfo{author}{Testa, F.}, \bibinfo{author}{Villanueva, C.},
  \bibinfo{author}{Cooke, D.}, \bibinfo{author}{Zhang, L.},
  \bibinfo{year}{2018}.
\newblock \bibinfo{title}{{Lithological and hydrothermal alteration mapping of
  epithermal, porphyry and tourmaline breccia districts in the Argentine Andes
  using ASTER imagery}}.
\newblock \bibinfo{journal}{Remote Sensing} \bibinfo{volume}{10},
  \bibinfo{pages}{203}.
\newblock \DOIprefix\doi{10.3390/rs10020203}.
\bibitem[{Tewksbury et~al.(2012)Tewksbury, Dokmak, Tarabees and
  Mansour}]{tewksbury2012}
\bibinfo{author}{Tewksbury, B.J.}, \bibinfo{author}{Dokmak, A.A.},
  \bibinfo{author}{Tarabees, E.A.}, \bibinfo{author}{Mansour, A.S.},
  \bibinfo{year}{2012}.
\newblock \bibinfo{title}{Google earth and geologic research in remote regions
  of the developing world: An example from the western desert of egypt}.
\newblock \bibinfo{journal}{Google Earth and Virtual Visualizations in
  Geoscience Education and Research: Geological Society of America Special
  Paper} \bibinfo{volume}{492}, \bibinfo{pages}{23--36}.
\newblock \DOIprefix\doi{10.1130/2012.2492(02)}.
\bibitem[{Toth and Jóźków(2016)}]{Toth2016}
\bibinfo{author}{Toth, C.}, \bibinfo{author}{Jóźków, G.},
  \bibinfo{year}{2016}.
\newblock \bibinfo{title}{Remote sensing platforms and sensors: A survey}.
\newblock \bibinfo{journal}{ISPRS Journal of Photogrammetry and Remote Sensing}
  \bibinfo{volume}{115}, \bibinfo{pages}{22--36}.
\newblock \DOIprefix\doi{10.1016/j.isprsjprs.2015.10.004}.
\bibitem[{Traore et~al.(2020)Traore, {Takodjou Wambo}, Ndepete, Tekin,
  Beiranvand~Pour and Muslim}]{Traore2020}
\bibinfo{author}{Traore, M.}, \bibinfo{author}{{Takodjou Wambo}, J.D.},
  \bibinfo{author}{Ndepete, C.P.}, \bibinfo{author}{Tekin, S.},
  \bibinfo{author}{Beiranvand~Pour, A.}, \bibinfo{author}{Muslim, A.M.},
  \bibinfo{year}{2020}.
\newblock \bibinfo{title}{{Lithological and alteration mineral mapping for
  alluvial gold exploration in the south east of Birao Area, Central African
  Republic using Landsat-8 operational land imager (OLI) data}}.
\newblock \bibinfo{journal}{Journal of African Earth Sciences}
  \bibinfo{volume}{170}, \bibinfo{pages}{103933}.
\newblock \DOIprefix\doi{10.1016/j.jafrearsci.2020.103933}.
\bibitem[{Tripathi et~al.(2020)Tripathi, Govil and Chattoraj}]{Tripathi2020}
\bibinfo{author}{Tripathi, M.K.}, \bibinfo{author}{Govil, H.},
  \bibinfo{author}{Chattoraj, S.}, \bibinfo{year}{2020}.
\newblock \bibinfo{title}{{Identification of hydrothermal altered/weathered and
  clay minerals through airborne AVIRIS-NG hyperspectral data in Jahajpur,
  India}}.
\newblock \bibinfo{journal}{Heliyon} \bibinfo{volume}{6},
  \bibinfo{pages}{e03487}.
\newblock \DOIprefix\doi{10.1016/j.heliyon.2020.e03487}.
\bibitem[{Usui and Okamoto(2010)}]{Usui2010}
\bibinfo{author}{Usui, A.}, \bibinfo{author}{Okamoto, N.},
  \bibinfo{year}{2010}.
\newblock \bibinfo{title}{Geophysical and geological exploration of cobalt-rich
  ferromanganese crusts: An attempt of small-scale mapping on a micronesian
  seamount}.
\newblock \bibinfo{journal}{Marine Georesources and Geotechnology}
  \bibinfo{volume}{28}, \bibinfo{pages}{192--206}.
\newblock \DOIprefix\doi{10.1080/10641190903521717}.
\bibitem[{Utgoff(1989)}]{Utgoff1989}
\bibinfo{author}{Utgoff, P.E.}, \bibinfo{year}{1989}.
\newblock \bibinfo{title}{Incremental induction of decision trees}.
\newblock \bibinfo{journal}{Machine Learning} \bibinfo{volume}{4},
  \bibinfo{pages}{161--186}.
\newblock \DOIprefix\doi{10.1023/A:1022699900025}.
\bibitem[{Varouchakis et~al.(2021)Varouchakis, Kamińska-Chuchmała, Kowalik,
  Spanoudaki and Graña}]{Varouchakis2021}
\bibinfo{author}{Varouchakis, E.}, \bibinfo{author}{Kamińska-Chuchmała, A.},
  \bibinfo{author}{Kowalik, G.}, \bibinfo{author}{Spanoudaki, K.},
  \bibinfo{author}{Graña, M.}, \bibinfo{year}{2021}.
\newblock \bibinfo{title}{Combining geostatistics and remote sensing data to
  improve spatiotemporal analysis of precipitation}.
\newblock \bibinfo{journal}{Sensors} \bibinfo{volume}{21}.
\newblock \DOIprefix\doi{10.3390/s21093132}.
\bibitem[{Velliangiri et~al.(2019)Velliangiri, Alagumuthukrishnan and
  Thankumar~joseph}]{Velliangiri2019}
\bibinfo{author}{Velliangiri, S.}, \bibinfo{author}{Alagumuthukrishnan, S.},
  \bibinfo{author}{Thankumar~joseph, S.I.}, \bibinfo{year}{2019}.
\newblock \bibinfo{title}{{A review of dimensionality reduction techniques for
  efficient computation}}.
\newblock \bibinfo{journal}{Procedia Computer Science} \bibinfo{volume}{165},
  \bibinfo{pages}{104--111}.
\newblock \DOIprefix\doi{10.1016/j.procs.2020.01.079}.
\bibitem[{Von~Luxburg(2007)}]{VonLuxburg2007}
\bibinfo{author}{Von~Luxburg, U.}, \bibinfo{year}{2007}.
\newblock \bibinfo{title}{A tutorial on spectral clustering}.
\newblock \bibinfo{journal}{Statistics and computing} \bibinfo{volume}{17},
  \bibinfo{pages}{395--416}.
\newblock \DOIprefix\doi{10.1007/s11222-007-9033-z}.
\bibitem[{Wang et~al.(2020a)Wang, Peng, Jiang and Liu}]{Wang2020}
\bibinfo{author}{Wang, B.}, \bibinfo{author}{Peng, X.}, \bibinfo{author}{Jiang,
  M.}, \bibinfo{author}{Liu, D.}, \bibinfo{year}{2020}a.
\newblock \bibinfo{title}{{Real-time fault detection for UAV based on model
  acceleration engine}}.
\newblock \bibinfo{journal}{IEEE Transactions on Instrumentation and
  Measurement} \bibinfo{volume}{69}, \bibinfo{pages}{9505--9516}.
\newblock \DOIprefix\doi{10.1109/TIM.2020.3001659}.
\bibitem[{Wang et~al.(2010)Wang, Yan, Zhang and Song}]{Wang2010}
\bibinfo{author}{Wang, G.}, \bibinfo{author}{Yan, C.}, \bibinfo{author}{Zhang,
  S.}, \bibinfo{author}{Song, Y.}, \bibinfo{year}{2010}.
\newblock \bibinfo{title}{{Probabilistic neural networks and fractal method
  applied to mineral potential mapping in Luanchuan region, Henan Province,
  China}}, in: \bibinfo{booktitle}{Sixth International Conference on Natural
  Computation}, pp. \bibinfo{pages}{1003--1007}.
\newblock \DOIprefix\doi{10.1109/ICNC.2010.5582906}.
\bibitem[{Wang and Yeung(2020)}]{Wang2020b}
\bibinfo{author}{Wang, H.}, \bibinfo{author}{Yeung, D.Y.},
  \bibinfo{year}{2020}.
\newblock \bibinfo{title}{A survey on bayesian deep learning}.
\newblock \bibinfo{journal}{ACM Computing Surveys} \bibinfo{volume}{53},
  \bibinfo{pages}{1--38}.
\newblock \DOIprefix\doi{10.1145/3409383}.
\bibitem[{Wang et~al.(2016)Wang, Li, Ge, Jin, Ma, Liu, Wen and Liu}]{Wang2016}
\bibinfo{author}{Wang, S.}, \bibinfo{author}{Li, X.}, \bibinfo{author}{Ge, Y.},
  \bibinfo{author}{Jin, R.}, \bibinfo{author}{Ma, M.}, \bibinfo{author}{Liu,
  Q.}, \bibinfo{author}{Wen, J.}, \bibinfo{author}{Liu, S.},
  \bibinfo{year}{2016}.
\newblock \bibinfo{title}{Validation of regional-scale remote sensing products
  in china: From site to network}.
\newblock \bibinfo{journal}{Remote Sensing} \bibinfo{volume}{8},
  \bibinfo{pages}{980}.
\newblock \DOIprefix\doi{10.3390/rs8120980}.
\bibitem[{Wang et~al.(2014)Wang, Huang, Wang and Wang}]{Wang2014}
\bibinfo{author}{Wang, W.}, \bibinfo{author}{Huang, Y.}, \bibinfo{author}{Wang,
  Y.}, \bibinfo{author}{Wang, L.}, \bibinfo{year}{2014}.
\newblock \bibinfo{title}{Generalized autoencoder: A neural network framework
  for dimensionality reduction}, in: \bibinfo{booktitle}{IEEE Conference on
  Computer Vision and Pattern Recognition (CVPR) Workshops}, pp.
  \bibinfo{pages}{490--497}.
\bibitem[{Wang and Jiang(2019)}]{Wang2019}
\bibinfo{author}{Wang, Y.}, \bibinfo{author}{Jiang, Y.}, \bibinfo{year}{2019}.
\newblock \bibinfo{title}{{A weighted minimum distance classifier based on
  relative offset}}, in: \bibinfo{booktitle}{Fourth International Conference on
  Cloud Computing and Big Data Analysis}, pp. \bibinfo{pages}{343--347}.
\newblock \DOIprefix\doi{10.1109/ICCCBDA.2019.8725734}.
\bibitem[{Wang and Zheng(2010)}]{Wang2010a}
\bibinfo{author}{Wang, Z.}, \bibinfo{author}{Zheng, C.}, \bibinfo{year}{2010}.
\newblock \bibinfo{title}{{Rocks/minerals information extraction from EO-1
  Hyperion data base on SVM}}, in: \bibinfo{booktitle}{International Conference
  on Intelligent Computation Technology and Automation}, pp.
  \bibinfo{pages}{229--232}.
\newblock \DOIprefix\doi{10.1109/ICICTA.2010.341}.
\bibitem[{Wang et~al.(2020b)Wang, Zuo and Dong}]{Wang2020a}
\bibinfo{author}{Wang, Z.}, \bibinfo{author}{Zuo, R.}, \bibinfo{author}{Dong,
  Y.}, \bibinfo{year}{2020}b.
\newblock \bibinfo{title}{{Mapping of Himalaya leucogranites based on ASTER and
  Sentinel-2A datasets using a hybrid method of metric learning and random
  forest}}.
\newblock \bibinfo{journal}{IEEE Journal of Selected Topics in Applied Earth
  Observations and Remote Sensing} \bibinfo{volume}{13},
  \bibinfo{pages}{1925--1936}.
\newblock \DOIprefix\doi{10.1109/JSTARS.2020.2989509}.
\bibitem[{Werbos(1990)}]{Werbos1990}
\bibinfo{author}{Werbos, P.J.}, \bibinfo{year}{1990}.
\newblock \bibinfo{title}{Backpropagation through time: what it does and how to
  do it}.
\newblock \bibinfo{journal}{Proceedings of the IEEE} \bibinfo{volume}{78},
  \bibinfo{pages}{1550--1560}.
\newblock \DOIprefix\doi{10.1109/5.58337}.
\bibitem[{Wold et~al.(1987)Wold, Esbensen and Geladi}]{Wold1987}
\bibinfo{author}{Wold, S.}, \bibinfo{author}{Esbensen, K.},
  \bibinfo{author}{Geladi, P.}, \bibinfo{year}{1987}.
\newblock \bibinfo{title}{Principal component analysis}.
\newblock \bibinfo{journal}{Chemometrics and Intelligent Laboratory Systems}
  \bibinfo{volume}{2}, \bibinfo{pages}{37--52}.
\newblock \DOIprefix\doi{10.1016/0169-7439(87)80084-9}.
\bibitem[{Wu et~al.(2019)Wu, Xiao, Wen, You and Hueni}]{Wu2019}
\bibinfo{author}{Wu, X.}, \bibinfo{author}{Xiao, Q.}, \bibinfo{author}{Wen,
  J.}, \bibinfo{author}{You, D.}, \bibinfo{author}{Hueni, A.},
  \bibinfo{year}{2019}.
\newblock \bibinfo{title}{Advances in quantitative remote sensing product
  validation: Overview and current status}.
\newblock \bibinfo{journal}{Earth-Science Reviews} \bibinfo{volume}{196},
  \bibinfo{pages}{102875}.
\newblock \DOIprefix\doi{10.1016/j.earscirev.2019.102875}.
\bibitem[{Wulder et~al.(2008)Wulder, White, Goward, Masek, Irons, Herold,
  Cohen, Loveland and Woodcock}]{Wulder2008}
\bibinfo{author}{Wulder, M.A.}, \bibinfo{author}{White, J.C.},
  \bibinfo{author}{Goward, S.N.}, \bibinfo{author}{Masek, J.G.},
  \bibinfo{author}{Irons, J.R.}, \bibinfo{author}{Herold, M.},
  \bibinfo{author}{Cohen, W.B.}, \bibinfo{author}{Loveland, T.R.},
  \bibinfo{author}{Woodcock, C.E.}, \bibinfo{year}{2008}.
\newblock \bibinfo{title}{{Landsat continuity: Issues and opportunities for
  land cover monitoring}}.
\newblock \bibinfo{journal}{Remote Sensing of Environment}
  \bibinfo{volume}{112}, \bibinfo{pages}{955--969}.
\newblock \DOIprefix\doi{10.1016/j.rse.2007.07.004}.
\bibitem[{Xia et~al.(2020)Xia, Huang and Wang}]{Xia2020}
\bibinfo{author}{Xia, K.}, \bibinfo{author}{Huang, J.}, \bibinfo{author}{Wang,
  H.}, \bibinfo{year}{2020}.
\newblock \bibinfo{title}{Lstm-cnn architecture for human activity
  recognition}.
\newblock \bibinfo{journal}{IEEE Access} \bibinfo{volume}{8},
  \bibinfo{pages}{56855--56866}.
\newblock \DOIprefix\doi{10.1109/ACCESS.2020.2982225}.
\bibitem[{Xie et~al.(2020)Xie, Liu and Wei}]{Xie2020}
\bibinfo{author}{Xie, T.}, \bibinfo{author}{Liu, R.}, \bibinfo{author}{Wei,
  Z.}, \bibinfo{year}{2020}.
\newblock \bibinfo{title}{{Improvement of the fast clustering algorithm
  improved by k-means in the big data}}.
\newblock \bibinfo{journal}{Applied Mathematics and Nonlinear Sciences}
  \bibinfo{volume}{5}, \bibinfo{pages}{1--10}.
\newblock \DOIprefix\doi{10.2478/amns.2020.1.00001}.
\bibitem[{Xu et~al.(2019)Xu, Wang, Kong, Feng, Liu and Wu}]{Xu2019}
\bibinfo{author}{Xu, K.}, \bibinfo{author}{Wang, X.}, \bibinfo{author}{Kong,
  C.}, \bibinfo{author}{Feng, R.}, \bibinfo{author}{Liu, G.},
  \bibinfo{author}{Wu, C.}, \bibinfo{year}{2019}.
\newblock \bibinfo{title}{{Identification of hydrothermal alteration minerals
  for exploring gold deposits based on SVM and PCA using ASTER data: A case
  study of Gulong}}.
\newblock \bibinfo{journal}{Remote Sensing} \bibinfo{volume}{11},
  \bibinfo{pages}{3003}.
\newblock \DOIprefix\doi{10.3390/rs11243003}.
\bibitem[{Yang et~al.(2010)Yang, {Hwa Yang}, {B. Zhou} and {Y.
  Zomaya}}]{Yang2010}
\bibinfo{author}{Yang, P.}, \bibinfo{author}{{Hwa Yang}, Y.},
  \bibinfo{author}{{B. Zhou}, B.}, \bibinfo{author}{{Y. Zomaya}, A.},
  \bibinfo{year}{2010}.
\newblock \bibinfo{title}{{A review of ensemble methods in bioinformatics}}.
\newblock \bibinfo{journal}{Current Bioinformatics} \bibinfo{volume}{5},
  \bibinfo{pages}{296--308}.
\newblock \DOIprefix\doi{10.2174/157489310794072508}.
\bibitem[{Ye et~al.(2017)Ye, Tian, Ge and Sun}]{Ye2017}
\bibinfo{author}{Ye, B.}, \bibinfo{author}{Tian, S.}, \bibinfo{author}{Ge, J.},
  \bibinfo{author}{Sun, Y.}, \bibinfo{year}{2017}.
\newblock \bibinfo{title}{{Assessment of WorldView-3 data for lithological
  mapping}}.
\newblock \bibinfo{journal}{Remote Sensing} \bibinfo{volume}{9},
  \bibinfo{pages}{1132}.
\newblock \DOIprefix\doi{10.3390/rs9111132}.
\bibitem[{Yetkin et~al.(2004)Yetkin, Toprak and S{\"{u}}zen}]{Yetkin2004}
\bibinfo{author}{Yetkin, E.}, \bibinfo{author}{Toprak, V.},
  \bibinfo{author}{S{\"{u}}zen, M.}, \bibinfo{year}{2004}.
\newblock \bibinfo{title}{{Alteration mapping by remote sensing: Application to
  Hasandağ – Melendiz volcanic complex, Central Turkey}}, in:
  \bibinfo{booktitle}{20th ISPRS Congress on Technical Commission VII}.
\bibitem[{Yu et~al.(2012)Yu, Porwal, Holden and Dentith}]{Yu2012}
\bibinfo{author}{Yu, L.}, \bibinfo{author}{Porwal, A.},
  \bibinfo{author}{Holden, E.J.}, \bibinfo{author}{Dentith, M.C.},
  \bibinfo{year}{2012}.
\newblock \bibinfo{title}{{Towards automatic lithological classification from
  remote sensing data using support vector machines}}.
\newblock \bibinfo{journal}{Computers {\&} Geosciences} \bibinfo{volume}{45},
  \bibinfo{pages}{229--239}.
\newblock \DOIprefix\doi{10.1016/j.cageo.2011.11.019}.
\bibitem[{Zhang et~al.(2020)Zhang, Hu, Fu, Yang, Wu and Feng}]{Zhang2020}
\bibinfo{author}{Zhang, F.}, \bibinfo{author}{Hu, Z.}, \bibinfo{author}{Fu,
  Y.}, \bibinfo{author}{Yang, K.}, \bibinfo{author}{Wu, Q.},
  \bibinfo{author}{Feng, Z.}, \bibinfo{year}{2020}.
\newblock \bibinfo{title}{{A New Identification Method for Surface Cracks from
  UAV Images Based on Machine Learning in Coal Mining Areas}}.
\newblock \bibinfo{journal}{Remote Sensing} \bibinfo{volume}{12},
  \bibinfo{pages}{1571}.
\newblock \DOIprefix\doi{10.3390/rs12101571}.
\bibitem[{Zhang et~al.(2016)Zhang, Yi, Li, Wang, Tang, Zhong, Li, Wang and
  Bie}]{Zhang2016}
\bibinfo{author}{Zhang, T.}, \bibinfo{author}{Yi, G.}, \bibinfo{author}{Li,
  H.}, \bibinfo{author}{Wang, Z.}, \bibinfo{author}{Tang, J.},
  \bibinfo{author}{Zhong, K.}, \bibinfo{author}{Li, Y.}, \bibinfo{author}{Wang,
  Q.}, \bibinfo{author}{Bie, X.}, \bibinfo{year}{2016}.
\newblock \bibinfo{title}{{Integrating data of ASTER and Landsat-8 OLI (AO) for
  hydrothermal alteration mineral mapping in Duolong porphyry Cu-Au deposit,
  Tibetan Plateau, China}}.
\newblock \bibinfo{journal}{Remote Sensing} \bibinfo{volume}{8},
  \bibinfo{pages}{890}.
\newblock \DOIprefix\doi{10.3390/rs8110890}.
\bibitem[{Zhao et~al.(2020)Zhao, Deng, Li, Wang and Wei}]{Zhao2020}
\bibinfo{author}{Zhao, H.}, \bibinfo{author}{Deng, K.}, \bibinfo{author}{Li,
  N.}, \bibinfo{author}{Wang, Z.}, \bibinfo{author}{Wei, W.},
  \bibinfo{year}{2020}.
\newblock \bibinfo{title}{{Hierarchical spatial-spectral feature extraction
  with long short term memory (LSTM) for mineral identification using
  hyperspectral imagery}}.
\newblock \bibinfo{journal}{Sensors} \bibinfo{volume}{20},
  \bibinfo{pages}{6854}.
\newblock \DOIprefix\doi{10.3390/s20236854}.
\bibitem[{Zhou and Guan(2011)}]{Zhou2011}
\bibinfo{author}{Zhou, M.L.}, \bibinfo{author}{Guan, Z.Q.},
  \bibinfo{year}{2011}.
\newblock \bibinfo{title}{Active-passive microwave remote sensing data
  combination for retrieval of soil moisture}, in:
  \bibinfo{booktitle}{Proceedings of SPIE - The International Society for
  Optical Engineering}.
\newblock \DOIprefix\doi{10.1117/12.912598}.
\bibitem[{Zhu and Goldberg(2009)}]{Zhu2009}
\bibinfo{author}{Zhu, X.}, \bibinfo{author}{Goldberg, A.B.},
  \bibinfo{year}{2009}.
\newblock \bibinfo{title}{Introduction to semi-supervised learning}.
\newblock \bibinfo{journal}{Synthesis Lectures on Artificial Intelligence and
  Machine Learning} \bibinfo{volume}{3}, \bibinfo{pages}{1--130}.
\newblock \DOIprefix\doi{10.2200/S00196ED1V01Y200906AIM006}.
\bibitem[{Zoheir et~al.(2019)Zoheir, Emam, Abdel-Wahed and
  Soliman}]{Zoheir2019}
\bibinfo{author}{Zoheir, B.}, \bibinfo{author}{Emam, A.},
  \bibinfo{author}{Abdel-Wahed, M.}, \bibinfo{author}{Soliman, N.},
  \bibinfo{year}{2019}.
\newblock \bibinfo{title}{{Multispectral and radar data for the setting of gold
  mineralization in the Southeastern Desert, Egypt}}.
\newblock \bibinfo{journal}{Remote Sensing} \bibinfo{volume}{11},
  \bibinfo{pages}{1450}.
\newblock \DOIprefix\doi{10.3390/rs11121450}.

\end{thebibliography}

\end{document}